%% file: neurips_2026.tex
\documentclass{article}

\PassOptionsToPackage{semicolon, round, sort}{natbib}

\usepackage[main, final]{neurips_2026}

\usepackage{graphicx}
\usepackage[utf8]{inputenc} 
\usepackage[T1]{fontenc}    
\usepackage{hyperref}       
\usepackage{url}            
\usepackage{booktabs}       
\usepackage{amsfonts}       
\usepackage{nicefrac}       
\usepackage{microtype}      
\usepackage{xcolor}         
\definecolor{darkblue}{rgb}{0.0,0.0,0.65}
\definecolor{darkred}{rgb}{0.65,0.0,0.0}
\definecolor{darkgreen}{rgb}{0.0,0.65,0.0}
\hypersetup{
	colorlinks = true,
	citecolor  = darkblue,
	linkcolor  = darkred,
	filecolor  = darkblue,
	urlcolor   = darkblue,
}
\usepackage{subcaption}
\usepackage{adjustbox}
\usepackage[cjk]{kotex}
\input{math_commands}

\usepackage{algorithm}
\usepackage{algpseudocode}
\usepackage{placeins}
\usepackage{hyperref}
\usepackage{threeparttable}
\usepackage{wrapfig}
\usepackage{tikz}
\usepackage{graphicx}
\usetikzlibrary{arrows.meta,calc}
\definecolor{surface}{RGB}{210,210,210}
\definecolor{mesh}{RGB}{120,120,120}
\definecolor{river}{RGB}{0,140,135}
\definecolor{gdblue}{RGB}{10,55,145}
\definecolor{iterred}{RGB}{190,35,45}
\definecolor{gradred}{RGB}{230,90,65}
\definecolor{gdgreen}{RGB}{30,140,85}
\definecolor{interpcolor}{RGB}{230,135,35}
\definecolor{floorbright}{RGB}{224,210,255}
\definecolor{walldark}{RGB}{78,42,132}

\usepackage{mathtools}
\usepackage{siunitx}
\usepackage[capitalize,noabbrev]{cleveref}
\definecolor{darkorange}{RGB}{190,85,0}

\theoremstyle{plain}
\newtheorem{theorem}{Theorem}[section]

\theoremstyle{definition}
\newtheorem{definition}[theorem]{Definition}

\theoremstyle{remark}

\title{AMUSE: \\ Anytime Muon with Stable Gradient Evaluation}

\author{
  \makebox[0.29\textwidth][c]{Jueun Kim\thanks{Authors contributed equally to this paper.}} \\
  KAIST \\
  \makebox[0pt][c]{\texttt{jueunkim@kaist.ac.kr}}
  \And
  \makebox[0.29\textwidth][c]{Baekrok Shin\footnotemark[1]} \\
  KAIST \\
  \makebox[0pt][c]{\texttt{br.shin@kaist.ac.kr}}
  \And
  \makebox[0.29\textwidth][c]{Jihun Yun} \\
  KRAFTON \\
  \makebox[0pt][c]{\texttt{jihuny@krafton.com}}
  \AND
  \makebox[0.29\textwidth][c]{Beomhan Baek\thanks{Work done during internships at KRAFTON.}} \\
  UCLA \\
  \makebox[0pt][c]{\texttt{beomhanbaek@ucla.edu}}
  \And
  \makebox[0.29\textwidth][c]{Minhak Song\footnotemark[2]} \\
  Stanford University \\
  \makebox[0pt][c]{\texttt{minhaks@stanford.edu}}
  \And
  \makebox[0.29\textwidth][c]{Chulhee Yun} \\
  KAIST \\
  \makebox[0pt][c]{\texttt{chulhee.yun@kaist.ac.kr}}
}

\usepackage{fontawesome5}

\begin{document}

\maketitle
\vspace{-1em}
\begin{center}
{\hypersetup{urlcolor=purple}
\href{https://github.com/kjeiun/amuse}{\faGithub\ \texttt{https://github.com/kjeiun/amuse}}}
\end{center}
\vspace{0.1em}

\begin{abstract}
    Modern deep learning commonly relies on AdamW with prescribed learning rate schedules, but recent works challenge both components: Schedule-Free optimization removes explicit schedules via iterate averaging, and Muon improves the update geometry by orthogonalizing momentum for matrix parameters. Despite Muon's strong empirical performance, its underlying mechanism remains partially understood. We study Muon through the river-valley loss landscape, where useful training progress occurs along a flat, low-curvature bulk subspace (the river), while high-curvature dominant directions form steep valley walls that induce oscillations. We empirically show that while Muon's orthogonalization accelerates river progress by increasing the bulk component, it also amplifies dominant-direction noise, causing oscillatory trajectories. Building on this, we propose Anytime MUon with Stable gradient Evaluation (AMUSE), which integrates Muon's rapid bulk progress with the stabilizing effect of Schedule-Free averaging. AMUSE uses a time-varying interpolation coefficient that initially evaluates gradients near the fast Muon sequence for rapid adaptation, then gradually shifts toward the stable averaged sequence to suppress valley-wall oscillations. As a result, AMUSE requires no learning rate schedules and supports anytime training. Across vision tasks and large language model pretraining, AMUSE consistently improves the performance-iteration Pareto frontier over (Schedule-Free) AdamW and Muon.
\end{abstract}

\input{01_Introduction}
\input{02_Related_Works}
\input{03_SF+Muon}
\input{04_AMUSE}
\input{05_Experiments}
\input{06_Conclusion}
\bibliographystyle{plainnat}    
\bibliography{references}

\newpage
\setcounter{tocdepth}{2}
\tableofcontents
\newpage
\appendix
\input{A_Quadratic_Example}
\clearpage
\input{B_Experiments_for_Section_3}

\clearpage
\input{C_AMUSE}
\clearpage
\input{D_Image_Domain_Experimental_Details}
\clearpage
\input{E_Large_Language_Model_Experimental_Details}
\clearpage

\input{F_Additional_Experiments}

\newpage
\input{checklist.tex}

\end{document}

%% file: math_commands.tex
\usepackage{amsthm}
\usepackage{latexsym}
\usepackage{amsmath}
\usepackage{amssymb}
\usepackage{bm}
\usepackage{mathrsfs}
\usepackage{url}
\usepackage{bbm}
\usepackage{enumitem}
\usepackage{nicefrac}
\usepackage{multirow}

\usepackage{multicol}

\newcommand{\mycomment}[1]{}
\newcommand{\GG}[1]{}

\def\eqref#1{(\ref{#1})}

\def\vm{{\bm{m}}}

\def\vu{{\bm{u}}}
\def\vv{{\bm{v}}}

\def\vx{{\bm{x}}}
\def\vy{{\bm{y}}}
\def\vz{{\bm{z}}}

\def\mA{{\bm{A}}}

\def\mG{{\bm{G}}}

\def\mI{{\bm{I}}}

\def\mM{{\bm{M}}}

\def\mO{{\bm{O}}}
\def\mP{{\bm{P}}}
\def\mQ{{\bm{Q}}}

\def\mU{{\bm{U}}}
\def\mV{{\bm{V}}}
\def\mW{{\bm{W}}}
\def\mX{{\bm{X}}}
\def\mY{{\bm{Y}}}
\def\mZ{{\bm{Z}}}

\DeclareMathAlphabet{\mathsfit}{\encodingdefault}{\sfdefault}{m}{sl}
\SetMathAlphabet{\mathsfit}{bold}{\encodingdefault}{\sfdefault}{bx}{n}

\def\gL{{\mathcal{L}}}

\def\gS{{\mathcal{S}}}
\def\gT{{\mathcal{T}}}

%% file: 01_Introduction.tex
\vspace{-2pt}
\section{Introduction}

Optimization remains a primary lever for improving training efficiency in deep learning. For many modern architectures, especially Transformer-based language models, the standard recipe has long combined the Adam(W) optimizer~\citep{kingma2015adam,loshchilov2018decoupled} with warmup and a prescribed learning rate decay schedule such as cosine decay~\citep{loshchilov2017sgdr}.

Recent work challenges both components of this recipe. Schedule-Free AdamW~\citep{defazio2024the} removes the need for explicit learning rate schedules by evaluating gradients at interpolated points between the current and averaged iterates, while using the averaged iterate for inference. It has shown strong empirical performance relative to schedule-based baselines, including winning the Self-Tuning track of the 2024 AlgoPerf Challenge~\citep{dahl2023benchmarking,kasimbeg2025accelerating}.

A complementary advance changes the underlying geometry of the update itself. Muon~\citep{jordan2024muon} exploits matrix structure by applying momentum to matrix-valued parameters and orthogonalizing the resulting update direction. Muon and its variants have been reported to outperform AdamW in language model pretraining~\citep{liu2025muon}, and have recently been adopted in large-scale foundation-model pretraining, including Kimi-K2/K3~\citep{kimi2025kimik2,team2026kimi}, GLM-5~\citep{glm2026glm5}, DeepSeek-V4/V4.1-Flash~\citep{deepseekai2026deepseekv4,deepseekai2026deepseekv41flash}, MAI-Base-2~\citep{microsoft2026frontier}, and Qwen3.8-Flash-Next~\citep{qwen2026flashnext}. Despite this empirical success, \emph{why orthogonalized momentum improves optimization} remains only partially understood.

We take this question as the first step toward optimizer design, and analyze Muon from the perspective of loss landscape geometry. Deep learning optimization takes place in high-dimensional parameter spaces whose loss
landscapes are highly anisotropic. Empirically, the Hessian spectrum often contains a small number of large outlier eigenvalues and a much larger bulk of relatively small eigenvalues. This spectral structure induces a local decomposition of the parameter space into a low-dimensional, high-curvature \emph{dominant subspace} and a high-dimensional, low-curvature \emph{bulk subspace}~\citep{sagun2016eigenvalues,sagun2017empirical,
papyan2018full,papyan2019measurements,JMLR:v21:20-933,ghorbani2019investigation}.

Recent work casts this anisotropy as the \emph{river-valley landscape}~\citep{song2025does,wen2025understanding}: the dominant subspace forms steep valley walls, while the bulk subspace forms a relatively flat river. In this picture, training progress occurs mainly along the river, whereas dominant subspace components cause valley-wall oscillations. This suggests two core principles for optimizer design: \emph{(i)} take large, consistent steps along the river and \emph{(ii)} suppress unnecessary updates across the valley.

We find that Muon naturally promotes the first part of this principle. By orthogonalizing the momentum matrix, Muon prevents the update from being dominated by a few large singular components, producing a more balanced matrix-valued step. 
Empirically, Muon updates contain substantially larger bulk components than SGD or Adam. Moreover, orthogonalization itself explicitly increases this relative bulk component. This provides a geometric explanation for Muon's efficiency: \emph{orthogonalized momentum accelerates progress along the river directions}.

However, this benefit comes with a tradeoff. Because orthogonalization is not selective, it can amplify noisy components as well as useful ones. In particular, small noise aligned with high-curvature dominant directions can induce oscillations across the valley walls after orthogonalization, leading to less stable trajectories~\citep{qi2026delving,zhang2026namo,he2025root,shen2026understandingpowerlimitsmuon}.

Schedule-Free (SF) optimization offers a complementary stabilizing mechanism by evaluating gradients at an interpolation between the fast optimizer sequence and an averaged inference sequence.
\citet{song2025through} show that this mechanism can keep the inference trajectory close to the river without requiring learning rate decay. We find this particularly beneficial for Muon: evaluating gradients closer to the averaged sequence reduces the high-curvature component before orthogonalization, thereby suppressing oscillations between the valley walls and stabilizing the resulting trajectory.

Motivated by this complementarity, we propose \textbf{AMUSE} (Anytime MUon with Stable gradient Evaluation). AMUSE combines Muon's orthogonalized matrix updates with the SF formulation, using a \emph{time-varying} interpolation coefficient.
AMUSE initially evaluates gradients near the fast Muon sequence for rapid adaptation, then gradually shifts toward the averaged sequence, where gradients are less dominated by high-curvature components.
Because this schedule is independent of the training horizon and requires no learning rate decay, AMUSE naturally supports \emph{anytime} training.

We evaluate AMUSE across a diverse set of benchmarks spanning image classification, segmentation, and large language model pretraining. The results show that AMUSE consistently improves the performance-iteration Pareto frontier over (Schedule-Free) AdamW and Muon. Notably, AMUSE reaches Muon's final performance with substantially fewer training steps across diverse settings. In particular, it requires \(1.51\times\), \(1.34\times\), and \(2.57\times\) fewer steps on 720M Llama pretraining, ImageNet with ResNet-50, and ImageNet with ViT fine-tuning, respectively.

%% file: 02_Related_Works.tex
\section{Related Work}
\label{sec:related-work}
\subsection{River-Valley Loss Landscape}

Previous studies have characterized neural network training through a low-rank Hessian structure, consisting of a few high-curvature directions and a much broader bulk component~\citep{ghorbani2019investigation, papyan2018full, papyan2019measurements, JMLR:v21:20-933, sagun2016eigenvalues, sagun2017empirical, yao2020pyhessian}. Recent work further interprets this geometry through the river-valley landscape (see Figure~\ref{fig:river_valley} for illustration), where dominant directions form steep valley walls while the bulk defines the flatter river floor where most useful progress occurs~\citep{song2025does, wen2025understanding, belloni2025universal, cohen2025understanding,zhou2025bsfa,wang2024improving, wang2025the,deng2026suspicious}.

\citet{wen2025understanding} use this geometry to explain the success of Warmup-Stable-Decay (WSD) schedules in large language model pretraining: the stable phase facilitates rapid traversal along the river, while the decay phase helps the iterate settle onto the valley floor. Extending this perspective, \citet{belloni2025universal} show that a similar geometry also arises in convolutional networks trained on image classification tasks. \citet{song2025does} show that projecting updates only onto the dominant subspace is insufficient to reduce training loss, while updates projected onto the bulk subspace remain effective.
Recently, several works have leveraged the river-valley landscape to design optimizers that either amplify updates along river directions or stabilize oscillations along valley directions~\citep{zhu2026accelerating, zhang2026mousse, liu2025focus, luo2026how}.

\subsection{Schedule-Free Optimizer}
\label{sec:sf}
\citet{defazio2024the} introduce the schedule-free (SF) optimizer as an interpolation between Polyak--Ruppert (PR) averaging \citep{ruppert1988efficient, polyak1992acceleration} and primal averaging~\citep{nesterov2009primal, tao2018primal}. Its update is given by
\begin{equation}
\begin{aligned}
    \vy_{t} & =(1-\beta)\vz_{t}+\beta\vx_{t}, \\
    \vz_{t+1} & =\vz_{t}-\eta \Delta_t, \\
    \vx_{t+1} & =\left(1-c_{t+1}\right)\vx_{t}+c_{t+1}\vz_{t+1},
\end{aligned}
\label{eq:sf}
\end{equation}  
where \(\eta\) is the learning rate, \(\Delta_t\) is the base optimizer update computed from the gradient at \(\vy_t\), \(c_{t+1}=1/(t+1)\), and the initialization satisfies \(\vz_1=\vx_1\). The framework can be combined with different base optimizers through the choice of $\Delta_t$. For example, defining $\Delta_t$ as a stochastic gradient at $\vy_t$ gives SF-SGD, whereas computing $\Delta_t$ using an RMSProp update with decoupled weight decay gives SF-AdamW.

The three sequences play distinct roles: $\vy_t$ is the gradient evaluation point, $\vz_t$ is the primary iterate that follows the base optimizer update, and $\vx_t$ is the averaged sequence used for inference. The interpolation parameter $\beta$ controls whether gradient evaluation tracks the fast-moving base trajectory ($\vz_t$) or the more stable averaged trajectory ($\vx_t$), thereby balancing rapid adaptation and stability. Under this formulation, SF optimizers are known to achieve a performance--time Pareto frontier, as evidenced by winning the Self-Tuning track of the 2024 AlgoPerf Challenge~\citep{dahl2023benchmarking}.

\citet{song2025through} later connect the empirical success of SF optimization to the river-valley landscape, showing that, with an appropriate choice of $\beta$, the inference trajectory $\vx_t$ of SF-AdamW follows the river throughout training.
Their findings suggest that the implicit averaging in SF optimizer filters out high-curvature fluctuations from the valley walls. As a result, SF optimizer can remain near the river without requiring an explicit decay phase or an additional average of the model parameters.

\subsection{Muon Optimizer}
\label{subsec:rel-muon}

Muon~\citep{jordan2024muon} is a recently introduced optimizer that leverages matrix structure by orthogonalizing matrix-valued updates. At each iteration $t$, given a matrix $\mW_t$ and its gradient $\mG_t$, the updates are defined as:
\begin{equation}
\begin{aligned}
    \mM_t &= \mu \mM_{t-1} + (1-\mu)\mG_t, \\
    \mW_{t+1} &= \mW_t - \eta \gT(\mM_t),
\end{aligned}
\label{eqn: muon}
\end{equation}
where $\eta$ is the learning rate, $\mu$ is the momentum coefficient, and $\gT$ denotes an orthogonalization operator, which is approximated using a Newton-Schulz iteration~\citep{bernstein2024old} for computational efficiency. For non-matrix-valued parameters, including embeddings and classification heads, updates are instead performed using standard optimizers such as SGD or Adam(W).

Recent theoretical studies have begun to explain Muon's rapid convergence through the normalization effect of its matrix-wise orthogonalization, especially in ill-conditioned settings. \citet{shen2025convergence} show that Muon can be advantageous when neural network Hessians exhibit low-rank or structured spectral geometry, while \citet{ma2026preconditioning} show that its orthogonalization can act as an effective preconditioner, yielding condition-number-free linear convergence in matrix factorization and linear transformer settings. Other recent studies further show that Muon's orthogonalized updates induce balanced component learning~\citep{kang2026uniform}, improving minority-group performance in imbalanced settings~\citep{vasudeva2026how}. However, this same mechanism can also strengthen noisy components, potentially leading to less stable dynamics~\citep{qi2026delving,zhang2026namo,shen2026understandingpowerlimitsmuon}.

%% file: 03_SF+Muon.tex
\section{A River-Valley Analysis of Muon and Schedule-Free Dynamics}
\label{sec:sfmuon_analysis}

In this section, we analyze Muon and Schedule-Free dynamics in controlled deep learning settings where the leading Hessian eigenspaces can be computed directly. We first define the dominant and bulk subspaces used in our measurements.  Section~\ref{subsec:muon_bulk} then shows that Muon produces more bulk-oriented updates than SGD or AdamW, but that orthogonalization can also amplify small components in the dominant subspace, inducing oscillations across the valley walls. Section~\ref{subsec:sfmuon} shows that Schedule-Free averaging mitigates this effect by evaluating gradients at points with smaller dominant components before orthogonalization.

\paragraph{Dominant \& Bulk Decomposition.}
Following~\citet{song2025does}, we decompose parameter-space directions using the Hessian eigenspaces of the loss $\gL:\mathbb{R}^d \to \mathbb{R}$.
\begin{definition}[Dominant and bulk subspaces]
    Let $\nabla^2 \gL(\theta) \in \mathbb{R}^{d \times d}$ have eigenvalues $\lambda_1 \ge \cdots \ge \lambda_d$ with corresponding eigenvectors $\vu_1(\theta), \dots, \vu_d(\theta)$. The top-$k$ dominant subspace is $\gS_k(\theta) := \mathrm{span}\{\vu_1(\theta), \dots, \vu_k(\theta)\}$, and its orthogonal complement $\mathcal{S}_k^\perp(\theta)$ is the bulk subspace.
    \label{def:dominant_bulk}
\end{definition}

\begin{definition}[Subspace projections]
    We define the projection matrix onto $\gS_k(\theta)$ as $\mP_k(\theta) := \sum_{i=1}^{k}\vu_i(\theta)\vu_i(\theta)^\top$, and the projection matrix onto $\gS_k^\perp(\theta)$ as $\mP_k^\perp(\theta) := \mI-\mP_k(\theta)$.
    For any nonzero vector $\vv \in \mathbb{R}^d$, we define
    \begin{equation}
        \label{eq:dom_bulk_ratio}
        r^{\mathrm{dom}}(\vv; \theta) = \frac{\|\mP_k(\theta) \vv\|_2}{\| \vv\|_2},
        \qquad
        r^{\mathrm{bulk}}(\vv; \theta) = \frac{\|\mP_k^\perp(\theta) \vv \|_2}{\|\vv\|_2}.
    \end{equation}
    \label{def:subspace_proj}
\end{definition}

\begin{wrapfigure}{r}{0.41\textwidth}
    \centering
    \input{river-valley_fig}
    \caption{Illustration of a river-valley landscape. SGD oscillates across the valley walls and progresses slowly along the river, while Muon advances faster but remains oscillatory.}
    \label{fig:river_valley}
    \vspace{-1em}
\end{wrapfigure}
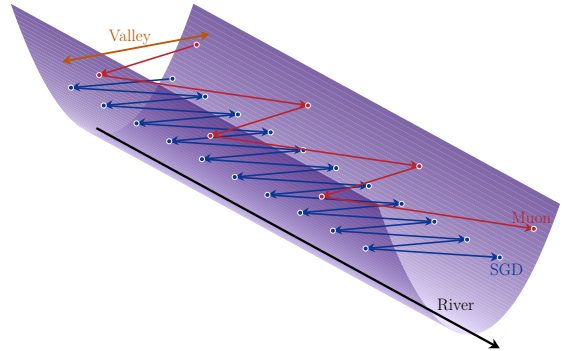

These quantities measure how strongly $\vv$ is aligned with the dominant and bulk subspaces, respectively. Since $\mP_k(\theta)$ and $\mP_k^\perp(\theta)$ are orthogonal projections, the ratios satisfy $(r^{\mathrm{dom}})^2+(r^{\mathrm{bulk}})^2=1$; hence, we report $r^{\mathrm{dom}}$ in most measurements. Following~\citet{song2025does}, we set $k$ to the number of classes, motivated by prior observations that classification Hessians exhibit exactly this many outlier eigenvalues~\citep{papyan2018full,papyan2019measurements}.

Before analyzing Muon's update decomposition, we verify that the dominant and bulk subspaces play distinct optimization roles in our controlled settings. Following \citet{zhou2025bsfa}, we independently scale the dominant and bulk components of each Muon update. As shown in Appendix~\ref{app:scaling_proj}, amplifying the \emph{bulk component} accelerates training, whereas amplifying the \emph{dominant component} can destabilize optimization. This supports the river-valley interpretation in our setting and motivates the dominant/bulk update-ratio analysis below.

\vspace{-6pt}
\subsection{Orthogonalization Makes Muon Updates More Bulk-Oriented}
\label{subsec:muon_bulk}

Muon applies an orthogonalization operator $\gT(\cdot)$ to the momentum matrix $\mM_t$. Although this operation acts on the singular-vector geometry of each matrix, we measure its effect in the Hessian eigenspace by projecting the flattened update onto the dominant and bulk subspaces. For notation simplicity, we slightly abuse notation and write $\gT(\vm_t) := \mathrm{vec}(\gT(\mM_t))$. 

To examine how orthogonalization redistributes optimizer updates across the Hessian eigenspace, we measure the dominant component ratio throughout training on a 5k-sample subset of MNIST using a 3-layer MLP. Experimental details are provided in Appendix~\ref{subsec:section3_exp_details}. Figure~\ref{fig:Muon bulk vs. SGD/AdamW bulk} compares the dominant ratios of optimizer updates. Muon produces much smaller dominant-subspace components than SGD or AdamW, indicating that its updates are less aligned with the high-curvature directions and more oriented toward the bulk subspace. The figure also includes AMUSE, introduced in Section~\ref{sec:amuse}, as a preview of the stabilized variant of Muon; its updates have even smaller dominant ratios.

Figure~\ref{fig:Muon pre bulk vs. post bulk} isolates the role of orthogonalization by comparing the raw momentum $\vm_t$ with the post-orthogonalized update $\Delta \theta_t (=\gT(\vm_t))$. For Muon, orthogonalization sharply reduces the dominant ratio relative to the raw momentum, showing that the orthogonalization step itself shifts the update away from dominant directions and toward the bulk subspace. This provides a geometric explanation for Muon's effectiveness: by increasing the relative bulk component, \emph{orthogonalized momentum promotes faster progress along the river directions}, where long-term learning takes place.

However, this normalization process prevents dominant components from dying out.
Even a small residual dominant component can induce non-negligible motion along high-curvature directions after normalization, leading to oscillations across the valley walls. This behavior resembles normalized-update dynamics, which can oscillate even on simple quadratic objectives~\citep{arora2022understanding}; see Appendix~\ref{app:quadratic_bouncing} for an example. 

\begin{figure}[t]
\centering

\begin{subfigure}[t]{0.47\linewidth}
    \centering
    \includegraphics[width=\linewidth]{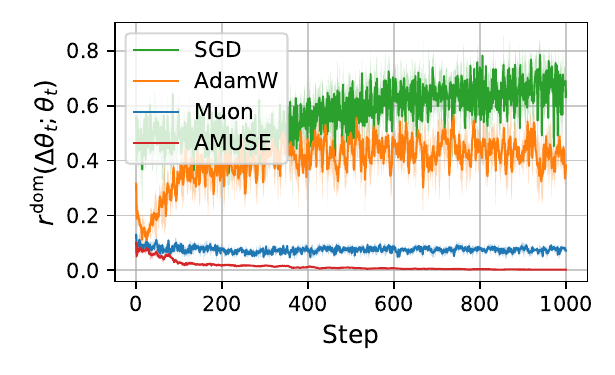}
    \caption{Muon vs. SGD/AdamW}
    \label{fig:Muon bulk vs. SGD/AdamW bulk}
\end{subfigure}
\hfill
\begin{subfigure}[t]{0.47\linewidth}
    \centering
    \includegraphics[width=\linewidth]{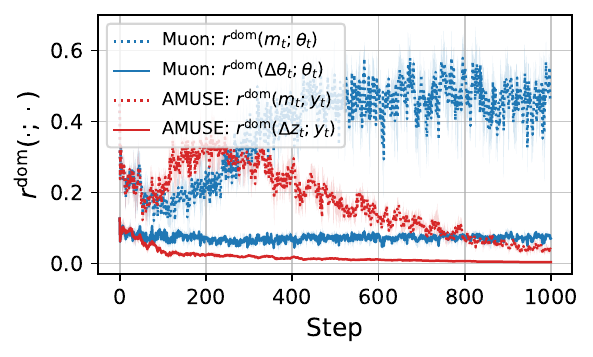}
    \caption{Momentum $\vm_t$ vs. Update $\Delta \theta_t \,(=\gT(\vm_t))$}
    \label{fig:Muon pre bulk vs. post bulk}
\end{subfigure}

\caption{Comparison of dominant component ratios. Evaluated on a 5k MNIST subset using a 3-layer MLP.(\textbf{a}) Muon consistently produces smaller dominant updates than SGD/AdamW, and AMUSE further suppresses the dominant component. (\textbf{b}) Orthogonalization reduces Muon’s dominant ratio compared to momentum $\vm_t$; in contrast, AMUSE maintains low dominant ratios throughout, reflecting more stable gradient dynamics. Averaged over three runs. Additional experiments on other datasets and architectures are provided in Appendix~\ref{app:additional experimental results for section 3}.}
\label{fig:muon_dom_bulk}
\end{figure}

\vspace{-3pt}
\subsection{Schedule-Free Stabilizes Muon Trajectories}
\label{subsec:sfmuon}

The previous subsection shows that Muon produces large bulk movement, but its orthogonalized updates may also amplify valley-wall oscillations. Prior work has shown that, with an appropriate choice of $\beta$, the averaged trajectory $\vx_t$ of SF-AdamW follows the river~\citep{song2025through}, suggesting that \textit{the averaging mechanism of the SF optimizer may mitigate this instability}. Motivated by this observation, we use Muon as the base optimizer under the SF optimizer (we refer to this variant as \emph{SF-Muon}) and analyze the resulting trajectories.

Taking Muon as the base optimizer, we obtain the following SF formulation for matrix-valued parameters:
\begin{equation}
\begin{aligned}
    \mY_t &= (1-\beta)\mZ_t + \beta \mX_t, \\
    \mM_t &= \mu \mM_{t-1} + (1-\mu) \nabla \gL(\mY_t), \\
    \mZ_{t+1} &= \mZ_t - \eta \gT(\mM_t), \\
    \mX_{t+1} &= \left(1-c_{t+1}\right)\mX_t
    + c_{t+1}\mZ_{t+1}.
\end{aligned}
\label{eq:sfmuon}
\end{equation}
Here, $\mZ_t$ is the base sequence updated by Muon, $\mY_t$ is the gradient-evaluation point, and $\mX_t$ is the averaged sequence used as the final model parameter, with $c_{t+1}=1/(t+1)$. 
The coefficient $\beta$ determines where the gradient is evaluated between the fast-moving base sequence $\mZ_t$ and the averaged sequence $\mX_t$. 
Although Eq.~\eqref{eq:sfmuon} is written for a single Muon-updated matrix block, not all parameters are updated by Muon. Non-matrix parameters are updated using their corresponding SF base optimizer, such as SF-SGD or SF-AdamW.

\begin{figure}
\begin{subfigure}[t]{0.48\linewidth}
    \centering
    \includegraphics[width=\linewidth]{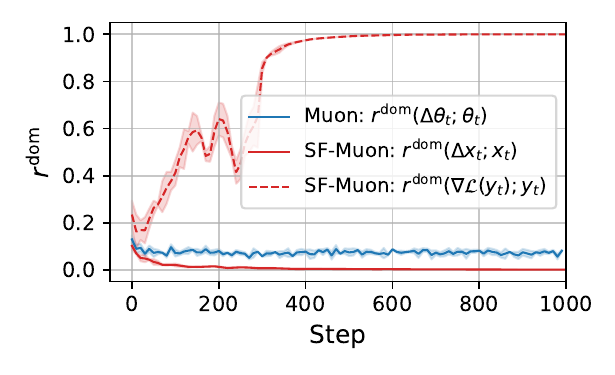}
    \caption{Comparison of dominant ratios}
    \label{fig: SF-Muon vs. Muon dom}
\end{subfigure}
\hfill
\begin{subfigure}[t]{0.48\linewidth}
    \centering
    \includegraphics[width=\linewidth]{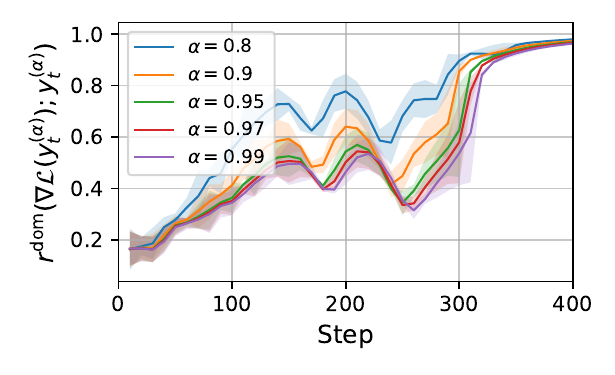}
    \caption{Dominant ratio of gradient at $\vy_t^{(\alpha)}$}
    \label{fig:SF-Muon gradient bulk dom}
\end{subfigure}
\caption{Comparison of dominant component ratios. Settings as in Figure~\ref{fig:muon_dom_bulk}. (\textbf{a}) Dominant ratios for Muon (blue) and SF-Muon (red). The solid and dashed red lines denote the dominant components of $\Delta \vx_t$ and $\nabla \gL(\vy_t)$, respectively. (\textbf{b}) Dominant ratios for gradients at $\vy_t^{(\alpha)}=(1-\alpha)\vz_t + \alpha \vx_t$ across varying $\alpha$, demonstrating that larger $\alpha$ values correlate with lower dominant components. Additional datasets and model architectures show consistent trends; see Appendix~\ref{app:additional experimental results for section 3}.}
\vspace{-3pt}
\label{fig:sfmuon_grad_components}
\end{figure}

\input{river-valley-interpolation}

\vspace{-3pt}
To understand how SF interacts with Muon, we measure subspace ratios in the full parameter space. Using the notation of Eq.~\eqref{eq:sf}, let $\vx_t,\vy_t,\vz_t \in \mathbb{R}^d$ denote the averaged sequence, gradient-evaluation point, and base sequence, respectively, obtained by flattening and concatenating all trainable parameters. With $\Delta \vx_t := \vx_{t+1}-\vx_t$, we compute $r^{\mathrm{dom}}(\Delta \vx_t;\vx_t)$ 
as well as the ratio for the gradient evaluated at $\vy_t$, $r^{\mathrm{dom}}(\nabla \gL(\vy_t);\vy_t)$.

The results in Figure~\ref{fig: SF-Muon vs. Muon dom} show that the update of $\vx_t$ is predominantly bulk-oriented, indicating that the trajectory primarily follows the river direction. In contrast, the gradient evaluated at $\vy_t$ still retains a substantial dominant component, which intensifies as training progresses. This suggests that while the averaging mechanism of SF stabilizes the $\vx_t$ trajectory, the gradient passed to Muon may still be computed at a noisy valley-wall location. This raises a natural question:

\vspace{-2pt}
\begin{center}
    \textit{Can we further reduce the dominant component before orthogonalization \\ by evaluating gradients at a point closer to the river?}
\end{center}

\vspace{-2pt}
To test this hypothesis, we use $\vx_t$ as a proxy for a point on the river and evaluate gradients at virtual interpolation points $\vy_t^{(\alpha)} = (1-\alpha)\vz_t + \alpha\vx_t$ for varying $\alpha \in [0,1]$, while keeping the actual training trajectory fixed by using a fixed $\beta$ during training (see Figure~\ref{fig:river_valley_sf} for an illustration). At each $\vy_t^{(\alpha)}$, we compute the gradient and measure its dominant ratio using Eq.~\eqref{eq:dom_bulk_ratio}.

Figure~\ref{fig:SF-Muon gradient bulk dom} shows that as $\alpha$ increases, the gradient becomes increasingly bulk-oriented. Thus, evaluating closer to $\vx_t$ suppresses valley-wall contributions before the Muon orthogonalization step, yielding a more stable update. Consistent with this mechanism, AMUSE---the time-varying SF-Muon variant introduced in Section~\ref{sec:amuse}---suppresses dominant components both before and after orthogonalization (Figure~\ref{fig:Muon pre bulk vs. post bulk}). In the language model experiment on 124M Llama pretraining (setup in Section~\ref{sec:amuse}), we observe a related effect at the trajectory level: increasing $\beta$ yields larger averaged-sequence updates $\|\Delta \vx_t\|$, indicating faster progress along the river direction (Figure~\ref{fig:sf-muon_varying_beta}, right).

This observation suggests evaluating gradients near the averaged sequence $\vx_t$ by using a large fixed $\beta$. 
However, using a large constant $\beta$ creates an early--late trade-off: early in training, $\vx_t$ may lag behind the rapidly adapting base sequence $\vz_t$, so evaluating too close to $\vx_t$ can slow initial progress toward the river. 
Conversely, a smaller $\beta$ preserves early adaptation but provides weaker late-stage stabilization.

%% file: river-valley_fig.tex
\resizebox{\linewidth}{!}{%
    \begin{tikzpicture}[
        x={(1.8cm,0.00cm)},
        y={(1.cm,0cm)},
        z={(0cm,1.3cm)},
        line cap=round,
        line join=round,
        >=Stealth
    ]

    \def\wall{1.28}
    \def\slope{0.42}
    \def\base{4.20}
    \def\xmax{1.55}   
    \def\ymax{11.20}  

    \foreach \k in {0,...,47} {
        \pgfmathsetmacro{\ua}{\k*\xmax/48}
        \pgfmathsetmacro{\ub}{(\k+1)*\xmax/48}
        \pgfmathtruncatemacro{\pct}{100*\k/47}

        \path[
            fill=walldark!\pct!floorbright,
            opacity=0.72
        ]
        (\ua,0,{\base+\wall*\ua*\ua})
        --
        (\ua,\ymax,{\base-\slope*\ymax+\wall*\ua*\ua})
        --
        (\ub,\ymax,{\base-\slope*\ymax+\wall*\ub*\ub})
        --
        (\ub,0,{\base+\wall*\ub*\ub})
        -- cycle;

        \path[
            fill=walldark!\pct!floorbright,
            opacity=0.72
        ]
        (-\ua,0,{\base+\wall*\ua*\ua})
        --
        (-\ua,\ymax,{\base-\slope*\ymax+\wall*\ua*\ua})
        --
        (-\ub,\ymax,{\base-\slope*\ymax+\wall*\ub*\ub})
        --
        (-\ub,0,{\base+\wall*\ub*\ub})
        -- cycle;
    }

    \draw[
        black,
        line width=2pt,
        -{Stealth[length=9.0pt,width=9.5pt]}
    ]
    (0,-0.15,{\base-\slope*(-0.15)+0.08})
    --
    (0,{\ymax+0.95},{\base-\slope*(\ymax+0.95)+0.08});
    \node[
    font=\Large,
    text=black,
    anchor=south west,
    xshift=-58pt,
    yshift=30pt
    ] at (0,{\ymax+0.95},{\base-\slope*(\ymax+0.95)+0.08}) {River};

    \draw[
        darkorange,
        line width=1.6pt,
        {Stealth[length=8.0pt,width=8.5pt]}-{Stealth[length=8.0pt,width=8.5pt]}
    ]
    (-1.15,0.85,{\base-\slope*0.85+\wall*1.15*1.15+0.38})
    --
    ( 1.35,0.85,{\base-\slope*0.85+\wall*1.35*1.35+0.38});

    \node[
        font=\Large,
        text=darkorange,
        anchor=south,
        xshift=0pt,
        yshift=0pt
    ] at (0,0.85,{\base-\slope*0.85+\wall*1.25*1.25+0.38}) {Valley};

    \foreach \i/\xx/\yy in {
        0/ 1.0/0.35,
        1/-1.0/0.85,
        2/ 1.0/1.35,
        3/-1.0/1.85,
        4/ 1.0/2.35,
        5/-1.0/2.85,
        6/ 1.0/3.35,
        7/-1.0/3.85,
        8/ 1.0/4.35,
        9/-1.0/4.85,
        10/1.0/5.35,
        11/-1.0/5.85,
        12/1.0/6.35,
        13/-1.0/6.85,
        14/1.0/7.35,
        15/-1.0/7.85,
        16/1.0/8.35,
        17/-1.0/8.85,
        18/1.0/9.35,
        19/-1.0/9.85,
        20/1.0/10.35
    } {
        \coordinate (G\i) at
        (\xx,\yy,{\base-\slope*\yy+\wall*\xx*\xx+0.18});
    }
    
    \foreach \i [evaluate=\i as \j using int(\i+1)] in {0,...,19} {
        \draw[
            gdblue,
            line width=1.4pt,
            -{Stealth[length=8.0pt,width=8.0pt]}
        ]
        (G\i) -- (G\j);
    }

    \foreach \i in {0,...,20} {
        \filldraw[
            fill=gdblue,
            draw=white,
            line width=0.48pt
        ]
        (G\i) circle[radius=2.2pt];
    }

    \coordinate (I0) at ( 1.30,0.55,{\base-\slope*0.55+\wall*1.30*1.30+0.18});
    \coordinate (I1) at (-1.30,2.25,{\base-\slope*2.25+\wall*1.30*1.30+0.18});
    \coordinate (I2) at ( 1.30,3.95,{\base-\slope*3.95+\wall*1.30*1.30+0.18});
    \coordinate (I3) at (-1.30,5.65,{\base-\slope*5.65+\wall*1.30*1.30+0.18});
    \coordinate (I4) at ( 1.30,7.35,{\base-\slope*7.35+\wall*1.30*1.30+0.18});
    \coordinate (I5) at (-1.30,9.05,{\base-\slope*9.05+\wall*1.30*1.30+0.18});
    \coordinate (I6) at ( 1.30,10.85,{\base-\slope*10.85+\wall*1.30*1.30+0.18});

    \foreach \i [evaluate=\i as \j using int(\i+1)] in {0,...,5} {
        \draw[
            iterred,
            line width=1.4pt,
            -{Stealth[length=8pt,width=8pt]}
        ]
        (I\i) -- (I\j);
    }

    \foreach \i in {0,...,6} {
        \filldraw[
            fill=iterred,
            draw=white,
            line width=0.48pt
        ]
        (I\i) circle[radius=2.2pt];
    }

    \node[
        font=\Large,
        text=iterred,
        anchor=west,
        xshift=-23pt,
        yshift=10pt
    ] at (I6) {Muon};
    
    \node[
        font=\Large,
        text=gdblue,
        anchor=west,
        xshift=-12pt,
        yshift=-10pt
    ] at (G20) {SGD};

    \end{tikzpicture}
    }

%% file: river-valley-interpolation.tex
\begin{wrapfigure}{r}{0.337\textwidth}
    \vspace{-1.6em}
    \centering
    \resizebox{\linewidth}{!}{
    \begin{tikzpicture}[
        x={(1.05cm,0.00cm)},
        y={(0.05cm,0.32cm)},
        z={(0cm,1.12cm)},
        line cap=round,
        line join=round,
        >=Stealth
    ]

    \def\wall{0.8}
    \def\slope{0.38}
    \def\base{4.20}
    \def\xmax{1.15}   
    \def\ymax{11.3}  

    \foreach \k in {0,...,47} {
        \pgfmathsetmacro{\ua}{\k*\xmax/48}
        \pgfmathsetmacro{\ub}{(\k+1)*\xmax/48}
        \pgfmathtruncatemacro{\pct}{100*\k/47}

        \path[
            fill=walldark!\pct!floorbright,
            opacity=0.72
        ]
        (\ua,0,{\base+\wall*\ua*\ua})
        --
        (\ua,\ymax,{\base-\slope*\ymax+\wall*\ua*\ua})
        --
        (\ub,\ymax,{\base-\slope*\ymax+\wall*\ub*\ub})
        --
        (\ub,0,{\base+\wall*\ub*\ub})
        -- cycle;

        \path[
            fill=walldark!\pct!floorbright,
            opacity=0.72
        ]
        (-\ua,0,{\base+\wall*\ua*\ua})
        --
        (-\ua,\ymax,{\base-\slope*\ymax+\wall*\ua*\ua})
        --
        (-\ub,\ymax,{\base-\slope*\ymax+\wall*\ub*\ub})
        --
        (-\ub,0,{\base+\wall*\ub*\ub})
        -- cycle;
    }
    \foreach \i/\xx/\yy in {
        0/ 0.018/1.00,
        1/-0.017/2.45,
        2/ 0.016/4.15,
        3/-0.015/5.85,
        4/ 0.014/7.65,
        5/-0.013/9.45,
        6/ 0.012/11.25
    }
    {
        \coordinate (G\i) at
        (\xx,\yy,{\base-\slope*\yy+\wall*\xx*\xx+0.18});
    }
    
    \foreach \i [evaluate=\i as \j using int(\i+1)] in {0,...,5} {
        \draw[
            gdblue,
            line width=0.7pt,
            -{Stealth[length=4.0pt,width=4.0pt]}
        ]
        (G\i) -- (G\j);
    }
    
    \foreach \i in {0,...,6} {
        \filldraw[
            fill=gdblue,
            draw=white,
            line width=0.28pt
        ]
        (G\i) circle[radius=1.5pt];
    }
    \coordinate (H0) at ( 0.82,1.00,{\base-\slope*1.00+\wall*0.82*0.82+0.26});
    \coordinate (H1) at (-0.82,6.10,{\base-\slope*6.10+\wall*0.82*0.82+0.26});
    \coordinate (H2) at ( 0.82,11.25,{\base-\slope*11.25+\wall*0.82*0.82+0.26});

    \coordinate (I0) at ( 0.32,1.00,{\base-\slope*1.00+\wall*0.32*0.32+0.18});
    \coordinate (I1) at (-0.32,2.60,{\base-\slope*2.60+\wall*0.32*0.32+0.18});
    \coordinate (I2) at ( 0.32,4.20,{\base-\slope*4.20+\wall*0.32*0.32+0.18});
    \coordinate (I3) at (-0.32,5.80,{\base-\slope*5.80+\wall*0.32*0.32+0.18});
    \coordinate (I4) at ( 0.32,7.40,{\base-\slope*7.40+\wall*0.32*0.32+0.18});
    \coordinate (I5) at (-0.32,9.00,{\base-\slope*9.00+\wall*0.32*0.32+0.18});
    
    \coordinate (I6) at ($(G6)!0.50!(H2)$);
    
    \foreach \i [evaluate=\i as \j using int(\i+1)] in {0,...,5} {
        \draw[
            iterred,
            line width=0.7pt,
            -{Stealth[length=4pt,width=4pt]}
        ]
        (I\i) -- (I\j);
    }
    
    \foreach \i in {0,...,6} {
        \filldraw[
            fill=iterred,
            draw=white,
            line width=0.28pt
        ]
        (I\i) circle[radius=2.pt];
    }
    \foreach \i [evaluate=\i as \j using int(\i+1)] in {0,...,1} {
        \draw[
            gdgreen,
            line width=0.8pt,
            -{Stealth[length=4.0pt,width=4.0pt]}
        ]
        (H\i) -- (H\j);
    }
    
    \foreach \i in {0,...,2} {
        \filldraw[
            fill=gdgreen,
            draw=white,
            line width=0.28pt
        ]
        (H\i) circle[radius=2.0pt];
    }
    \draw[
        black,
        line width=0.55pt,
        opacity=0.85
    ]
    (G6) -- (I6) -- (H2);

    
    \coordinate (Cbottom) at ($(G6)!0.13!(H2)$);
    \coordinate (Cmiddle) at ($(G6)!0.3!(H2)$);
    \coordinate (Ctop)    at ($(G6)!0.68!(H2)$);
    \coordinate (Cttop)   at ($(G6)!0.88!(H2)$);

    \foreach \C in {Cbottom,Cmiddle,Ctop,Cttop} {
        \filldraw[
            fill=interpcolor,
            draw=white,
            line width=0.28pt
        ]
        (\C) circle[radius=1.15pt];
    }

    \coordinate (Rbottom) at ($(Cbottom)+(0.00,1.55,{-\slope*1.55})$);
    \coordinate (Rmiddle) at ($(Cmiddle)+(0.00,1.55,{-\slope*1.55})$);
    \coordinate (Rtop)    at ($(Ctop)   +(0.00,1.35,{-\slope*1.35})$);
    \coordinate (Rttop)   at ($(Cttop)  +(0.00,0.8,{-\slope*0.8})$);

    \coordinate (Lbottom) at ($(Cbottom)!0.18!(G6)$);
    \coordinate (Lmiddle) at ($(Cmiddle)!0.24!(G6)$);
    \coordinate (Ltop)    at ($(Ctop)!0.30!(G6)$);
    \coordinate (Lttop)   at ($(Cttop)!0.2!(G6)$);

    \coordinate (Ebottom) at (Rbottom);
    \coordinate (Emiddle) at ($(Rmiddle)!0.2!(Lmiddle)$);
    \coordinate (Etop)    at ($(Rtop)!0.5!(Ltop)$);
    \coordinate (Ettop)   at (Lttop);

    \draw[
        interpcolor,
        line width=0.5pt,
        -{Stealth[length=3.0pt,width=3.0pt]},
        opacity=0.95
    ]
    (Cbottom) -- (Ebottom);
    
    \draw[
        interpcolor,
        line width=0.5pt,
        -{Stealth[length=3.0pt,width=3.0pt]},
        opacity=0.95
    ]
    (Cmiddle) -- (Emiddle);
    
    \draw[
        interpcolor,
        line width=0.5pt,
        -{Stealth[length=3.0pt,width=3.0pt]},
        opacity=0.95
    ]
    (Ctop) -- (Etop);
    
    \draw[
        interpcolor,
        line width=0.5pt,
        -{Stealth[length=3.0pt,width=3.0pt]},
        opacity=0.95
    ]
    (Cttop) -- (Ettop);

    \node[
        text=gdgreen,
        font=\scriptsize\bfseries,
        anchor=west,
        xshift=-8pt,
        yshift=4.5pt
    ] at (H2) {$z_t$};
    
    \node[
        text=iterred,
        font=\scriptsize\bfseries,
        anchor=west,
        xshift=-4.5pt,
        yshift=-4pt
    ] at (I6) {$y_t$};
    
    \node[
        text=gdblue,
        font=\scriptsize\bfseries,
        anchor=east,
        xshift=3pt,
        yshift=-1pt
    ] at (G6) {$x_t$};
    
    \end{tikzpicture}
    }
    \caption{Schedule-free iterates in a river-valley landscape. Orange arrows indicate negative gradients computed at each $\vy_t^{(\alpha)}$, exhibiting reduced dominant components when evaluated closer to the river.}
    \label{fig:river_valley_sf}
    \vspace{-1em}
\end{wrapfigure}
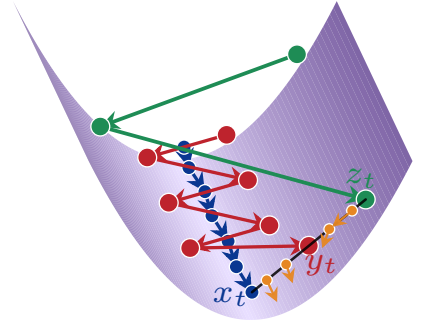

%% file: 04_AMUSE.tex
\section{AMUSE: Anytime MUon with Stable gradient Evaluation}
\vspace{-4pt}
\label{sec:amuse}
To address the trade-off between early progress and later stability induced by a fixed $\beta$, we introduce \textbf{Anytime MUon with Stable gradient Evaluation (AMUSE)}. Instead of fixing the interpolation coefficient throughout training, AMUSE employs a time-varying coefficient $\beta_t$. It starts with a small $\beta_t$ to allow fast early adaptation and gradually increases it, shifting the gradient-evaluation point from the fast base trajectory ($\vz_t$) toward the smoother averaged trajectory ($\vx_t$). Formally, AMUSE uses the same update rule as Eq.~\eqref{eq:sfmuon}, except the gradient is now evaluated at 
$\mY_t = (1-\beta_t)\mZ_t + \beta_t\mX_t.$

Let $T_0$ denote the number of warmup steps. During warmup, we fix $\beta_t = \beta_{\rm init}$. For $t \ge T_0$, we set:
\begin{equation}
    \beta_t =  1 - \left(\frac{c_{t+1}(1-c_{T_0+1})}{c_{T_0+1}(1-c_{t+1})}\right)^\rho(1-\beta_{\rm init}),
\label{eq:amuse_beta}
\end{equation}
where $\rho \in [0,1]$ controls how quickly $\beta_t$ increases. Under the convention $c_{t+1}=1/(t+1)$, Eq.~\eqref{eq:amuse_beta} simplifies to
\begin{equation*}
    \beta_t = 1-\left(\frac{T_0}{t}\right)^\rho(1-\beta_{\rm init}).
\end{equation*}
When $\rho=0$, AMUSE reduces to fixed-$\beta$ SF-Muon with $\beta_t=\beta_{\rm init}$. Larger values of $\rho$ move the gradient-evaluation point toward $\vx_t$ more rapidly. We defer the detailed rationale and derivation of Eq.~\eqref{eq:amuse_beta} to Appendix~\ref{app:amuse_design}.

Notably, \textit{this time-varying $\beta_t$ is independent of the total number of training steps}, gradually approaching $1$ as training progresses. Furthermore, like the standard SF method, AMUSE does not require any learning rate schedule. Algorithm~\ref{alg:amuse} in Appendix~\ref{app:algorithm} summarizes the complete AMUSE procedure.

\begin{figure}[t]
    \centering
    \includegraphics[width=0.85\linewidth]{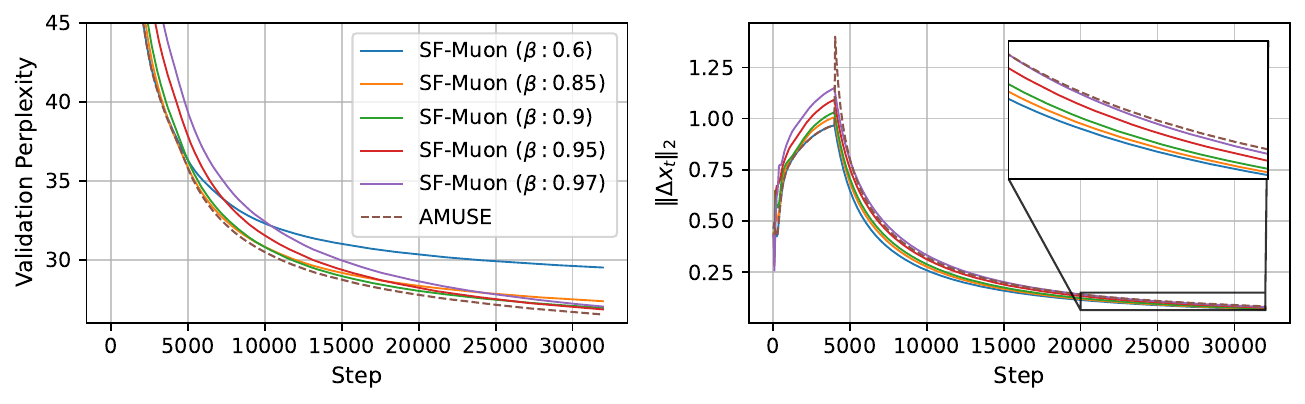}
    \caption{Comparison of fixed-$\beta$ SF-Muon with different $\beta$ values and AMUSE in the 124M Llama pretraining on FineWeb. Solid lines show fixed-$\beta$ SF-Muon, and dashed lines show AMUSE with $\beta_{\rm init}=0.6$ and $\rho=0.8$. We report validation perplexity \textbf{(left)} and the update norm $\|\Delta \vx_t\|$ \textbf{(right)}.}
    \label{fig:sf-muon_varying_beta}
\end{figure}

To verify whether AMUSE behaves as intended, we compare it with fixed-$\beta$ SF-Muon in Figure~\ref{fig:sf-muon_varying_beta}. We train a 124M Llama model on FineWeb, keeping all hyperparameters fixed except the interpolation coefficient: SF-Muon uses different constant $\beta$ values, while AMUSE uses the time-varying $\beta_t$ from Eq.~\eqref{eq:amuse_beta}. 
A small fixed $\beta$ achieves better early performance but degrades late-stage performance, whereas a large fixed $\beta$ improves performance in the later stages at the cost of higher early perplexity. AMUSE balances these effects by gradually increasing $\beta_t$, yielding a Pareto-optimal perplexity curve while maintaining a larger $\|\Delta \vx_t\|$, which indicates faster progress along useful directions. 
Fixed-$\beta$ SF-Muon variants also underperform AMUSE throughout training (Appendix~\ref{app: ablation studies}). We further evaluate the design of the $\beta_t$ schedule in Eq.~\eqref{eq:amuse_beta} by comparing it with tuned linear and cosine alternatives. The proposed schedule achieves lower validation perplexity (Table~\ref{tab:beta_schedule_comparison} in Appendix~\ref{app: ablation studies}).

We also compare AMUSE with post-hoc weight averaging methods, including Stochastic Weight Averaging~\citep{izmailov2018averaging}, Latest Weight Averaging~\citep{kaddour2022stop}, and Pre-trained Model Average~\citep{li2025model}, in Appendix~\ref{app:weight_averaging}. These methods average an existing trajectory, whereas AMUSE incorporates averaged iterates into gradient evaluation, modifying subsequent updates.

\vspace{-4pt}
\subsection{AMUSE Follows the River}
\vspace{-3pt}
\label{subsec:amuse_river_proxy}
Following~\citet{song2025through}, we test whether AMUSE already follows the river using two post-hoc evaluations: Exponential Weight Averaging (EWA) and learning rate decay. If the optimizer still oscillates across the valley walls, applying either to the $\vx_t$ trajectory should substantially reduce the loss by moving it closer to the river; if it already tracks the river, they should offer little benefit.

\begin{figure}[t]
    \centering
    \includegraphics[width=1\linewidth]{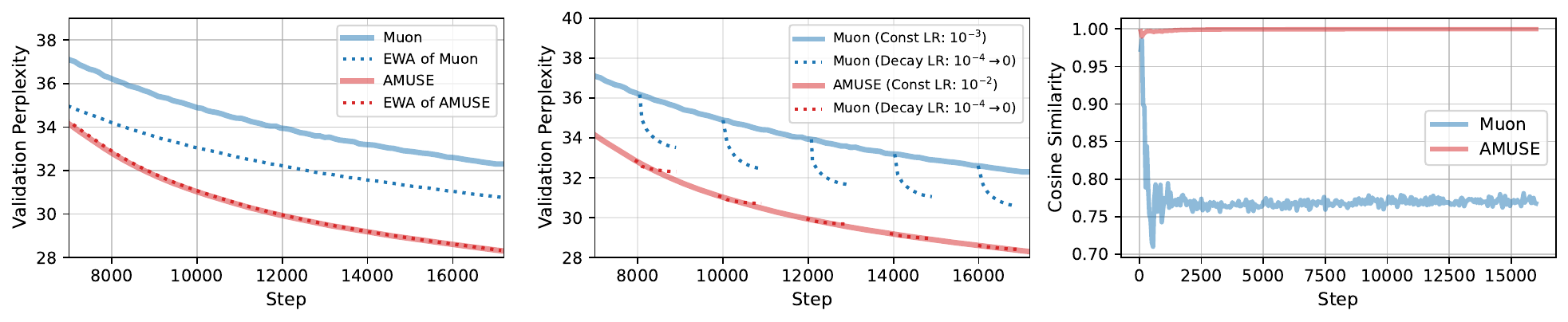}
    \caption{
    Comparison between constant learning rate Muon and AMUSE in the 124M Llama setting.
    \textbf{The left panel} shows the effect of EWA, where solid lines represent the original training trajectories and dotted lines represent their corresponding EWA trajectories. \textbf{The middle panel} shows the effect of learning rate decay, where after warmup we linearly decay the learning rate from $10^{-4}$ to $0$ at selected iterations using the Muon optimizer; solid lines denote the original trajectories, while dotted lines show the corresponding decay-phase trajectories. \textbf{The right panel} shows cosine similarity between consecutive updates, plotting $\cos(\Delta \vx_t, \Delta \vx_{t+1})$ for AMUSE and $\cos(\Delta \theta_t, \Delta \theta_{t+1})$ for Muon.
    }
    \vspace{-3pt}
    \label{fig:AMUSE river}
\end{figure}

As shown in Figure~\ref{fig:AMUSE river}, constant learning rate Muon benefits significantly from both EWA and learning rate decay, indicating strong valley-wall oscillation. 
In contrast, AMUSE shows little improvement from either procedure, suggesting it already stays near the river.
This stable progress is further confirmed by AMUSE's high cosine similarity between consecutive updates of $\vx_t$.

%% file: 05_Experiments.tex
\vspace{-6pt}
\section{Experiments}
\vspace{-4pt}
We evaluate AMUSE across a diverse set of benchmarks covering standard image classification, image segmentation, and large language model (LLM) pretraining. For the standard image domain experiments, we primarily follow the setup of \citet{defazio2024the}. For LLM pretraining, we follow \citet{semenov2026benchmarking} and adopt their Llama-like decoder-only Transformer setup trained on FineWeb-100B~\citep{penedo2024the}. In particular, we consider 124M, 720M, and 1.3B model sizes.

As baselines, we use SGD with momentum and SF-SGD for CIFAR-10, CIFAR-100, SVHN, ImageNet-1k, and ISIC 2018. AdamW and SF-AdamW are used for the remaining tasks. Across all settings, we additionally include Muon as a shared baseline. For SGD, AdamW and Muon, we employ cosine learning rate schedules across all experiments. 

\vspace{-3pt}
\subsection{Image Domain Experiments}
\vspace{-2pt}
\noindent \textbf{Experimental Setup.} 
We evaluate AMUSE on image classification benchmarks on diverse datasets and architectures: Wide-ResNet-16-8 on CIFAR-10~\citep{DBLP:journals/corr/ZagoruykoK16,krizhevsky2009learning}, DenseNet on CIFAR-100~\citep{Huang_2017_CVPR}, ResNet-3-96 on SVHN~\citep{He_2016_CVPR,netzer2011reading}, and ResNet-50 on ImageNet-1k~\citep{russakovsky2015imagenet}. 
We further evaluate two additional settings: U-Net on ISIC 2018 for image segmentation~\citep{ronneberger2015u,codella2019skin}, and MAE fine-tuning with a ViT-B/16 on ImageNet~\citep{he2022masked}. 
For existing baselines, we use the tuned hyperparameters from~\citet{defazio2024the} when available. 
For Muon and AMUSE, whose settings are not provided, we fairly tune the learning rate, weight decay, and optimizer-specific hyperparameters; details are provided in Appendix~\ref{app:img_hyper}.

\noindent \textbf{Experimental Results.} 
As illustrated in Figure~\ref{fig:image_domain}, AMUSE consistently achieves the best performance among the compared optimizers across all evaluated datasets, tasks, and architectures. It demonstrates \textit{superior anytime performance}: while vanilla Muon requires the full training horizon to reach competitive performance, AMUSE exhibits rapid early progress and maintains the highest accuracy throughout training. Notably, AMUSE remains highly effective for MAE fine-tuning even when initialized from an AdamW-pretrained model. In contrast, vanilla Muon struggles in this setting, consistent with recent reports of optimizer mismatch when fine-tuning AdamW-pretrained models with Muon~\citep{qu2026can, liu2025muon, liu2025reg}. This suggests that AMUSE can be applied effectively to models pretrained with conventional optimizers.

\vspace{-3pt}
\subsection{Large Language Model Pretraining}
\vspace{-2pt}
\label{sec:llm_training}
\noindent \textbf{Experimental Setup.}
We follow the Llama-style Transformer setup of~\citet{semenov2026benchmarking}, using tied input/output embeddings, SwiGLU, RMSNorm, and RoPE.
We train 124M, 720M, and 1.3B models on FineWeb-100B~\citep{penedo2024the} with sequence length 512 and batch sizes 256, 1984, and 2048, respectively.
The 124M and 720M models are trained for 16k iterations ($\approx$2.1B and $\approx$16.3B tokens, respectively), while the 1.3B model is trained for 24.8k iterations ($\approx$26B tokens).
For baseline optimizers, we use the highly tuned hyperparameters from~\citet{semenov2026benchmarking}, including learning rate, weight decay, scheduler, warmup steps, and momentum.
At the 124M scale, we additionally conduct an expanded hyperparameter search for Muon, starting from the reported best configuration.
For AMUSE, we tune the corresponding hyperparameters under the same experimental setup, along with $\beta_{\rm init} \in \{0.4, 0.6\}$ and $\rho \in \{0.6, 0.8\}$.
See Appendix~\ref{app:llm_hyper} for more details.

We additionally evaluate the final 720M checkpoints after 16k pretraining iterations on ARC-Easy, ARC-Challenge, BoolQ, HellaSwag, PIQA, WinoGrande, and OpenBookQA in a zero-shot setting.

\input{05_Image_Domain_Figures}

\begin{figure}[t]
    \centering
    \includegraphics[width=\linewidth]{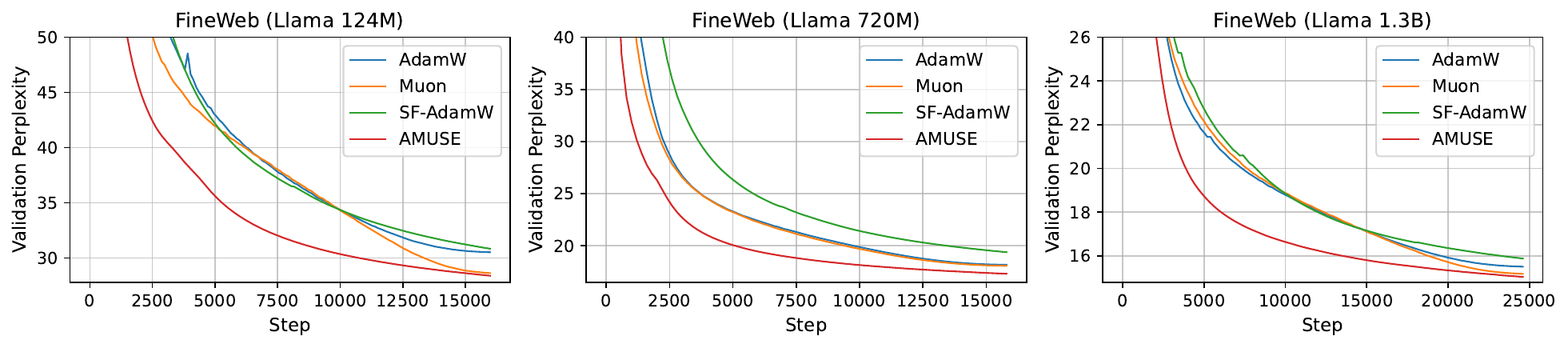}
    \caption{Validation perplexity on FineWeb pretraining across Llama model scales.}
    \vspace{-2pt}
    \label{fig:fineweb_pretraining}
\end{figure}

\noindent \textbf{Experimental Results.}
The results in Figure~\ref{fig:fineweb_pretraining} show that AMUSE consistently improves LLM pretraining. Across all Llama model scales, it achieves lower validation perplexity than all baselines throughout the entire training horizon. Extended training results across 124M and 720M model scales show that AMUSE remains effective beyond the standard training horizon (Figure~\ref{fig:fineweb_longer} in Appendix~\ref{app:training_longer_iteration}). We also compare AMUSE with other recent optimizers, demonstrating that it consistently improves the performance-iteration Pareto frontier (Figure~\ref{fig:all_opts} in Appendix~\ref{app:llm_more_baselines}). While SF-AdamW struggles with the large batch sizes typically required for modern LLMs~\citep{morwani2025connections}, AMUSE scales effectively, maintaining strong performance on the 720M and 1.3B models trained with large batches. AMUSE's advantage persists across five random seeds at 124M; see Table~\ref{tab:llm_multiseed_124m} in Appendix~\ref{app:llm_multiseed}.

As shown in Table~\ref{tab:downstream_720m}, AMUSE achieves the highest accuracy across all seven downstream tasks, with an average accuracy of 46.3\% compared with 45.3\% for Muon. These results indicate that AMUSE's improvements in validation perplexity also translate into improved downstream performance.

We further test whether adding exponential weight averaging or a decay phase to constant learning rate Muon recovers AMUSE's performance. As detailed in Appendix~\ref{app:muon_ewa_wsd}, while these variants improve Muon, they still fall short of AMUSE. This suggests AMUSE inherently maintains a stable trajectory, rather than merely mimicking weight averaging or learning rate decay.
AMUSE retains its performance advantage in wall-clock time, with runtime comparable to Muon (Appendix~\ref{app:wall_clock}).

\begin{table}[t]
    \centering
    \caption{Zero-shot downstream accuracy of 720M models after 16k pretraining iterations.}
    \label{tab:downstream_720m}
    \small
    \setlength{\tabcolsep}{5pt}
    \begin{tabular}{lcccccccc}
        \toprule
        Optimizer & ARC-E & ARC-C & BoolQ & HS & PIQA & WG & OBQA & Avg. \\
        \midrule
        AdamW   & 0.534 & 0.228 & 0.587 & 0.372 & 0.700 & 0.535 & 0.172 & 0.447 \\
        Muon    & 0.542 & 0.235 & 0.567 & 0.376 & 0.717 & 0.536 & 0.198 & 0.453 \\
        SF-Muon & 0.542 & 0.224 & 0.576 & 0.384 & 0.706 & 0.511 & 0.206 & 0.450 \\
        AMUSE   & \textbf{0.547} & \textbf{0.240} & \textbf{0.588}
                & \textbf{0.388} & \textbf{0.720} & \textbf{0.539}
                & \textbf{0.220} & \textbf{0.463} \\
        \bottomrule
    \end{tabular}
\end{table}

\vspace{-4pt}
\subsection{Hyperparameter Sensitivity}
\vspace{-3pt}
In all AMUSE experiments, we fix the Muon momentum $\mu$ to $0.95$ and do not treat it as a tunable parameter. Therefore, compared to standard SF optimizers, AMUSE introduces only one additional hyperparameter, $\rho$, which controls how quickly the gradient evaluation point moves toward the averaged iterate. We evaluate the hyperparameter sensitivity of AMUSE to $\beta_{\rm init}$ and $\rho$ in the Llama 124M setting in Appendix~\ref{app:hyper_sensitivity}, demonstrating its robustness.

\vspace{-4pt}
\subsection{Implementation Details}
\label{sec:Impldetails}
\vspace{-3pt}
For language model pretraining and MAE fine-tuning, we optimize non-Muon parameters, such as embeddings, normalization layers, and output heads, with SF-AdamW; for other image-domain experiments, we use SF-SGD, following standard Muon practice. For Muon updates, we use Nesterov momentum rather than the Polyak momentum formulation in Eqs.~\eqref{eqn: muon} and~\eqref{eq:sfmuon}, following the practical Muon implementation~\citep{jordan2024muon}.

Although AMUSE is formulated with three sequences $(\vx_t, \vy_t, \vz_t)$ from the standard SF formulation and a Muon momentum buffer $\vm_t$, its practical implementation requires storing only the model parameters (kept at $\vy_t$), the sequence state $\vz_t$, and the momentum buffer $\vm_t$. For evaluation, we temporarily compute $\vx_t$ via in-place interpolation using $\vy_t$ and $\vz_t$, and revert the parameters afterward. Because of this design, AMUSE requires exactly one extra state copy ($\vz_t$) compared to vanilla Muon, \textit{incurring no additional memory overhead compared to AdamW and SF-AdamW}.
While AMUSE still utilizes a standard linear warmup phase, $\eta_t=\eta\min(1,t/T_{0})$, it does not require a subsequent learning rate decay phase. 
For other implementation details, see Appendix~\ref{app:implementation_details}.

%% file: 05_Image_Domain_Figures.tex
\begin{figure}[t]
    \centering
    \includegraphics[width=\linewidth]{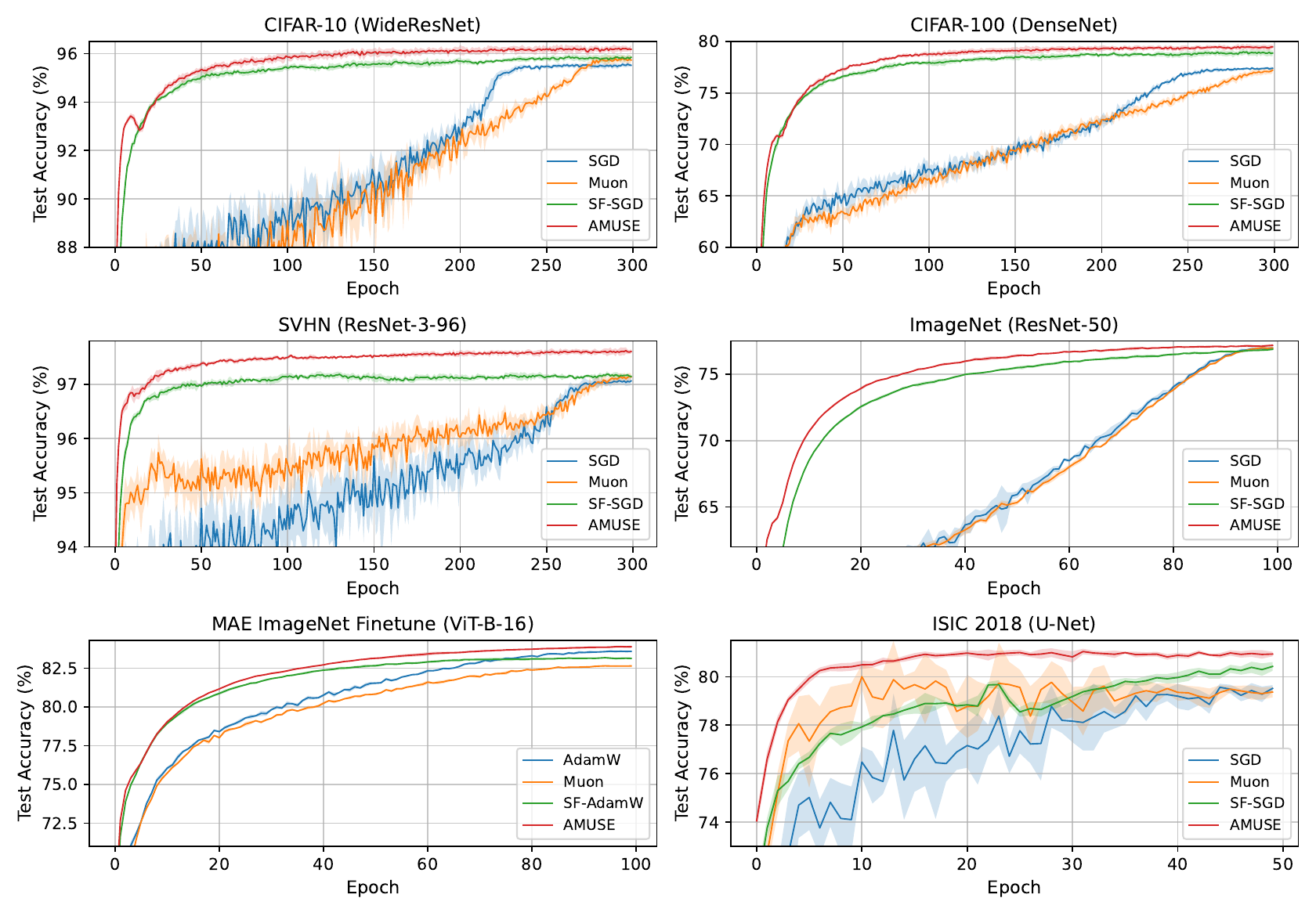}
    \caption[Test accuracy across image-domain experiments.]{
    Test accuracy across image domain experiments. Averaged over five random seeds.}
    \vspace{-7pt}
    \label{fig:image_domain}
\end{figure}

%% file: 06_Conclusion.tex
\vspace{-5pt}
\section{Conclusion}
\label{sec:conclusion}
\vspace{-4pt}
We introduce AMUSE, an optimization algorithm designed to stabilize Muon using the Schedule-Free (SF) framework and a time-varying interpolation parameter $\beta_t$. We observe that Muon's orthogonalization operator promotes training by leveraging large bulk components, but also induces oscillations along valley walls, thereby contributing to its inherent instability. AMUSE adopts the SF framework, a principled approach that controls where the gradient is evaluated and stabilizes training by keeping the optimization trajectory close to the river. With a carefully designed schedule for the interpolation parameter $\beta_t$, AMUSE outperforms conventional learning-rate-scheduled Adam(W) or Muon across a wide range of deep learning tasks. Our results highlight how a systematic understanding of training dynamics and the loss landscape provides a principled approach to optimizer design.

Despite the superior performance of AMUSE compared with various optimizers including Muon, it requires additional memory overhead relative to vanilla Muon. Developing an optimization strategy that enhances updates along the river direction and suppresses valley-wall oscillations without incurring such overhead is a promising direction for future investigation. Further evaluation at larger model scales and over longer training horizons is also needed.

\begin{ack}
This work was supported by an Institute of Information \& Communications Technology Planning \& Evaluation (IITP) grant (No.\ RS-2024-00457882 National AI Research Lab Project, 20\%) funded by the Korean government (MSIT), two National Research Foundation of Korea (NRF) grants (No.\ RS-2023-00211352 Development of Efficient Stochastic Optimization Algorithms via Analyzing Without-replacement Sampling Methods, 40\%; No.\ RS-2026-25612350 Development of an End-to-End Design and Validation Framework for Mathematically Principled Next-Generation AI Architectures and Learning Algorithms, 30\%) funded by the Korean government (MSIT), and a KRAFTON research grant (10\%).
\end{ack}

%% file: A_Quadratic_Example.tex
\section{A Quadratic Example of Bouncing in Matrix-Normalized Updates}

\label{app:quadratic_bouncing}

We compare standard gradient descent (GD) with an idealized matrix-normalized update on a two-dimensional matrix quadratic objective. This example is not intended to model the full momentum dynamics of Muon, but rather isolates the magnitude-removing effect of polar normalization. We use an anisotropic setting to show that matrix normalization removes gradient magnitude information, which can cause bouncing of small nonzero components near the optimum.

Let $\mW\in\mathbb{R}^{2\times 2}$ be the parameter matrix, and let $\mA\in\mathbb{R}^{2\times 2}$ be a symmetric positive definite matrix with an anisotropic spectrum:
\begin{equation*}
    \mA = \mQ \boldsymbol{\Lambda} \mQ^\top, \qquad \boldsymbol{\Lambda} = \begin{pmatrix} \lambda & 0 \\ 0 & 1 \end{pmatrix},
\end{equation*}
where $\mQ$ is an orthogonal matrix, and $\lambda \gg 1$ controls the curvature gap between the first and second principal directions. 
We consider the quadratic objective
\begin{equation*}
    f(\mW)=\frac{1}{2}\operatorname{tr}(\mW^\top \mA \mW),
\end{equation*}
where the gradient is $\nabla f(\mW)=\mA\mW$. To analyze the dynamics, we project the parameters into the eigenbasis of $\mA$ by defining $\mV_t = \mQ^\top \mW_t$. Consider a diagonal iterate in this eigenbasis:
\begin{equation*}
    \mV_t=
    \begin{pmatrix}
        a_t & 0 \\
        0 & b_t
    \end{pmatrix}.
\end{equation*}

Then, $\mW_t = \mQ \mV_t$, and the gradient is:
\begin{equation*}
    \nabla f(\mW_t) = \mA \mW_t = (\mQ \boldsymbol{\Lambda} \mQ^\top) (\mQ \mV_t) = \mQ \boldsymbol{\Lambda} \mV_t = \mQ
    \begin{pmatrix}
        \lambda a_t & 0 \\
        0 & b_t
    \end{pmatrix}.
\end{equation*}

\paragraph{Gradient Descent Dynamics.}
GD with step size $\eta$ gives $\mW_{t+1}^{\mathrm{GD}} = \mW_t-\eta\nabla f(\mW_t)$. Projecting this update into the eigenbasis yields:
\begin{equation*}
    \mV_{t+1}^{\mathrm{GD}} = \mV_t - \eta \boldsymbol{\Lambda} \mV_t.
\end{equation*}
For the first coordinate, we have $a_{t+1}^{\mathrm{GD}} = (1-\eta\lambda)a_t$.

If $0<\eta<1/\lambda$, then $0<1-\eta\lambda<1$, so
\begin{equation*}
    |a_{t+1}^{\mathrm{GD}}|<|a_t|,
    \qquad
    \operatorname{sign}(a_{t+1}^{\mathrm{GD}})
    =
    \operatorname{sign}(a_t).
\end{equation*}

The update magnitude in the first coordinate is $|\Delta a_t^{\mathrm{GD}}| = \eta\lambda |a_t|$,
which goes to zero as $a_t\to 0$.

\paragraph{Matrix-Normalized Dynamics.}

Now consider a matrix-normalized update, which idealizes the normalization effect of Muon. For a full-rank matrix $\mG$, define its polar factor as:
\begin{equation*}
    \operatorname{Polar}(\mG) = \mG(\mG^\top \mG)^{-1/2}.
\end{equation*}

We apply this to the gradient $\mG_t = \nabla f(\mW_t) = \mQ \boldsymbol{\Lambda} \mV_t$. When $a_t\neq 0$ and $b_t\neq 0$, we have:
\begin{equation*}
    \mG_t^\top \mG_t
    =
    \mV_t^\top \boldsymbol{\Lambda}^\top \mQ^\top \mQ \boldsymbol{\Lambda} \mV_t
    =
    \begin{pmatrix}
        (\lambda a_t)^2 & 0 \\
        0 & b_t^2
    \end{pmatrix}.
\end{equation*}

The inverse square root is $\operatorname{diag}(|\lambda a_t|^{-1}, |b_t|^{-1})$. Therefore,
\begin{equation*}
    \operatorname{Polar}(\mG_t)
    =
    \mQ \begin{pmatrix} \lambda a_t & 0 \\ 0 & b_t \end{pmatrix} \begin{pmatrix} |\lambda a_t|^{-1} & 0 \\ 0 & |b_t|^{-1} \end{pmatrix}
    =
    \mQ \begin{pmatrix} \operatorname{sign}(a_t) & 0 \\ 0 & \operatorname{sign}(b_t) \end{pmatrix}.
\end{equation*}

The matrix-normalized update is $\mW_{t+1}^{\mathrm{MN}} = \mW_t-\eta\,\operatorname{Polar}(\nabla f(\mW_t))$. Projecting this back into the eigenbasis ($\mV_{t+1}^{\mathrm{MN}} = \mQ^\top \mW_{t+1}^{\mathrm{MN}}$) gives $a_{t+1}^{\mathrm{MN}} = a_t-\eta\,\operatorname{sign}(a_t)$.

The contrast is therefore clear. GD preserves the magnitude information of the gradient, while the matrix-normalized update removes it through polar decomposition. For every nonzero $a_t$ in this full-rank regime, the matrix-normalized step magnitude remains constant:
\begin{equation*}
    |\Delta a_t^{\mathrm{MN}}|=\eta.
\end{equation*}

If $0<|a_t|<\eta$, then
\begin{equation*}
    a_t a_{t+1}^{\mathrm{MN}}<0,
\end{equation*}
so the first coordinate changes sign in one step. Moreover, if $0<a_t<\eta$, then $a_{t+2}^{\mathrm{MN}}=a_t$,
so the dynamics forms a two-cycle along this coordinate. Thus, unlike GD, the matrix-normalized update does not shrink as the component becomes arbitrarily small but nonzero. A small residual component can therefore receive a constant update and bounce across zero.

%% file: B_Experiments_for_Section_3.tex
\section{Additional Results and Experimental Details for Section~3}
\label{app:experiments_section3}
In this section, we provide additional experimental details and results for the analysis in Section~\ref{sec:sfmuon_analysis}. 
We include experimental setups and additional results that were omitted from the main text due to space constraints.

\subsection{Dominant Updates Can Cause Instability, while Bulk Updates Accelerate Training}
\label{app:scaling_proj}

In this subsection, we empirically show that updates along the dominant directions can cause training instability, while amplifying the bulk component can accelerate Muon training. 
Following~\citet{zhou2025bsfa}, we do this by decomposing each Muon update into its dominant and bulk components and scaling them separately.

We conduct experiments on three classification datasets: MNIST, CIFAR-10, and SST-2. 
Detailed experimental settings, including dataset construction and model architectures, are provided in Section~\ref{subsec:section3_exp_details}; hyperparameters are reported in the corresponding figure captions.

We use the dominant and bulk subspaces defined in Definition~\ref{def:dominant_bulk}. 
That is, for the Hessian \(\nabla^2\gL(\theta_t)\), the dominant subspace \(\gS_k(\theta_t)\) is spanned by the top-\(k\) eigenvectors, and the bulk subspace is its orthogonal complement \(\gS_k^\perp(\theta_t)\). 
Following the observation that the Hessian spectrum of classification models contains a small number of outlier eigenvalues related to the number of classes, we set \(k\) to the number of classes: \(k=10\) for MNIST-5k and CIFAR-10-5k, and \(k=2\) for SST-2-1k.

Let \(\mM_t\) denote the matrix-valued Muon momentum and define the flattened post-orthogonalized update direction as \(\vu_t:=\operatorname{vec}(\gT(\mM_t))\). Using the subspace projection matrices from Definition~\ref{def:subspace_proj}, we decompose $\vu_t$ into dominant and bulk components as
\(\mP_k(\theta_t)\vu_t\) and \(\mP_k^\perp(\theta_t)\vu_t\), respectively.
We then rescale the two components separately:
\[
    \widetilde{\vu}_t
    =
    \alpha \mP_k(\theta_t)\vu_t
    +
    \gamma \mP_k^\perp(\theta_t)\vu_t,
\]
where \(\alpha\) and \(\gamma\) are multiplicative scaling factors for the dominant and bulk components, respectively. 
For example, when \(\alpha=\gamma=1\), this recovers the original Muon update. 
The parameter is updated by
\[
    \theta_{t+1}
    =
    \theta_t - \eta \widetilde{\vu}_t.
\]

\begin{figure}[ht]
    \centering
    \includegraphics[width=1\linewidth]{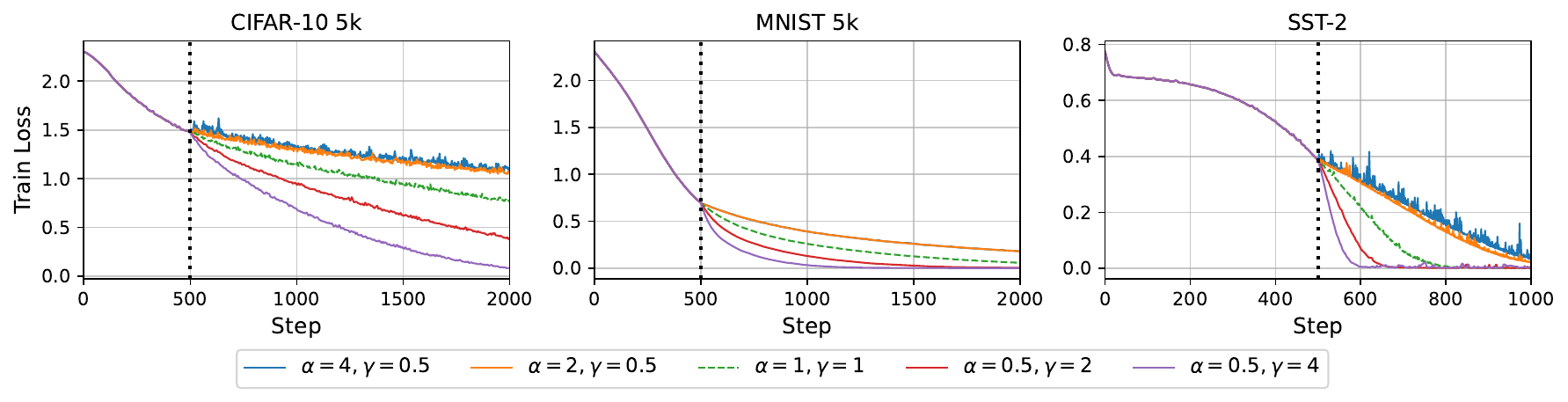}
    \caption{\textbf{Scaling Dominant (\(\alpha\)) and Bulk (\(\gamma\)) components of Muon update.}
    We apply subspace-wise scaling to Muon updates starting at step 500.
    For CIFAR-10-5k and MNIST-5k, Muon is combined with SGD for non-Muon parameters, using momentum \(0.9\) and learning rate \(5\times10^{-4}\).
    For SST-2-1k, Muon is combined with AdamW for non-Muon parameters, using Muon momentum \(0.9\), AdamW coefficients \((0.9,0.99)\), and learning rate \(1\times10^{-4}\).
    }
    \label{fig:proj_trainloss}
\end{figure}

As shown in Figure~\ref{fig:proj_trainloss}, larger \(\alpha\) leads to more unstable training, indicating that dominant-subspace components can induce valley-wall oscillations. 
In contrast, larger \(\gamma\) improves training speed and leads to more stable convergence, suggesting that bulk directions are responsible for useful optimization progress. 
These results suggest that reducing dominant components before orthogonalization is important for training stability, while preserving Muon's strong bulk-oriented updates is important for fast training.

\subsection{Experimental Details}
\label{subsec:section3_exp_details}
We follow the experimental setup of
\citet{song2025does,cohen2021gradient} for all experiments in Appendix~\ref{app:additional experimental results for section 3}.
For experiments in Appendix~\ref{app:scaling_proj}, we use the same experimental setup except for introducing scaling factors following ~\citet{zhou2025bsfa}.

\subsubsection{Architecture Details}

We consider three model architectures, chosen according to the input modality.
For fully connected image classification, we use a three-layer MLP with width 200 and Tanh activations, following~\citet{cohen2021gradient,song2025does}. For convolutional image classification, we use a three-layer CNN with width 32 and ReLU activations, following~\citet{cohen2021gradient,song2025does}. For text classification, we use a two-layer Transformer with hidden dimension 64 and 8 attention heads, following~\citet{damian2022self,song2025does}.

\subsubsection{Datasets}

We use three small-scale classification benchmarks constructed from MNIST, CIFAR-10, and SST-2. For MNIST and CIFAR-10, we use the first 5,000 training examples, each with 10 classes. For SST-2, we use the first 1,000 examples for binary sentiment classification. The main text reports the MNIST-5k experiment with MLP architecture in Section~\ref{sec:sfmuon_analysis}.

\subsubsection{Experimental Setup}

In all experiments in this section, including those in Section~\ref{sec:sfmuon_analysis}, we use cross-entropy loss with batch size 50. We use a constant learning rate for all optimizers: SGD, AdamW, Muon, and SF-Muon. For Muon and SF-Muon, non-hidden matrix layers are optimized with SGD in the MNIST-5k and CIFAR-10-5k experiments, and with AdamW (and its SF variant in SF-Muon) in the SST-2-1k experiment.

The hyperparameters for each experiment are reported in Tables~\ref{tab:baseline_hparams_mnist}--\ref{tab:baseline_hparams_sst2}. 
Unless otherwise specified, each experiment uses the hyperparameters corresponding to its dataset. Specifically, we use the following settings:
\begin{itemize}
    \item Figure~\ref{fig:muon_dom_bulk}, Figure~\ref{fig:sfmuon_grad_components}, and Figure~\ref{fig:eigenspectrum_all}: MNIST hyperparameters in Table~\ref{tab:baseline_hparams_mnist}.
    \item Figure~\ref{fig:eigenspectrum_all}, Figure~\ref{fig:bulk_ratio_cifar}, and Figure~\ref{fig:sfmuon_grad_components_cifar}: CIFAR-10 hyperparameters in Table~\ref{tab:baseline_hparams_cifar}.
    \item Figure~\ref{fig:eigenspectrum_all}, Figure~\ref{fig:bulk_ratio_sst2}, and Figure~\ref{fig:sfmuon_grad_components_sst2}: SST-2 hyperparameters in Table~\ref{tab:baseline_hparams_sst2}.
    \item Figure~\ref{fig:muon_dom_bulk}: AMUSE with learning rate \(0.01\), \(\beta_{\rm init}=0.8\), and \(\rho=0.6\).
    \item Figure~\ref{fig:bulk_ratio_cifar}: AMUSE with learning rate \(0.02\), \(\beta_{\rm init}=0.8\), and \(\rho=0.6\).
    \item Figure~\ref{fig:bulk_ratio_sst2}: AMUSE with learning rate \(0.02\), \(\beta_{\rm init}=0.8\), and \(\rho=0.6\).
\end{itemize}

\begin{table}[!ht]
\centering
\caption{\textbf{Hyperparameters for MNIST-5k Experiments.}}
\label{tab:baseline_hparams_mnist}
\begin{tabular}{lcc}
\toprule
Optimizer & Learning rate & Momentum \\
\midrule
SGD   & 1e-2 & -- \\
AdamW & 5e-4 & (0.9,0.99) \\
Muon  & 1e-3 & 0.9 \\
SF-Muon ($\beta=0.9$)  & 2e-3 & 0.95 \\
\bottomrule
\end{tabular}
\end{table}

\begin{table}[!ht]
\centering
\caption{\textbf{Hyperparameters for CIFAR-10-5k Experiments.}}
\label{tab:baseline_hparams_cifar}
\begin{tabular}{lcc}
\toprule
Optimizer & Learning rate & Momentum \\
\midrule
SGD   & 1e-2 & -- \\
AdamW & 5e-4 & (0.9,0.99) \\
Muon  & 2e-3 & 0.9 \\
SF-Muon ($\beta=0.8$)  & 2e-3 & 0.9 \\
\bottomrule
\end{tabular}
\end{table}

\begin{table}[!ht]
\centering
\caption{\textbf{Hyperparameters for SST-2-1k Experiments.}}
\label{tab:baseline_hparams_sst2}
\begin{tabular}{lccc}
\toprule
Optimizer & Learning rate & AdamW Momentum & Muon momentum \\
\midrule
AdamW & 5e-4 & (0.9, 0.99) & -- \\
Muon  & 2e-3 & (0.9, 0.99) & 0.9 \\
SF-Muon ($\beta=0.8$)  & 1e-4 & (--, 0.99) & 0.9\\
\bottomrule
\end{tabular}
\end{table}

\subsection{Additional Experimental Results}
\label{app:additional experimental results for section 3}

In this section, we provide additional results for the analysis in
Section~\ref{sec:sfmuon_analysis}. We organize the results into three parts: Hessian eigenspectra, Muon dominant-ratio behavior, and SF-Muon dynamics.

\begin{figure}[ht]
    \centering
    \begin{tabular}{@{}c@{}}
        \includegraphics[width=\linewidth]{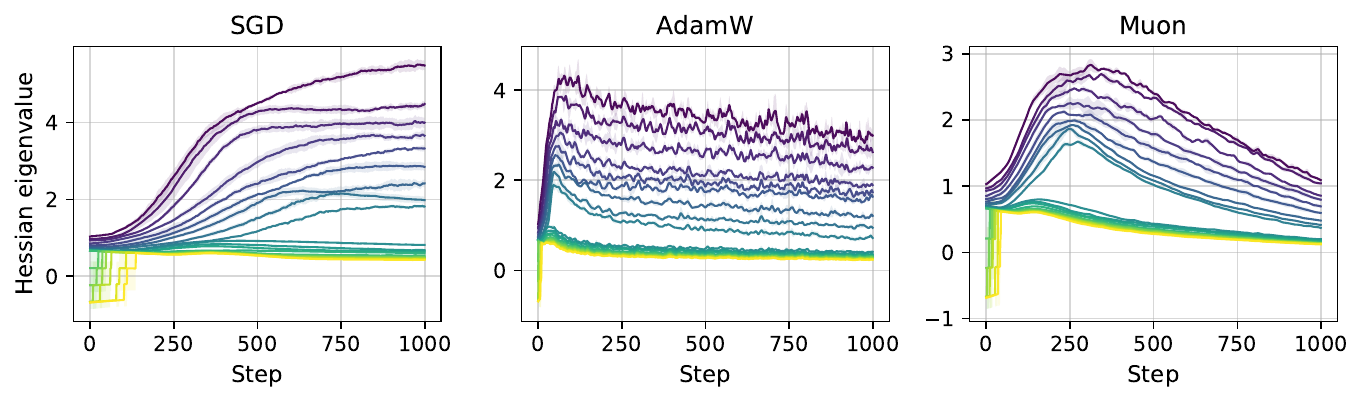} \\
        \small (a) MNIST with MLP \\[0.4em]
        \includegraphics[width=\linewidth]{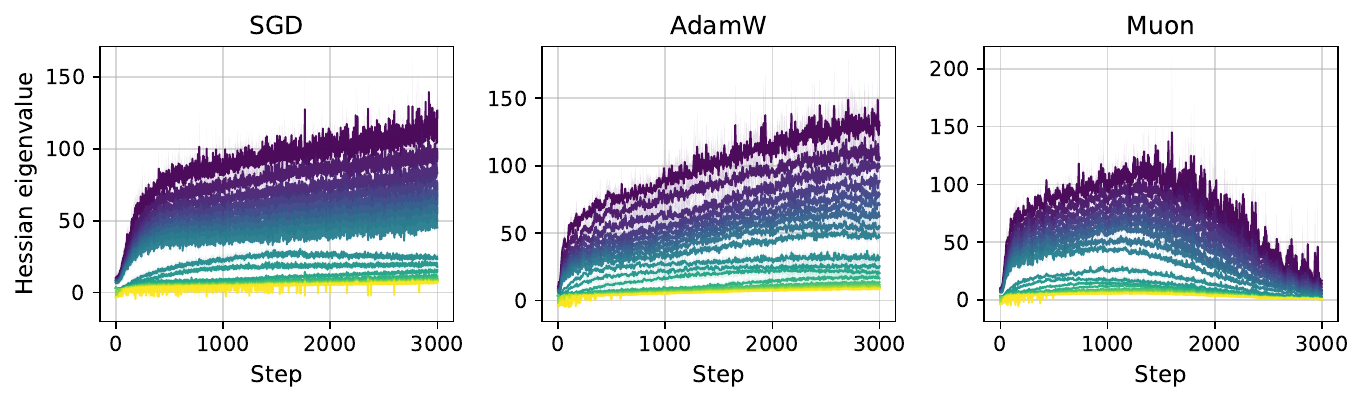} \\
        \small (b) CIFAR-10 with CNN \\[0.4em]
        \includegraphics[width=0.7\linewidth]{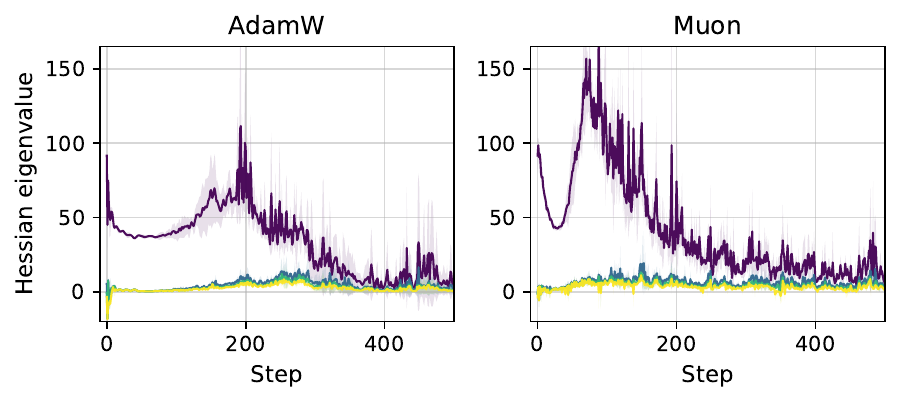} \\
        \small (c) SST-2 with Transformer
    \end{tabular}
    \caption{
    \textbf{Eigenspectra of the loss Hessian across optimizers, datasets, and architectures.} Across all cases, a small number of top-$k$ eigenvalues, where $k$ corresponds to the number of classes, form clear outliers, while the remaining eigenvalues are significantly smaller. This separation highlights a low-dimensional dominant subspace and a high-dimensional bulk subspace.
    }
    \label{fig:eigenspectrum_all}
\end{figure}

\paragraph{Hessian Eigenspectrum.}
We first examine the loss Hessian eigenspectrum across datasets and architectures.
Figure~\ref{fig:eigenspectrum_all} shows that the Hessian spectrum consistently contains
a small number of large outlier eigenvalues, while the remaining eigenvalues are
much smaller. This separation supports the dominant/bulk decomposition used in the
main text.

\begin{figure}[ht]
\centering
\begin{subfigure}[t]{0.49\linewidth}
    \centering
    \includegraphics[width=\linewidth]{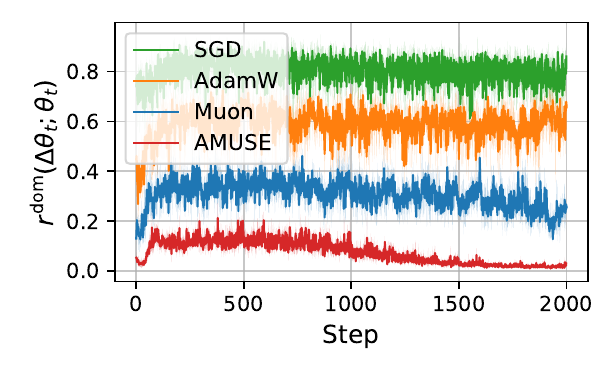}
    \caption{Muon vs. SGD/AdamW}
    \label{fig:Muon bulk vs. SGD/AdamW bulk cifar}
\end{subfigure}
\hfill
\begin{subfigure}[t]{0.49\linewidth}
    \centering
    \includegraphics[width=\linewidth]{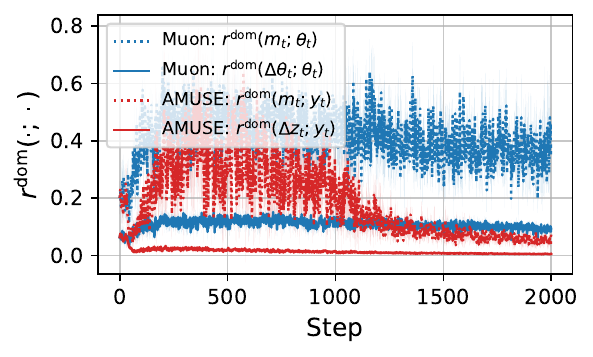}
    \caption{Momentum $\vm_t$ vs. Update $\Delta \theta_t(=\gT(\vm_t))$}
    \label{fig:Muon pre bulk vs. post bulk cifar}
\end{subfigure}
\caption{\textbf{Comparison of dominant component ratios on CIFAR-10 with CNN.} Metrics are measured on a CIFAR-10 5k subset using a CNN architecture. (\textbf{a}) Muon consistently produces substantially smaller dominant component updates than SGD and AdamW, and AMUSE further suppresses the dominant component. (\textbf{b}) Orthogonalization reduces Muon's dominant ratio compared to momentum $\vm_t$; in contrast, AMUSE maintains low dominant ratios throughout, reflecting more stable gradient dynamics. Results are averaged over three runs.}
\label{fig:bulk_ratio_cifar}
\end{figure}

\begin{figure}[ht]
\centering
\begin{subfigure}[t]{0.49\linewidth}
    \centering
    \includegraphics[width=\linewidth]{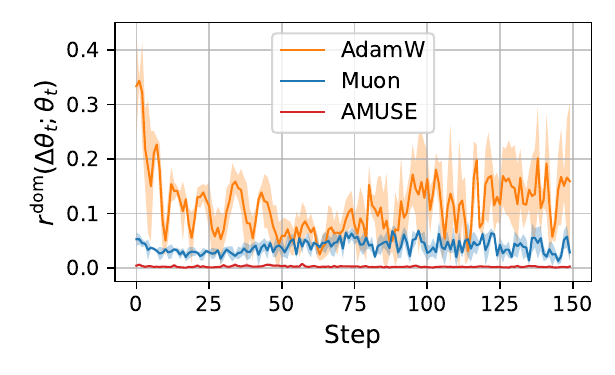}
    \caption{Muon vs. AdamW}
    \label{fig:Muon bulk vs. AdamW dominant sst2}
\end{subfigure}
\hfill
\begin{subfigure}[t]{0.49\linewidth}
    \centering
    \includegraphics[width=\linewidth]{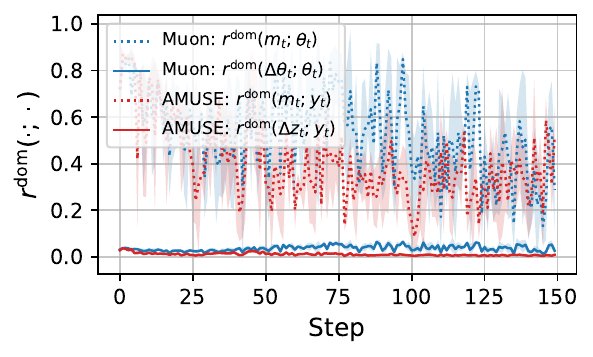}
    \caption{Momentum $\vm_t$ vs. Update $\Delta \theta_t(=\gT(\vm_t))$}
    \label{fig:Muon pre dom vs. post dom sst2}
\end{subfigure}
\caption{\textbf{Comparison of dominant component ratios on SST-2 with Transformer.} Metrics are measured on SST-2 using a Transformer architecture. (\textbf{a}) Muon consistently produces substantially smaller dominant component updates than AdamW, and AMUSE further suppresses the dominant component. (\textbf{b}) Orthogonalization reduces Muon's dominant ratio compared to momentum $\vm_t$; in contrast, AMUSE maintains low dominant ratios throughout, reflecting more stable gradient dynamics. Results are averaged over three runs.}
\label{fig:bulk_ratio_sst2}
\end{figure}

\paragraph{Bulk-oriented Updates of Muon.}
We next verify that the bulk-oriented behavior of Muon extends beyond the main setting. Figures~\ref{fig:bulk_ratio_cifar} and~\ref{fig:bulk_ratio_sst2} show that Muon produces substantially smaller dominant-ratio updates than SGD or AdamW across CIFAR-10 with a CNN and SST-2 with a Transformer. The same figures also compare the pre- and post-orthogonalized Muon updates, showing that the orthogonalization step itself decreases the dominant component of the update. These results reinforce the explanation that Muon accelerates training by promoting movement in bulk directions, as mentioned in Section~\ref{subsec:muon_bulk}.

\begin{figure}[ht]
\begin{subfigure}[ht]{0.49\linewidth}
    \centering
    \includegraphics[width=\linewidth]{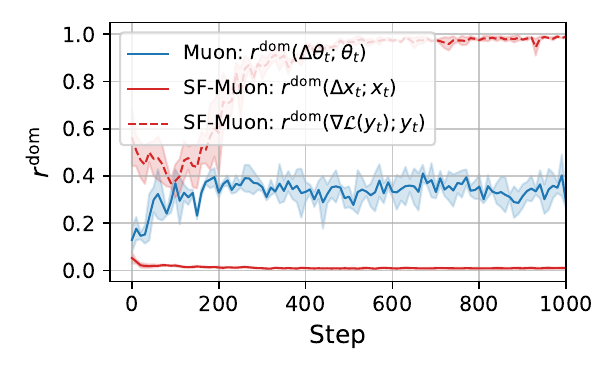}
    \caption{Comparison of dominant ratios}
\end{subfigure}
\hfill
\begin{subfigure}[ht]{0.49\linewidth}
    \centering
    \includegraphics[width=\linewidth]{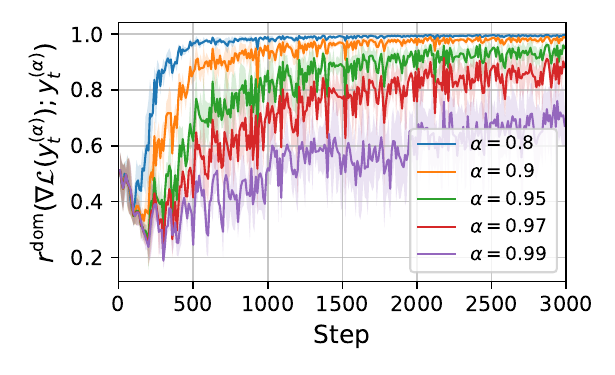}
    \caption{Dominant ratio of gradient at $\vy_t^{(\alpha)}$}
\end{subfigure}

\caption{\textbf{Comparison of dominant component ratios on CIFAR-10 with CNN.} (\textbf{a}) Dominant ratios for Muon (blue) and SF-Muon (red). The solid and dashed red lines denote the dominant components of $\Delta \vx_t$ and $\nabla \gL(\vy_t)$, respectively. (\textbf{b}) Dominant ratios for gradients at $\vy_t^{(\alpha)}=(1-\alpha)\vz_t + \alpha \vx_t$ across varying $\alpha$, demonstrating that larger $\alpha$ values correlate with lower dominant components.}
\label{fig:sfmuon_grad_components_cifar}

\end{figure}

\begin{figure}[ht]
\begin{subfigure}[ht]{0.49\linewidth}
    \centering
    \includegraphics[width=\linewidth]{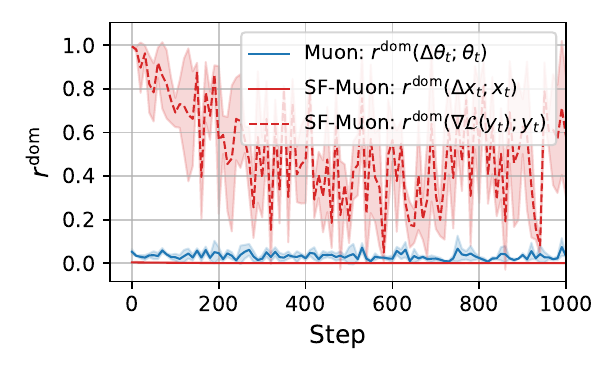}
    \caption{Comparison of dominant ratios}
\end{subfigure}
\hfill
\begin{subfigure}[ht]{0.49\linewidth}
    \centering
    \includegraphics[width=\linewidth]{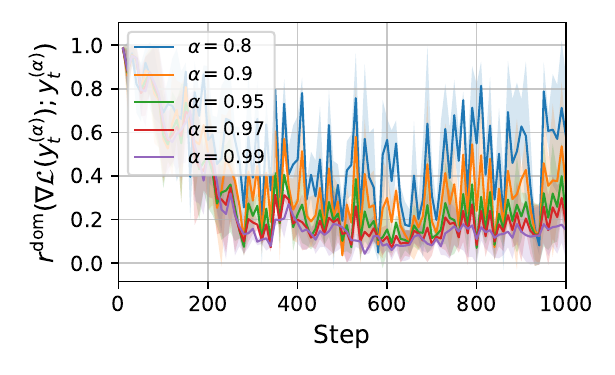}
    \caption{Dominant ratio of gradient at $\vy_t^{(\alpha)}$}
\end{subfigure}

\caption{\textbf{Comparison of dominant component ratios on SST-2 with Transformer.} (\textbf{a}) Dominant ratios for Muon (blue) and SF-Muon (red). The solid and dashed red lines denote the dominant components of $\Delta \vx_t$ and $\nabla \gL(\vy_t)$, respectively. (\textbf{b}) Dominant ratios for gradients at $\vy_t^{(\alpha)}=(1-\alpha)\vz_t + \alpha \vx_t$ across varying $\alpha$, demonstrating that larger $\alpha$ values correlate with lower dominant components.}
\label{fig:sfmuon_grad_components_sst2}

\end{figure}

\paragraph{Schedule-Free Muon Dynamics.}
Finally, we repeat the SF-Muon analysis from the main text (Figure~\ref{fig:sfmuon_grad_components}) on additional
settings. Figures~\ref{fig:sfmuon_grad_components_cifar} and \ref{fig:sfmuon_grad_components_sst2} compare the dominant components of Muon and SF-Muon, and also gradients at multiple virtual interpolation points $\vy_t^{(\alpha)}=(1-\alpha)\vz_t+\alpha\vx_t$. In both settings, the averaged SF-Muon trajectory has a much smaller dominant component than vanilla Muon, and gradients evaluated closer to $\vx_t$ become more bulk-oriented. These results further support the claim that evaluating gradients near the averaged iterate can suppress valley-wall contributions before the Muon orthogonalization step, as mentioned in Section~\ref{subsec:sfmuon}.

\clearpage

\subsection{Extending the Dominant--Bulk Analysis to LLM Pretraining}
\label{app:llm_dominant_ratio}

We further extend dominant--bulk analysis in Section~\ref{sec:sfmuon_analysis}
to 124M Llama pretraining setup.
At each checkpoint, we compute the gradient and Hessian
on the same mini-batch of 256 sequences used for the
parameter update.
We construct the dominant subspace using the top 30
Hessian eigenvectors and measure the dominant ratio
following Definition~\ref{def:dominant_bulk}.

\paragraph{Dominant Ratios of Muon Gradients and Updates.}
We first compare the dominant ratios of the gradient
$\nabla L(\theta_t)$ and the full parameter update
$\Delta_t^{\mathrm{full}} = \theta_{t+1}-\theta_t$
along a Muon training trajectory.
As shown in Table~\ref{tab:llm_muon_orthogonalization},
the gradient exhibits a high dominant ratio.
In contrast, the full optimizer update maintains a
dominant ratio of approximately 0.01 throughout training.
This is consistent with our small-scale observations
that Muon orthogonalization promotes bulk-oriented updates.

\begin{table*}[!ht]
\centering
\small
\caption{Dominant component ratios of the gradient and
full optimizer update along a Muon trajectory during
124M Llama pretraining. The dominant subspace is
constructed using the top 30 Hessian eigenvectors.}
\label{tab:llm_muon_orthogonalization}
\begin{tabular}{lccccccccc}
\toprule
Method & 0K & 2K & 4K & 6K & 8K & 10K & 12K & 14K & 16K \\
\midrule
Muon $\nabla L(\theta_t)$
& 0.43 & 0.81 & 0.76 & 0.78 & 0.74 & 0.76 & 0.84 & 0.73 & 0.74 \\
Muon $\Delta_t^{\mathrm{full}}$
& 0.01 & 0.01 & 0.01 & 0.00 & 0.01 & 0.01 & 0.01 & 0.01 & 0.01 \\
\bottomrule
\end{tabular}
\end{table*}

\paragraph{Gradient-Evaluation Point in AMUSE.}
We next compare the dominant ratios of gradients along
a separately trained Muon trajectory and at the fast
iterate $z_t$ and gradient-evaluation point $y_t$
along an AMUSE trajectory.
As shown in Table~\ref{tab:llm_amuse_gradient},
the gradients along the Muon trajectory and at $z_t$
exhibit relatively high dominant ratios.
In contrast, the gradients at $y_t$ consistently
exhibit lower dominant ratios than those at $z_t$
after initialization, decreasing overall from 0.79
at 2K steps to 0.31 at 16K steps.
These results suggest that moving the gradient-evaluation
point toward the averaged iterate $x_t$ suppresses
the dominant gradient component before Muon orthogonalization.

\begin{table*}[!ht]
\centering
\small
\caption{Dominant component ratios of gradients along
Muon and AMUSE trajectories during 124M Llama pretraining.
For AMUSE, gradients are evaluated at $z_t$ and $y_t$.}
\label{tab:llm_amuse_gradient}
\begin{tabular}{lccccccccc}
\toprule
Method & 0K & 2K & 4K & 6K & 8K & 10K & 12K & 14K & 16K \\
\midrule
Muon $\nabla L(\theta_t)$
& 0.43 & 0.81 & 0.76 & 0.78 & 0.74 & 0.76 & 0.84 & 0.73 & 0.74 \\
AMUSE $\nabla L(z_t)$
& 0.43 & 0.85 & 0.93 & 0.91 & 0.94 & 0.89 & 0.69 & 0.66 & 0.69 \\
AMUSE $\nabla L(y_t)$
& 0.43 & \textbf{0.79} & \textbf{0.70} & \textbf{0.52} & \textbf{0.39} & \textbf{0.41} & \textbf{0.47} & \textbf{0.37} & \textbf{0.31} \\
\bottomrule
\end{tabular}
\end{table*}

Together, these results extend our small-scale observations
to LLM pretraining: Muon produces updates concentrated
in the bulk subspace, while AMUSE further reduces the
dominant gradient component through its time-varying
gradient-evaluation point.

%% file: C_AMUSE.tex
\section{Details for AMUSE}

\subsection{Algorithmic Description of AMUSE}
\label{app:algorithm}

In this section, we present the full procedure of AMUSE, summarized in Algorithm~\ref{alg:amuse}.

\begin{algorithm}[H]
\caption{AMUSE: Anytime Muon with Stable Gradient Evaluation}
\begin{algorithmic}[1]
\setlength{\itemsep}{2pt}
\Require learning rate $\eta$, weight decay $\lambda$, AMUSE $\beta_{\mathrm{init}}$, AMUSE $\rho$, Muon momentum $\mu$, AdamW second-moment coefficient $\beta_2$, numerical constant $\epsilon$, Muon scaling factor $s_{\mathrm{NS}}$, warmup step $T_0$
\State Initialize $\mX_1 \gets \mZ_1$, $\mM_0 \gets \mathbf{0}$, $\mV_0 \gets \mathbf{0}$

\For{$t=1,2,\dots,T$}
    \State $\eta_t \gets \eta \cdot \min(1, t/T_{0})$
    \State $c_{t+1} \gets \frac{\eta_t^2}{\sum_{i=1}^{t}\eta_i^2}$
    \If{$t = T_0$}
        \State $c_{\mathrm{warm}} \gets c_{t+1}$
    \EndIf
    \If{$t \le T_{0}$}
        \State $\beta_t \gets \beta_{\mathrm{init}}$
    \Else
        \State $\beta_t \gets 1 - \left(\frac{c_{t+1} (1-c_{\mathrm{warm}})}{c_{\mathrm{warm}}(1-c_{t+1})}\right)^\rho (1-\beta_{\mathrm{init}})$
        \Comment{Eq.~\eqref{eq:amuse_beta}}
    \EndIf

    \State $\mY_t \gets (1-\beta_t)\mZ_t + \beta_t \mX_t$ \Comment{Gradient evaluation point}
    \State $\mG_t \gets \nabla \gL(\mY_t)$

    \If{the parameter group uses Muon}
        \State $\mM_t \gets \mu\mM_{t-1} + (1-\mu)\mG_t$
        \State $\widehat{\mM}_t \gets \mu\mM_t + (1-\mu)\mG_t$ \Comment{Nesterov momentum}
        \State $\mO_t \gets s_{\mathrm{NS}} \cdot \mathrm{Newton\text{-}Schulz}(\widehat{\mM}_t)$
        \State $\mZ_{t+1} \gets (1-\eta_t\lambda)\mZ_t - \eta_t\mO_t$
    \ElsIf{the parameter group uses AdamW}
        \State $\mV_t \gets \beta_2\mV_{t-1} + (1-\beta_2)(\mG_t \odot \mG_t)$
        \State $\widehat{\mV}_t \gets \mV_t / (1-\beta_2^t)$
        \State $\mU_t \gets \mG_t \oslash \left(\sqrt{\widehat{\mV}_t} + \epsilon\right)$
        \State $\mZ_{t+1} \gets (1-\eta_t \lambda)\mZ_t - \eta_t\mU_t$
    \ElsIf{the parameter group uses SGD}
        \State $\mZ_{t+1} \gets (1-\eta_t\lambda)\mZ_t - \eta_t \mG_t$
    \EndIf
    \State $\mX_{t+1} \gets (1-c_{t+1})\mX_t + c_{t+1}\mZ_{t+1}$ \State $\mX_{t+1} \gets (1-c_{t+1})\mX_t + c_{t+1}\mZ_{t+1}$ \Comment{Update averaged iterate}
\EndFor

\State \Return $\mX_{T+1}$

\end{algorithmic}
\label{alg:amuse}
\end{algorithm}

\newpage
\subsection{Muon Momentum and Averaging After Orthogonalization}
\label{app:amuse_delta_x}
In this section, we justify why AMUSE retains the Muon momentum buffer $\mM_t$. \citet{defazio2024the} observe that Schedule-Free SGD and AdamW do not require an explicit first-moment momentum buffer, since the displacement of the averaged sequence itself induces a momentum-like update. In particular, for uniform averaging and constant $\beta$, Theorem 4 in \citet{defazio2024the} gives
\begin{align*}
    \Delta \mX_t &= \frac{t-1}{t+1}\Delta \mX_{t-1} + \frac{1}{t+1}\Delta \mZ_t, \\
    \Delta \mY_t &= \beta \Delta \mX_t + (1-\beta)\Delta \mZ_t,
\end{align*}
where $\Delta \mX_t := \mX_{t+1}-\mX_t$, $\Delta \mY_t := \mY_{t+1}-\mY_t$, and $\Delta \mZ_t := \mZ_{t+1}-\mZ_t$. Thus, $\Delta \mX_t$ acts as an implicit momentum term that stabilizes the movement of the gradient-evaluation sequence $\mY_t$.

For AMUSE, however, this implicit Schedule-Free momentum does not replace the Muon momentum buffer. Muon's momentum $\mM_t$ serves a distinct role: it accumulates recent gradients before the orthogonalization step, thereby determining the direction to which the Muon update is applied. Therefore, AMUSE retains $\mM_t$ while using the Schedule-Free interpolation to stabilize the point at which gradients are evaluated.

We derive the averaged-sequence update using the Nesterov direction $\widehat{\mM}_t := \mu\mM_t+(1-\mu)\nabla\gL(\mY_t)$ from Algorithm~\ref{alg:amuse}:
\begin{equation*}
    \mZ_{t+1} = \mZ_t - \eta \gT(\widehat{\mM}_t),
    \qquad
    \mX_{t+1} = \left(1-c_{t+1}\right)\mX_t + c_{t+1}\mZ_{t+1},
\end{equation*}
with initialization $\mZ_1=\mX_1$.
Since $c_{t+1}=1/(t+1)$, $\mX_t$ is the uniform average of the $\mZ_t$ iterates, i.e., $\mX_t = \frac{1}{t}\sum_{i=1}^{t}\mZ_i$.

Each base iterate can be written as $\mZ_i = \mZ_1 - \eta \sum_{j=1}^{i-1}\gT(\widehat{\mM}_j)$. Substituting this expression into the uniform average gives
\begin{align*}
    \mX_t
    &= \frac{1}{t} \sum_{i=1}^{t} \left(\mZ_1 - \eta \sum_{j=1}^{i-1}\gT(\widehat{\mM}_j)\right) \\
    &= \mZ_1 - \frac{\eta}{t} \sum_{i=1}^{t} \sum_{j=1}^{i-1} \gT(\widehat{\mM}_j) \\
    &= \mZ_1 - \frac{\eta}{t} \sum_{j=1}^{t-1}(t-j)\gT(\widehat{\mM}_j).
\end{align*}

We now derive the averaged-sequence update $\Delta \mX_t$ using the expression above:
\begin{align*}
    \Delta \mX_t &= -\frac{\eta}{t+1} \sum_{j=1}^{t} (t+1-j)\gT(\widehat{\mM}_j) + \frac{\eta}{t}\sum_{j=1}^{t-1}(t-j)\gT(\widehat{\mM}_j).
\end{align*}
For $j=1,\dots,t-1$, the coefficient of $\gT(\widehat{\mM}_j)$ is
\begin{align*}
    \eta \left(\frac{t-j}{t} - \frac{t+1-j}{t+1}\right) &= -\frac{\eta j}{t(t+1)}.
\end{align*}
For $j=t$, only the first sum contributes, giving $-\frac{\eta}{t+1} = -\frac{\eta t}{t(t+1)}$.
Combining both cases, we obtain
\begin{equation*}
    \Delta \mX_t= -\frac{\eta}{t(t+1)} \sum_{j=1}^{t} j\gT(\widehat{\mM}_j).
\end{equation*}

Therefore, the averaged sequence in AMUSE averages a trajectory after orthogonalized Muon updates have already been applied, and $\Delta \mX_t$ becomes a weighted average of previous orthogonalized updates. This is distinct from Muon momentum: the momentum buffer $\mM_t$ smooths gradients before orthogonalization, whereas the averaged sequence $\mX_t$ averages the trajectory induced by the already orthogonalized updates. Empirically, we also find that removing the Muon momentum buffer substantially degrades performance compared to the full AMUSE and SF-Muon update, as shown in Figure~\ref{fig:fineweb_muon_wo_momentum} in Appendix~\ref{app: ablation studies}.

\subsection{Derivation of the \texorpdfstring{$\beta_t$}{beta\_t} Schedule}
\label{app:amuse_design}

In designing the schedule for $\beta_t$ in AMUSE, one might initially consider a monotonically increasing schedule toward 1. However, following the derivation by~\citet{song2025through}, a closer inspection of the schedule-free (SF) update reveals that the $\vx_t$ iterates perform an implicit averaging of the $\vy_t$ iterates. To see this, first consider the fixed-$\beta$ case. Recall the updates from Eq.~\eqref{eq:sf}:
\begin{align*}
\vy_{t} &= (1-\beta)\vz_{t}+\beta\vx_{t}, \\
\vz_{t+1} &= \vz_{t}-\eta \Delta_t, \\
\vx_{t+1} &= \left(1-c_{t+1}\right)\vx_{t}+c_{t+1}\vz_{t+1}.
\end{align*}
From the definition of $\vy_{t+1}$, we have
\begin{equation*}
\vz_{t+1} = \frac{\vy_{t+1}-\beta\vx_{t+1}}{1-\beta}.
\end{equation*}
Substituting this expression into the update of $\vx_{t+1}$ gives
\begin{align*}
\vx_{t+1} &= (1-c_{t+1})\vx_t + c_{t+1}\frac{\vy_{t+1}-\beta\vx_{t+1}}{1-\beta}.
\end{align*}
Rearranging the terms involving $\vx_{t+1}$ yields
\begin{align*}
\left((1-\beta)+\beta c_{t+1}\right)\vx_{t+1} &= (1-c_{t+1})(1-\beta)\vx_t + c_{t+1}\vy_{t+1}.
\end{align*}
Equivalently,
\begin{align*}
\vx_{t+1} &= \frac{(1-c_{t+1})(1-\beta)}{(1-c_{t+1})(1-\beta)+c_{t+1}}\vx_t + \frac{c_{t+1}}{(1-c_{t+1})(1-\beta)+c_{t+1}}\vy_{t+1}.
\end{align*}
Defining
\begin{equation*}
\omega_{t+1} := \frac{c_{t+1}}{(1-c_{t+1})(1-\beta)+c_{t+1}},
\end{equation*}
we obtain the recursion
\begin{equation*}
\vx_{t+1} = (1-\omega_{t+1})\vx_t + \omega_{t+1}\vy_{t+1}.
\end{equation*}
Thus, $\vx_{t+1}$ is an implicit weighted average of the evaluation points $\vy_s$, with time-varying averaging coefficient $\omega_{t+1}$. If $c_{t+1}=1/(t+1)$, the effective weight $\omega_{t+1}$ simplifies to
\begin{equation*}
\omega_{t+1} = \frac{1}{t(1-\beta)+1}.
\end{equation*}
This expression suggests that, when designing a time-varying schedule, replacing $\beta$ by $\beta_t$ makes the effective averaging coefficient controlled by $t(1-\beta_t)$. If $\beta_t$ increases too rapidly toward 1, the term $1-\beta_t$ vanishes, causing $t(1-\beta_t)$ to decrease and thereby increasing the effective weight. An increasing effective weight implies that the implicit averaging window actually \textit{shrinks}. From an optimization perspective, shrinking the averaging window at later stages of training is undesirable, since it reduces the stabilizing effect of the averaged sequence. Therefore, the schedule for $\beta_t$ should be chosen so that $t(1-\beta_t)$ does not decay too rapidly, preventing the implicit averaging window from collapsing.

Based on this observation, we define the upper boundary by considering the fastest schedule that still maintains the implicit averaging window after a given step $T_0$. Using the $\beta_t$ notation, the effective averaging coefficient is controlled by
\begin{equation*}
    \widetilde{\omega}_{t+1} := \frac{c_{t+1}}{(1-c_{t+1})(1-\beta_t)+c_{t+1}}.
\end{equation*}
Since the window size is inversely related to $\widetilde{\omega}_{t+1}$, keeping the window size fixed amounts to keeping $\widetilde{\omega}_{t+1}$ constant. Setting $\widetilde{\omega}_{t+1}=\widetilde{\omega}_{T_0+1}$ for $t \ge T_0$, we obtain
\begin{align*}
    \widetilde{\omega}_{t+1} &= \frac{c_{t+1}}{(1-c_{t+1})(1-\beta_t)+c_{t+1}} \\
    &= \frac{c_{T_0+1}}{(1-c_{T_0+1})(1-\beta_{T_0})+c_{T_0+1}} = \widetilde{\omega}_{T_0+1}.
\end{align*}
Solving this equality for $\beta_t$ gives the constant-window boundary
\begin{equation*}
    \beta_t = 1 - \frac{c_{t+1}(1-c_{T_0+1})}{c_{T_0+1}(1-c_{t+1})}(1-\beta_{T_0}).
\end{equation*}

Building on this boundary condition, we introduce an interpolation parameter $\rho$ to control how quickly $\beta_t$ approaches this boundary. Taking $\beta_{T_0}=\beta_{\rm init}$, we define, for $t \ge T_0$,
\begin{equation*}
    \beta_t = 1 - \left(\frac{c_{t+1}(1-c_{T_0+1})}{c_{T_0+1}(1-c_{t+1})}\right)^\rho(1-\beta_{\rm init}).
\end{equation*}
This schedule interpolates between the standard fixed-$\beta$ schedule and the constant-window boundary: when $\rho=0$, we recover $\beta_t=\beta_{\rm init}$, while when $\rho=1$, we recover the boundary that keeps $\widetilde{\omega}_t$ constant. By restricting $\rho \in [0,1]$, we ensure that the implicit averaging window is allowed to grow for $0 \le \rho < 1$ and remains constant at the boundary when $\rho=1$. Thus, the schedule prevents the averaging window from shrinking while still allowing $\beta_t$ to increase over training.

\begin{figure}
    \centering
    \includegraphics[width=1\linewidth]{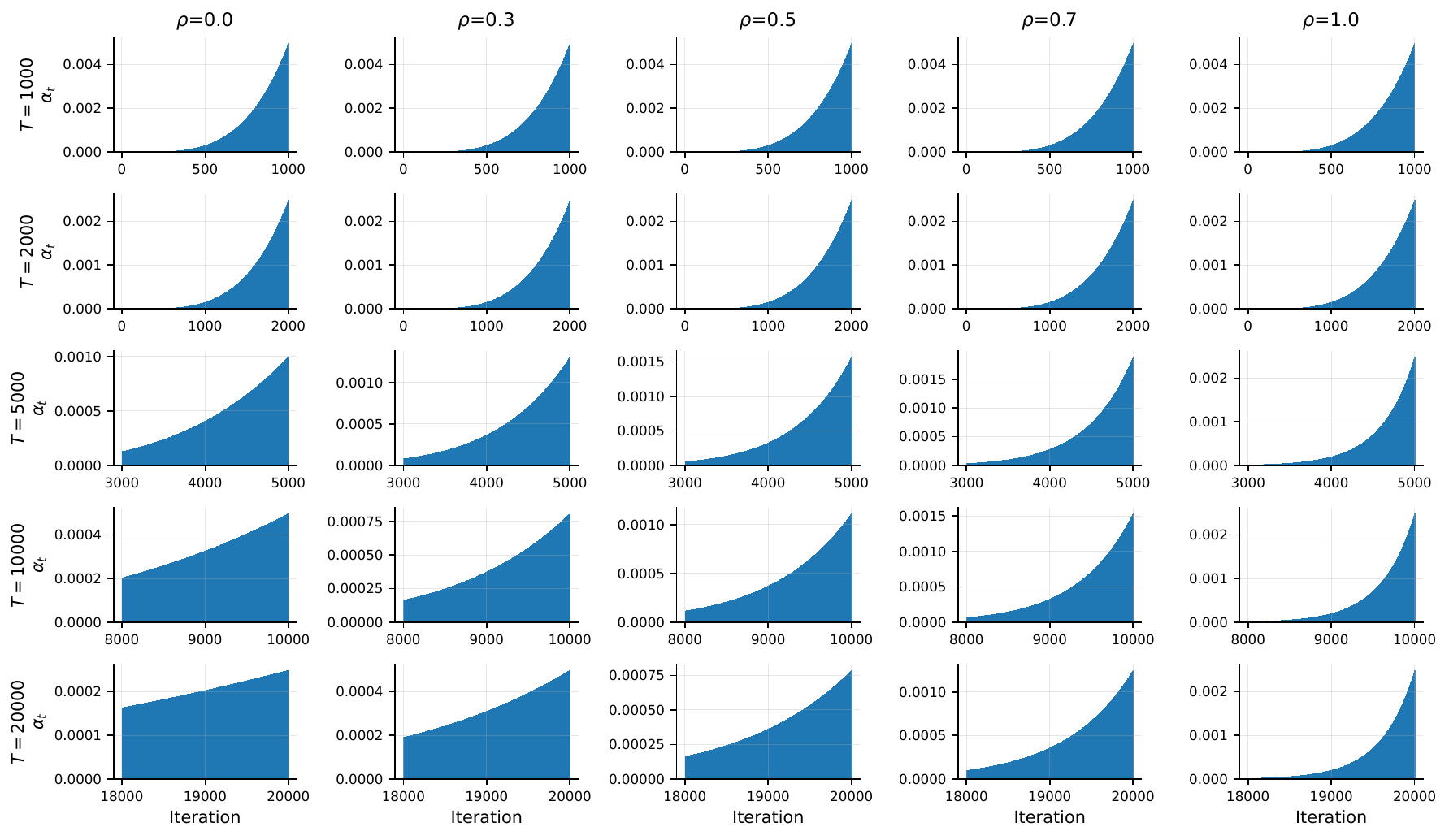}
    \caption{\textbf{Effect of $\rho$ on effective weights $\alpha_t$.} Histograms of $\alpha_t$ across varying sequence lengths $T$. For $T > T_0=2000$, lower $\rho$ values allow the effective averaging window to continuously grow, whereas $\rho = 1$ strictly fixes the window to its size at $T_0$.}
\label{fig:alpha_histograms}
    \label{fig:alpha_hist_beta0.8}
\end{figure}

\begin{figure}
    \centering
    \includegraphics[width=1\linewidth]{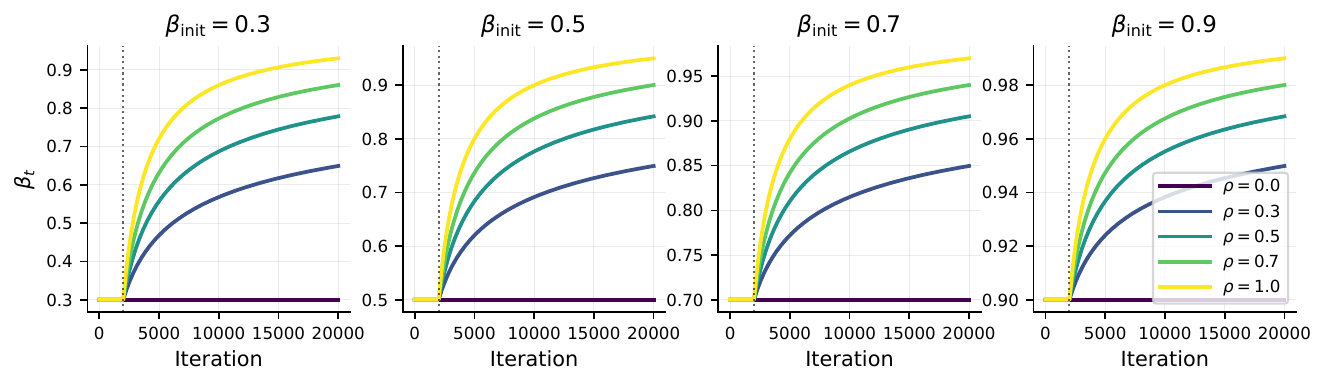}
    \caption{\textbf{Evolution of $\beta_t$ controlled by $\rho$.} After the initial warmup phase ($T_0=2000$, dotted vertical line), $\rho$ smoothly interpolates between a constant $\beta$ baseline ($\rho=0$) and the strict constant-window upper bound ($\rho=1$).}
\label{fig:beta_evolution}
\end{figure}

\paragraph{Visualization of Averaging Window Size and $\beta_t$.} To better understand the impact of our proposed schedule on the averaging dynamics, we visualize both the effective averaging window and the trajectory of $\beta_t$. By unrolling the recursive schedule-free update, the final averaged parameter $\vx_T$ can be explicitly expressed as a weighted sum of all past iterates $\vy_t$, i.e., $\vx_T=\sum_{t=1}^T \alpha_t \vy_t,$ where the global effective weight $\alpha_t$ assigned to each step $t$ is defined as:
$$
    \alpha_t \triangleq \frac{c_t}{(1-c_t)(1-\beta_{t-1}) + c_t} \prod_{s=t+1}^T \left[ \frac{(1 - c_s)(1-\beta_{s-1})}{(1-c_s)(1-\beta_{s-1}) + c_s} \right].
$$
This weight $\alpha_t$ directly dictates the size and shape of the effective averaging window. In Figure~\ref{fig:alpha_hist_beta0.8}, we plot the histogram of these weights $\alpha_t$ (initialized with $\beta_{\rm init}=0.8$) to illustrate how much ``memory'' the final model retains from past steps. The exact shape of this distribution is governed by the interpolation parameter $\rho$. Specifically, when $\rho = 0$, the schedule allows the averaging window size to grow approximately linearly throughout training. In contrast, setting $\rho = 1$ strictly fixes the window size after the initial warmup phase ($T_0 = 2000$). For any intermediate value ($0 < \rho < 1$), the schedule smoothly interpolates between these two extremes, providing a controlled, sub-linear growth of the memory window.
Furthermore, we visualize the actual evolution of $\beta_t$ over time in Figure~\ref{fig:beta_evolution}.

\subsection{Implementation Details}
\label{app:implementation_details}

\paragraph{Averaging Coefficient.}
In the main text, we use the simplified intuition $c_t = 1/t$. 
In the implementation, we follow the learning-rate-weighted averaging rule used in ~\citet{defazio2024the}. 
Let $\eta_t$ denote the effective learning rate at step $t$, including linear warmup. 
Rather than using a uniform average during warmup, we set
\[
c_{t+1}
=
\frac{\eta_t^2}{\sum_{i=1}^{t}\eta_i^2}.
\]
This prevents very early iterations, where the learning rate is small, from being overweighted in the averaged sequence. 
After warmup, $\eta_t$ becomes constant and the coefficient recovers the usual $1/t$-type decay. 

\paragraph{Batch Normalization.}
\label{app:batchnorm}
Averaging methods, including Schedule-Free~\citep{defazio2024the}, require special care for models with BatchNorm. 
During training, gradients are evaluated at the interpolated sequence $\mY_t$, so BatchNorm running statistics are accumulated for the training-time parameter state. 
During evaluation, however, the returned model is the averaged sequence $\mX_t$. 
Therefore, the stored BatchNorm running statistics may not match the parameters used for evaluation.

For models with BatchNorm, we first convert the optimizer/model to the $\mX_t$-state and then refresh BatchNorm statistics by forwarding a small number of training batches without gradient updates. 
This is analogous to PreciseBN-style recalibration~\citep{wu2021rethinking}. 
No such step is needed for normalization layers that do not maintain batch-dependent running statistics, such as LayerNorm or RMSNorm, which are used in our language model experiments.

\paragraph{Weight Decay.}
Following ~\citet{defazio2024the}, weight decay may be applied either to the gradient-evaluation sequence $\vy_t$ or to the base sequence $\vz_t$. In our implementation, we apply the standard weight-decay term to $\vz_t$. We keep decay at $\vy_t$ as a separate optional variant, but do not use it unless explicitly stated.

\paragraph{Computational Resources.}
We use up to 8 NVIDIA RTX A6000 GPUs for the image-domain experiments and the 124M Llama experiments. For experiments at the 720M scale and above, we use up to 8 NVIDIA A100 GPUs and 4 B200 GPUs with data-parallel training.

\subsection{Comparison with Concurrent Methods}
\label{sec:concurrent_work}
We discuss two concurrent works that are closely related to AMUSE: SF-NorMuon, which combines Schedule-Free optimization with normalized Muon-style updates, and ScheduleFree+, which extends Schedule-Free AdamW to large-scale language model pretraining.

\paragraph{SF-NorMuon.} SF-NorMuon~\citep{apte2026anytime} extends NorMuon~\citep{li2025normuon} to the Schedule-Free framework. They provide convergence guarantees for schedule-free spectral optimization and analyze the role of weight decay in long-horizon training. Their analysis suggests that applying weight decay to the fast iterate $\vz_t$ improves stability. This finding is consistent with AMUSE, which also applies decoupled weight decay directly to $\vz_t$ (Algorithm~\ref{alg:amuse}). A key difference is that SF-NorMuon considers a fixed interpolation coefficient $\beta$ throughout training, whereas AMUSE considers a time-varying coefficient $\beta_t$.

\paragraph{ScheduleFree+.}
\citet{defazio2026schedulefree+} proposes ScheduleFree+, an extension of
Schedule-Free AdamW designed for large-scale language model pretraining.
Among other changes, ScheduleFree+ reintroduces Adam's inner first-moment
momentum and incorporates fully decoupled AdamC weight decay and a
Polyak-style adaptive step-size rule. For long-duration training, it also
anneals the Schedule-Free interpolation coefficient $\beta_t$. The closest
connection to our method lies in the design of this coefficient: both methods
increase $\beta_t$ over training, thereby shifting the gradient evaluation
point \(
y_t=(1-\beta_t)z_t+\beta_t x_t
\)
from the fast iterate $z_t$ toward the averaged iterate $x_t$.

ScheduleFree+ implements this transition by log-linearly interpolating the
residual coefficient $1-\beta_t$ between fixed initial and final endpoints:
\[
1-\beta_t
=
(1-\beta_{\mathrm{initial}})^{1-\tau_t}
(1-\beta_{\mathrm{final}})^{\tau_t},
\qquad
\tau_t=\min\left\{\frac{t}{T_{\mathrm{anneal}}},1\right\}.
\]
In its long-duration experiments, the annealing horizon spans the full
training run, whereas the shorter-duration experiments use a fixed
$\beta$. In contrast, following the schedule design in
Appendix~\ref{app:amuse_design}, our method uses the power-law form
\[
1-\beta_t
=
(1-\beta_{\mathrm{init}})
\left(T_0/t\right)^\rho,
\]
where $T_0$ is a warmup iteration and $\rho>0$ controls the rate at which the gradient evaluation point
shifts toward the averaged iterate. Unlike the endpoint-based rule of
ScheduleFree+, our schedule does not require specifying either a terminal
value $\beta_{\mathrm{final}}$ or a total annealing horizon.

We further examine related SF-AdamW variants in
Appendix~\ref{app:sfadamw_beta}, where we study the effect of applying our
$\beta_t$ schedule both with and without Adam's inner first-moment
accumulation.

%% file: D_Image_Domain_Experimental_Details.tex
\section{Image Domain Experiments}
\label{app:image domain detail}
\subsection{Experimental Details}
For the image-domain experiments, we mostly follow the experimental setup of~\citet{defazio2024the}. 
We additionally include Muon with cosine annealing as a new baseline and add the ISIC 2018 image segmentation task.

We follow the standard Muon hybrid setup~\citep{jordan2024muon}: Muon is applied to hidden matrix-valued layers, while SGD with momentum is used for normalization layers, biases, and classifier heads.
For MAE fine-tuning in the image-domain experiments, we instead adopt the Muon+AdamW hybrid setup commonly used for Transformer models~\citep{jordan2024muon,liu2025muon}. Specifically, Muon is applied to the hidden-layer weight matrices of the ViT backbone, while AdamW is used for the classifier head and all one-dimensional parameters, including biases and LayerNorm parameters.

When available, we use the tuned hyperparameters reported by~\citet{defazio2024the} for SGD, SF-SGD, AdamW, and SF-AdamW. 
For the added baseline Muon, we tune the learning rate, weight decay, and Muon momentum following the same tuning protocol. 
For vanilla Muon, we also tune the learning rate of the auxiliary SGD optimizer separately, since using the same learning rate for Muon and SGD led to poor performance.

For AMUSE, we fix the Muon momentum to $0.95$ and use the same learning rate for the auxiliary optimizer as for Muon, reducing the number of tuned hyperparameters. In the image-domain experiments, except for MAE fine-tuning, AMUSE uses SGD without momentum for the non-Muon parameters, which corresponds to an SF-SGD update. For MAE fine-tuning, AMUSE instead uses AdamW with no first-moment momentum and fixes the second-moment coefficient to $\beta_2=0.999$, which corresponds to an SF-AdamW update. Thus, beyond the base learning rate and weight decay, AMUSE only tunes the AMUSE-specific parameters $\beta_{\rm init}$ and $\rho$.

\label{app:img_hyper}
\paragraph{CIFAR-10.}\mbox{}
 We train a Wide-ResNet-16-8 architecture~\citep{DBLP:journals/corr/ZagoruykoK16} on the CIFAR-10 image classification~\citep{krizhevsky2009learning} dataset. We use a batch size of 128 and train for 300 epochs.

\begin{table}[H]\mbox{}
\centering
\caption{\textbf{Hyperparameters for CIFAR-10.}}
\label{tab:hparams_cifar10}
\small
\begin{minipage}[t]{0.48\linewidth}
\centering
\textbf{SGD and SF-SGD}\vspace{0.5em}

\begin{tabular}{lcc}
\hline
\textbf{Hyperparameter} & \textbf{SGD} & \textbf{SF-SGD} \\
\hline
Epochs & \multicolumn{2}{c}{300}  \\
Batch size & \multicolumn{2}{c}{128} \\
Warmup ratio & \multicolumn{2}{c}{0.05} \\
Seeds & \multicolumn{2}{c}{5} \\
Learning Rate & 0.2 & 10.0 \\
Weight decay & 0.0001 & 0.0001 \\
SGD momentum $\beta$ & 0.9 & -- \\
SF-SGD $\beta_1$ & -- & 0.9 \\
LR decay scheduler & cosine & -- \\
\hline
\end{tabular}
\end{minipage}
\hfill
\begin{minipage}[t]{0.48\linewidth}
\centering
\textbf{Muon and AMUSE}\vspace{0.5em}

\begin{tabular}{lcc}
\hline
\textbf{Hyperparameter} & \textbf{Muon} & \textbf{AMUSE} \\
\hline
Epochs & \multicolumn{2}{c}{300} \\
Batch size & \multicolumn{2}{c}{128} \\
Warmup ratio & \multicolumn{2}{c}{0.05} \\
Seeds & \multicolumn{2}{c}{5} \\
Muon LR & 0.05 & 0.2 \\
SGD LR & 0.0001 & 0.2 \\
Weight decay & 0.01 & 0.02 \\
Muon momentum $\mu$ & 0.9 & 0.95 \\
SGD momentum $\beta$ & 0.9 & -- \\
AMUSE $\beta_{\rm init}$ & -- & 0.8 \\
AMUSE $\rho$ & -- & 0.3 \\
LR decay scheduler & cosine & -- \\
\hline
\end{tabular}
\end{minipage}
\end{table}

\newpage
\paragraph{CIFAR-100.}\mbox{}
 We train a DenseNet architecture~\citep{Huang_2017_CVPR} on the CIFAR-100 image classification dataset. We use a batch size of 64 and train for 300 epochs.
 
\begin{table}[H]
\centering
\caption{\textbf{Hyperparameters for CIFAR-100.}}
\label{tab:hparams_cifar100}
\small
\begin{minipage}[t]{0.48\linewidth}
\centering
\textbf{SGD and SF-SGD}\vspace{0.5em}

\begin{tabular}{lcc}
\hline
\textbf{Hyperparameter} & \textbf{SGD} & \textbf{SF-SGD} \\
\hline
Epochs & \multicolumn{2}{c}{300}  \\
Batch size & \multicolumn{2}{c}{64} \\
Warmup ratio & \multicolumn{2}{c}{0.05} \\
Seeds & \multicolumn{2}{c}{5} \\
Learning Rate & 0.05 & 5.0 \\
Weight decay & 2e-4 & 2e-4 \\
SGD momentum $\beta$ & 0.9 & -- \\
SF-SGD $\beta_1$ & -- & 0.9 \\
LR decay scheduler & cosine & -- \\
\hline
\end{tabular}
\end{minipage}
\hfill
\begin{minipage}[t]{0.48\linewidth}
\centering
\textbf{Muon and AMUSE}\vspace{0.5em}

\begin{tabular}{lcc}
\hline
\textbf{Hyperparameter} & \textbf{Muon} & \textbf{AMUSE} \\
\hline
Epochs & \multicolumn{2}{c}{300} \\
Batch size & \multicolumn{2}{c}{64} \\
Warmup ratio & \multicolumn{2}{c}{0.05} \\
Seeds & \multicolumn{2}{c}{5} \\
Muon LR & 0.05 & 0.5 \\
SGD LR & 1e-3 & 0.5 \\
Weight decay & 5e-3 & 2e-3 \\
Muon momentum $\mu$ & 0.9 & 0.95 \\
SGD momentum $\beta$ & 0.9 & -- \\
AMUSE $\beta_{\rm init}$& -- & 0.7 \\
AMUSE $\rho$ & -- & 0.4 \\
LR decay scheduler & cosine & -- \\
\hline
\end{tabular}
\end{minipage}
\end{table}

\paragraph{SVHN.}\mbox{}
 We train a ResNet-3-96 architecture~\citep{He_2016_CVPR} on the Street View House Numbers (SVHN) dataset~\citep{netzer2011reading}. We use a batch size of 32 and train for 300 epochs.

\begin{table}[H]
\centering
\caption{\textbf{Hyperparameters for SVHN.}}
\label{tab:hparams_svhn}
\small
\begin{minipage}[t]{0.48\linewidth}
\centering
\textbf{SGD and SF-SGD}\vspace{0.5em}

\begin{tabular}{lcc}
\hline
\textbf{Hyperparameter} & \textbf{SGD} & \textbf{SF-SGD} \\
\hline
Epochs & \multicolumn{2}{c}{300}  \\
Batch size & \multicolumn{2}{c}{32} \\
Warmup ratio & \multicolumn{2}{c}{0.05} \\
Seeds & \multicolumn{2}{c}{5} \\
Learning Rate & 0.1 & 1.0 \\
Weight decay & 1e-4 & 2e-4 \\
SGD momentum $\beta$ & 0.9 & -- \\
SF-SGD $\beta_1$ & -- & 0.9 \\
LR decay scheduler & cosine & -- \\
\hline
\end{tabular}
\end{minipage}
\hfill
\begin{minipage}[t]{0.48\linewidth}
\centering
\textbf{Muon and AMUSE}\vspace{0.5em}

\begin{tabular}{lcc}
\hline
\textbf{Hyperparameter} & \textbf{Muon} & \textbf{AMUSE} \\
\hline
Epochs & \multicolumn{2}{c}{300} \\
Batch size & \multicolumn{2}{c}{32} \\
Warmup ratio & \multicolumn{2}{c}{0.05} \\
Seeds & \multicolumn{2}{c}{5} \\
Muon LR & 0.05 & 0.05 \\
SGD LR & 1e-3 & 0.05 \\
Weight decay & 2e-3 & 0.01 \\
Muon momentum $\mu$ & 0.9 & 0.95 \\
SGD momentum $\beta$ & 0.9 & -- \\
AMUSE $\beta_{\rm init}$ & -- & 0.8 \\
AMUSE $\rho$ & -- & 0.2 \\
LR decay scheduler & cosine & -- \\
\hline
\end{tabular}
\end{minipage}
\end{table}

\newpage

\paragraph{ImageNet-1k.}\mbox{}
We train a ResNet-50 architecture~\citep{He_2016_CVPR} on the ILSVRC 2012~\citep{russakovsky2015imagenet} ImageNet-1k classification dataset. We use a batch size of 256 and train for 100 epochs.

\begin{table}[H]
\centering
\caption{\textbf{Hyperparameters for ImageNet-1k.}}
\label{tab:hparams_imagenet}
\small
\begin{minipage}[t]{0.48\linewidth}
\centering
\textbf{SGD and SF-SGD}\vspace{0.5em}

\begin{tabular}{lcc}
\hline
\textbf{Hyperparameter} & \textbf{SGD} & \textbf{SF-SGD} \\
\hline
Epochs & \multicolumn{2}{c}{100}  \\
Batch size & \multicolumn{2}{c}{256} \\
Warmup ratio & \multicolumn{2}{c}{0.05} \\
Seeds & \multicolumn{2}{c}{5} \\
Learning Rate & 0.05 & 1.5 \\
Weight decay & 1e-4 & 5e-5 \\
SGD momentum $\beta$ & 0.9 & -- \\
SF-SGD $\beta_1$ & -- & 0.9 \\
LR decay scheduler & cosine & -- \\
\hline
\end{tabular}
\end{minipage}
\hfill
\begin{minipage}[t]{0.48\linewidth}
\centering
\textbf{Muon and AMUSE}\vspace{0.5em}

\begin{tabular}{lcc}
\hline
\textbf{Hyperparameter} & \textbf{Muon} & \textbf{AMUSE} \\
\hline
Epochs & \multicolumn{2}{c}{100} \\
Batch size & \multicolumn{2}{c}{256} \\
Warmup ratio & \multicolumn{2}{c}{0.05} \\
Seeds & \multicolumn{2}{c}{5} \\
Muon LR & 0.1 & 0.5 \\
SGD LR & 1e-5 & 0.5 \\
Weight decay & 1e-3 & 5e-4 \\
Muon momentum $\mu$ & 0.9 & 0.95 \\
SGD momentum $\beta$ & 0.9 & -- \\
AMUSE $\beta_{\rm init}$ & -- & 0.5 \\
AMUSE $\rho$ & -- & 0.8 \\
LR decay scheduler & cosine & -- \\
\hline
\end{tabular}
\end{minipage}
\end{table}

\paragraph{ISIC 2018.}\mbox{} We train a base U-Net architecture~\citep{ronneberger2015u} on the ISIC 2018 dataset~\citep{codella2019skin} for binary medical image segmentation. 
We use a batch size of 8 and train for 50 epochs. 
Our implementation is based on the U-Bench codebase\footnote{\url{https://github.com/FengheTan9/U-Bench}}.

Since this setup is newly added in our paper, we extensively tune the learning rate, weight decay, and momentum for each baseline. 
For vanilla Muon, we additionally tune the auxiliary SGD learning rate used for non-Muon parameters.

\begin{table}[H]
\centering
\caption{\textbf{Hyperparameters for ISIC 2018.}}
\label{tab:hparams_ISIC}
\small
\begin{minipage}[t]{0.48\linewidth}
\centering
\textbf{SGD and SF-SGD}\vspace{0.5em}

\begin{tabular}{lcc}
\hline
\textbf{Hyperparameter} & \textbf{SGD} & \textbf{SF-SGD} \\
\hline
Epochs & \multicolumn{2}{c}{50} \\
Batch size & \multicolumn{2}{c}{8} \\
Warmup ratio & \multicolumn{2}{c}{0.05} \\
Seeds & \multicolumn{2}{c}{5} \\
Learning Rate & 0.02 & 2e-3 \\
Weight decay & 5e-4 & 0.01 \\
SGD momentum $\beta$ & 0.9 & -- \\
SF-SGD $\beta_1$ & -- & 0.9 \\
LR decay scheduler & cosine & -- \\
\hline
\end{tabular}
\end{minipage}
\hfill
\begin{minipage}[t]{0.48\linewidth}
\centering
\textbf{Muon and AMUSE}\vspace{0.5em}

\begin{tabular}{lcc}
\hline
\textbf{Hyperparameter} & \textbf{Muon} & \textbf{AMUSE} \\
\hline
Epochs & \multicolumn{2}{c}{50} \\
Batch size & \multicolumn{2}{c}{8} \\
Warmup ratio & \multicolumn{2}{c}{0.05} \\
Seeds & \multicolumn{2}{c}{5} \\
Muon LR & 0.01 & 0.2 \\
SGD LR & 1e-3 & 0.2 \\
Weight decay & 1e-3 & 0.05 \\
Muon momentum $\mu$ & 0.9 & 0.95 \\
SGD momentum $\beta$ & 0.9 & -- \\
AMUSE $\beta_{\rm init}$ & -- & 0.6 \\
AMUSE $\rho$ & -- & 0.2 \\
LR decay scheduler & cosine & -- \\

\hline
\end{tabular}
\end{minipage}
\end{table}

\paragraph{MAE.}\mbox{}
We fine-tune a pretrained MAE ViT-Base model (\texttt{vit\_base\_patch16}) on ImageNet-1K for 100 epochs. 
The model uses $16\times16$ patches and the standard ViT-Base architecture with 12 transformer blocks, hidden size 768, and 12 attention heads. 

All runs use the same AdamW-pretrained MAE initialization and differ only in optimizer-specific hyperparameters. Since MAE fine-tuning starts from an AdamW-pretrained checkpoint, we carefully tune the initial interpolation value $\beta_{\rm init}$ for AMUSE. 
We use smaller $\beta_{\rm init}$ values than in the other experiments to avoid a large early discrepancy between the base sequence and the averaged sequence, allowing the base trajectory to adapt smoothly to the fine-tuning objective. Our implementation uses the official mae code\footnote{\url{https://github.com/facebookresearch/mae}}.

\begin{table}[H]
\centering
\caption{\textbf{Hyperparameters for MAE fine-tuning.}}
\label{tab:hparams_MAE}
\small
\begin{minipage}[t]{0.48\linewidth}
\centering
\textbf{AdamW and SF-AdamW}\vspace{0.5em}

\begin{tabular}{lcc}
\hline
\textbf{Hyperparameter} & \textbf{AdamW} & \textbf{SF-AdamW} \\
\hline
Epochs & \multicolumn{2}{c}{100} \\
Batch size & \multicolumn{2}{c}{256} \\
Warmup ratio & \multicolumn{2}{c}{0.05} \\
Learning rate & 5e-4 & 2e-3 \\
Weight decay & 0.05 & 0.05 \\
AdamW $\beta_1$ & 0.9 & -- \\
SF-AdamW $\beta_1$ & -- & 0.9 \\
AdamW $\beta_2$ & 0.999 & 0.999 \\
LR decay scheduler & cosine & -- \\
Drop Path & \multicolumn{2}{c}{0.1} \\
Reprob & \multicolumn{2}{c}{0.25} \\
Mixup & \multicolumn{2}{c}{0.8} \\
Cutmix & \multicolumn{2}{c}{1.0} \\

\hline
\end{tabular}
\end{minipage}
\hfill
\begin{minipage}[t]{0.48\linewidth}
\centering
\textbf{Muon and AMUSE}\vspace{0.5em}

\begin{tabular}{lcc}
\hline
\textbf{Hyperparameter} & \textbf{Muon} & \textbf{AMUSE} \\
\hline
Epochs & \multicolumn{2}{c}{100} \\
Batch size & \multicolumn{2}{c}{256} \\
Warmup ratio & \multicolumn{2}{c}{0.05} \\
Learning rate & 2e-4 & 2e-3 \\
Weight decay & 0.01 & 0.05 \\
Muon momentum & 0.95 & 0.95 \\
AdamW $\beta_1$ & 0.9 & -- \\
AMUSE $\beta_{\rm init}$ & -- & 0.4 \\
AdamW $\beta_2$ & 0.999 & 0.999 \\
AMUSE $\rho$ & -- & 0.3 \\
LR decay scheduler & cosine & -- \\
Drop Path & \multicolumn{2}{c}{0.1} \\
Reprob & \multicolumn{2}{c}{0.25} \\
Mixup & \multicolumn{2}{c}{0.8} \\
Cutmix & \multicolumn{2}{c}{1.0} \\
\hline
\end{tabular}
\end{minipage}
\end{table}

%% file: E_Large_Language_Model_Experimental_Details.tex
\section{Large Language Model Experiments}
\label{app:llm_detail}
\subsection{Architecture Details}
Our experimental setup follows~\citet{semenov2026benchmarking}, using their public codebase.\footnote{\url{https://github.com/epfml/llm-optimizer-benchmark}}
We use three decoder-only Llama-style models with approximately 124M, 720M, and 1.3B parameters. 
Across all model sizes, we use a sequence length of 512, the GPT-2 tokenizer with vocabulary size 50,304, RMSNorm with $\epsilon=10^{-5}$, bias-free projections, SwiGLU MLPs with \texttt{multiple\_of}=256, no dropout, and bfloat16 training. 
Table~\ref{tab:llm_architecture} summarizes the model-specific dimensions.
\begin{table}[!h]
\centering
\caption{\textbf{Model configurations for LLM experiments.}}
\label{tab:llm_architecture}
\begin{tabular}{lccc}
\toprule
Model size & Layers & Attention heads & Hidden size \\
\midrule
124M & 12 & 12 & 768 \\
720M & 12 & 16 & 2048 \\
1.3B & 24 & 32 & 2048 \\
\bottomrule
\end{tabular}
\end{table}
\subsection{Hyperparameter Tuning}
\label{app:llm_hyper}
For AdamW and SF-AdamW, we use the best-performing hyperparameter configurations reported by~\citet{semenov2026benchmarking}.

For Muon, we conduct an expanded hyperparameter search of 60 configurations, starting from the best configuration reported by~\citet{semenov2026benchmarking}. The search varies learning rate, warmup steps, weight decay, Muon momentum, and the AdamW second-moment coefficient $\beta_2$, covering both the learning-rate regime near the reported configuration and substantially larger learning rates. For fixed-$\beta$ SF-Muon, we evaluate 37 configurations in the 124M setting. We use the same hyperparameter grids as AMUSE and test fixed interpolation coefficients $\beta \in \{0.6, 0.8, 0.9, 0.95, 0.98\}$.

For AMUSE, we first tune the learning rate and then explore selected combinations of weight decay, warmup steps, and AMUSE-specific hyperparameters. Specifically, we sweep the initial interpolation value $\beta_{\rm init} \in \{0.4, 0.6\}$ and $\rho \in \{0.6, 0.8\}$, with $\rho=0.8$ consistently performing well across our experiments. We do not additionally tune the base-optimizer momentum parameters, fixing the Muon momentum to $\mu=0.95$ and AdamW's second-moment coefficient for non-Muon parameters to $\beta_2=0.999$. In total, we evaluate 34 AMUSE configurations, compared with 60 additional configurations for Muon.

Since~\citet{semenov2026benchmarking} report experiments only up to the 720M scale, we additionally evaluate a 1.3B model to test whether the observed results persist at a larger scale. 
For this setting, we initialize each optimizer from its corresponding 720M hyperparameter configuration and perform a learning-rate sweep for all methods, including AMUSE. See Tables~\ref{tab:hparams_124m}--\ref{tab:sfadamw_amuse_hparams_1.3b} for the selected hyperparameters, and Table~\ref{tab:amuse_search_grid_llm} for the AMUSE hyperparameter search grid.

\clearpage
\subsubsection{Hyperparameters for 124M parameters model}\mbox{}

\begin{table}[!ht]
\centering
\caption{\textbf{Selected hyperparameters at the 124M scale.}}
\label{tab:hparams_124m}
\small
\begin{minipage}[t]{0.48\linewidth}
\centering
\textbf{AdamW and Muon (D-Muon)}\vspace{0.5em}

\begin{tabular}{lcc}
\hline
\textbf{Hyperparameter} & \textbf{AdamW} & \textbf{Muon} \\
\hline
Learning rate & 0.001 & 0.01 \\
Batch size & 256 & 256 \\
Sequence length & 512 & 512 \\
Warmup steps & 2000 & 2000 \\
Weight decay & 0.1 & 0.05 \\
Gradient clipping & 0.5 & 0.5 \\
AdamW $\beta_1$ & 0.8 & 0.8 \\
AdamW $\beta_2$ & 0.999 & 0.999 \\
Muon momentum & -- & 0.95 \\
LR decay scheduler & cosine & cosine \\
\hline
\end{tabular}
\end{minipage}
\hfill
\begin{minipage}[t]{0.48\linewidth}
\centering
\textbf{SF-AdamW and AMUSE}\vspace{0.5em}

\begin{tabular}{lcc}
\hline
\textbf{Hyperparameter} & \textbf{SF-AdamW} & \textbf{AMUSE} \\
\hline
Learning rate & 0.002 & 0.01 \\
Batch size & 256 & 256 \\
Sequence length & 512 & 512 \\
Warmup steps & 8000 & 4000 \\
Weight decay & 0.1 & 0.05 \\
Gradient clipping & 0.5 & 0.5 \\
AMUSE $\beta_{\rm init}$ & -- & 0.6 \\
AMUSE $\rho$ & -- & 0.8 \\
SF-AdamW $\beta_1$ & 0.9 & -- \\
AdamW $\beta_2$ & 0.9999 & 0.999 \\
Muon momentum & -- & 0.95 \\
LR decay scheduler & -- & -- \\
\hline
\end{tabular}
\end{minipage}
\end{table}

\subsubsection{Hyperparameters for 720M parameters model}
\begin{table}[H]
\centering
\caption{\textbf{Selected hyperparameters at the 720M scale.}}
\label{tab:hparams_720m}
\small
\begin{minipage}[t]{0.48\linewidth}
\centering
\textbf{AdamW and Muon (D-Muon)}\vspace{0.5em}

\begin{tabular}{lcc}
\hline
\textbf{Hyperparameter} & \textbf{AdamW} & \textbf{Muon} \\
\hline
Learning rate & 0.001 & 0.001 \\
Batch size & 1984 & 1984 \\
Sequence length & 512 & 512 \\
Warmup steps & 2000 & 2000 \\
Weight decay & 0.1 & 0.1 \\
Gradient clipping & 0.1 & 0.1 \\
AdamW $\beta_1$ & 0.9 & 0.9 \\
AdamW $\beta_2$ & 0.999 & 0.99 \\
Muon momentum & -- & 0.95 \\
LR decay scheduler & cosine & cosine \\
\hline
\end{tabular}
\end{minipage}
\hfill
\begin{minipage}[t]{0.48\linewidth}
\centering
\textbf{SF-AdamW and AMUSE}\vspace{0.5em}

\begin{tabular}{lcc}
\hline
\textbf{Hyperparameter} & \textbf{SF-AdamW} & \textbf{AMUSE} \\
\hline
Learning rate & 0.001 & 0.01 \\
Batch size & 1984 & 1984 \\
Sequence length & 512 & 512 \\
Warmup steps & 8000 & 2000 \\
Weight decay & 0.1 & 0.1 \\
Gradient clipping & 0.1 & 0.1 \\
AMUSE $\beta_{\rm init}$ & -- & 0.4 \\
AMUSE $\rho$ & -- & 0.8 \\
SF-AdamW $\beta_1$ & 0.9 & -- \\
AdamW $\beta_2$ & 0.9999 & 0.999 \\
Muon momentum & -- & 0.95 \\
LR decay scheduler & -- & -- \\
\hline
\end{tabular}
\end{minipage}
\end{table}

\subsubsection{Hyperparameters for 1.3B parameters model}

\begin{table}[!ht]
\centering
\caption{\textbf{Selected hyperparameters for AdamW and Muon (D-Muon) at the 1.3B scale.}
We initialize from the 720M configuration and sweep the learning rate. Bold values indicate the selected learning rate.}
\label{tab:adamw_muon_hparams_1.3b}
\small
\begin{tabular}{lcc}
\hline
\textbf{Hyperparameter} & \textbf{AdamW} & \textbf{Muon} \\
\hline
Learning rate & 0.0005, \textbf{0.001}, 0.002 & 0.0005, 0.001, \textbf{0.002} \\
Batch size & 2048 & 2048 \\
Sequence length & 512 & 512 \\
Warmup steps & 2000 & 2000 \\
Weight decay & 0.1 & 0.1 \\
Gradient clipping & 0.1 & 0.1 \\
AdamW $\beta_1$ & 0.9 & 0.9 \\
AdamW $\beta_2$ & 0.999 & 0.99 \\
Muon momentum & -- & 0.95 \\
LR decay scheduler & cosine & cosine \\
\hline
\end{tabular}
\end{table}

\vspace{-0.5em}

\begin{table}[!ht]
\centering
\caption{\textbf{Selected hyperparameters for SF-AdamW and AMUSE at the 1.3B scale.}
We initialize from the 720M configuration and sweep the learning rate. Bold values indicate the selected learning rate.}
\label{tab:sfadamw_amuse_hparams_1.3b}
\small
\begin{tabular}{lcc}
\hline
\textbf{Hyperparameter} & \textbf{SF-AdamW} & \textbf{AMUSE} \\
\hline
Learning rate & 0.0005, 0.001, \textbf{0.002} & 0.005, \textbf{0.01}, 0.02 \\
Batch size & 2048 & 2048 \\
Sequence length & 512 & 512 \\
Warmup steps & 8000 & 2000 \\
Weight decay & 0.1 & 0.1 \\
Gradient clipping & 0.1 & 0.1 \\
AMUSE $\beta_{\rm init}$ & -- & 0.4 \\
AMUSE $\rho$ & -- & 0.8 \\
SF-AdamW $\beta_1$ & 0.9 & -- \\
AdamW $\beta_2$ & 0.9999 & 0.999 \\
Muon momentum & -- & 0.95 \\
LR decay scheduler & -- & -- \\
\hline
\end{tabular}
\end{table}
\newpage
\subsubsection{Search grid table for AMUSE}

\begin{table}[!h]
\centering
\caption{\textbf{AMUSE hyperparameter search grid for LLM pretraining.}
Selected values are shown in bold.}
\label{tab:amuse_search_grid_llm}
\small
\begin{tabular}{llc}
\hline
\textbf{Scale} & \textbf{Hyperparameter} & \textbf{Search grid} \\
\hline
\multirow{7}{*}{124M}
& Learning rate & $0.002,\ 0.005,\ \mathbf{0.01},\ 0.02$ \\
& Batch size & $256$ \\
& Sequence length & $512$ \\
& Warmup steps & $2000,\ \mathbf{4000}$ \\
& Weight decay & $0.01,\ \mathbf{0.05},\ 0.1$ \\
& Gradient clipping & $\mathbf{0.5}$ \\
& AMUSE $\beta_{\rm init}$ & $0.4,\ \mathbf{0.6}$ \\
& AMUSE $\rho$ & $0.6,\ \mathbf{0.8}$ \\
\hline
\multirow{7}{*}{720M}
& Learning rate & $0.002,\ 0.005,\ \mathbf{0.01}, 0.02$ \\
& Batch size & $1984$ \\
& Sequence length & $512$ \\
& Warmup steps & $\mathbf{2000}$ \\
& Weight decay & $\mathbf{0.1}$ \\
& Gradient clipping & $\mathbf{0.1}$ \\
& AMUSE $\beta_{\rm init}$ & $\mathbf{0.4},\ 0.6$ \\
& AMUSE $\rho$ & $\mathbf{0.8}$ \\
\hline
\multirow{7}{*}{1.3B}
& Learning rate & $0.005$,\ $\mathbf{0.01}$,\ $0.02$ \\
& Batch size & $2048$ \\
& Sequence length & $512$ \\
& Warmup steps & $\mathbf{2000}$ \\
& Weight decay & $\mathbf{0.1}$ \\
& Gradient clipping & $\mathbf{0.1}$ \\
& AMUSE $\beta_{\rm init}$ & $\mathbf{0.4}$ \\
& AMUSE $\rho$ & $\mathbf{0.8}$ \\
\hline
\multicolumn{3}{l}{\textit{Fixed across sweeps:} AdamW $\beta_2=0.999$, Muon momentum $0.95$, no LR decay.} \\
\hline
\end{tabular}
\end{table}

%% file: F_Additional_Experiments.tex
\section{Additional Experiments}
\label{app:additional_experiments}

\subsection{Comparison with Additional Optimizer Baselines}
\label{app:llm_more_baselines}
\label{app:muon_variants}

\begin{figure}[H]
    \centering
    \includegraphics[width=0.7\linewidth]{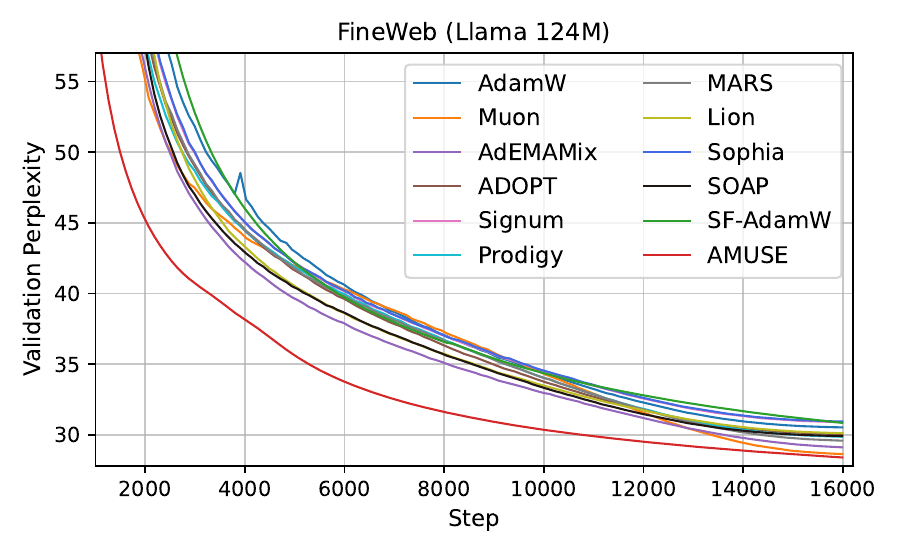}
    \caption{
    \textbf{Comparison with Other Baselines on Llama 124M model.}
    }
    \label{fig:all_opts}
\end{figure}

We first compare AMUSE against a broad set of optimizer baselines for pretraining a 124M-parameter Llama model. In addition to AdamW, Muon, and SF-AdamW, we evaluate AdEMAMix, ADOPT, Signum, Prodigy, MARS, Lion, Sophia, and SOAP using the tuned hyperparameters reported by~\citet{semenov2026benchmarking}. As shown in Figure~\ref{fig:all_opts}, AMUSE consistently achieves the lowest validation perplexity throughout training. AMUSE also attains the best final validation perplexity, indicating that its advantage is not limited to the main baselines considered in the paper.

\paragraph{Recent Muon Variants.}\mbox{}
We also compare AMUSE with two recent Muon variants:
AdaMuon~\citep{si2025adamuon}, an adaptive variant of Muon,
and Mousse~\citep{zhang2026mousse}, a curvature-aware variant.
Both methods are evaluated on the same 124M Llama pretraining setup.

For Mousse, we evaluate 45 hyperparameter configurations,
varying learning rate, weight decay, warmup steps, and Shampoo
momentum, while keeping the curvature exponent fixed.
For AdaMuon, we tune learning rate, weight decay, warmup steps,
the auxiliary AdamW second-moment coefficient, and the shared
momentum/second-moment coefficient.

Table~\ref{tab:muon_variants} compares validation perplexity,
wall-clock time, optimizer-state memory, and median throughput.
While both AdaMuon and Mousse achieve competitive performance,
AMUSE attains the lowest validation perplexity.
Moreover, AMUSE achieves comparable runtime and throughput
to AdaMuon, while Mousse incurs substantially higher
computational and memory overhead.

\begin{table}[!ht]
\centering
\small
\caption{Comparison with recent Muon variants on 124M Llama
pretraining on one NVIDIA B200 GPU. Lower validation perplexity is better.}
\label{tab:muon_variants}
\begin{tabular}{lcccc}
\toprule
Method & Val. PPL $\downarrow$ & Time & State (GiB)
& Throughput (k tok/s) \\
\midrule
AdaMuon & 28.70 & 1h 54m & \textbf{0.76} & 376.8 \\
Mousse  & 28.56 & 2h 48m & 2.52 & 366.4 \\
\midrule
AMUSE   & \textbf{28.48} & \textbf{1h 53m}
        & 0.92 & \textbf{377.8} \\
\bottomrule
\end{tabular}
\end{table}

\FloatBarrier

\subsection{Comparison with Muon with EWA and WSD}
\label{app:muon_ewa_wsd}

We further compare AMUSE against two extensions of constant step size Muon with warmup: exponential weight averaging (EWA) and warmup-stable-decay (WSD) scheduling. These baselines test two different alternatives to AMUSE. EWA examines whether AMUSE's gains can be recovered by applying an external averaging mechanism to a Muon trajectory, whereas WSD examines whether they can be matched by adding an explicit learning rate decay phase.

For the 124M Llama experiment, we first tune constant step size Muon over learning rate, warmup steps, and weight decay, while keeping the momentum parameters identical to those used in the cosine-decay Muon baseline. The best constant step size Muon configuration uses learning rate $10^{-3}$, weight decay $0.05$, and $1,000$ warmup steps. With these tuned hyperparameters, we evaluate EWA with multiple averaging coefficients $(0.99, 0.999, 0.9995)$. Separately, to evaluate the effect of WSD, we start from multiple Muon checkpoints and apply a 2k-iteration decay phase.

\begin{figure}[H]
\centering
\begin{subfigure}{0.48\linewidth}
    \centering
    \includegraphics[width=\linewidth]{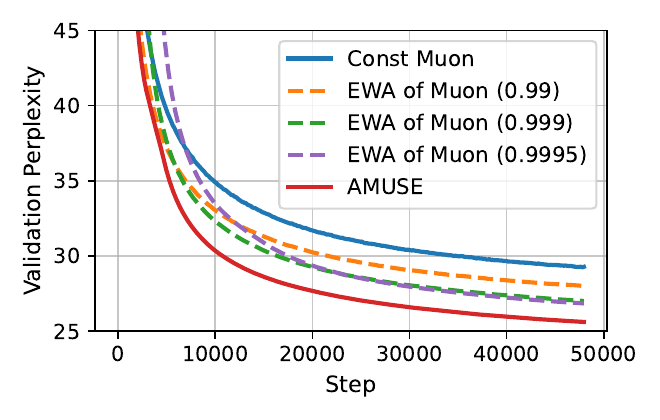}
    \caption{Muon with multiple EWA coefficients}
    \label{fig:fineweb_muon_ewa}
\end{subfigure}
\hfill
\begin{subfigure}{0.48\linewidth}
    \centering
    \includegraphics[width=\linewidth]{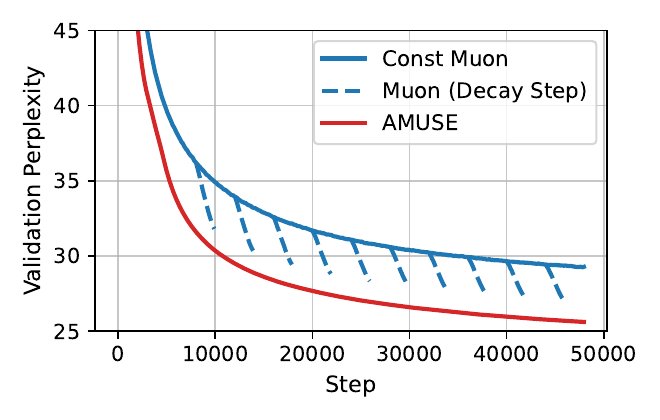}
    \caption{Muon with WSD decay}
    \label{fig:fineweb_muon_wsd}
\end{subfigure}

\caption{
\textbf{Comparison with Muon+EWA and Muon+WSD on 124M Llama pretraining.}
(\textbf{a}): We apply exponential weight averaging (EWA) with multiple averaging coefficients to the tuned constant-LR Muon trajectory.
(\textbf{b}): We apply 2k-step decay phase starting from multiple Muon checkpoints.
Both EWA and WSD improve constant-LR Muon, but none of these variants matches AMUSE.
}
\label{fig:comparison_constmuon_ema}
\end{figure}

As shown in Figure~\ref{fig:comparison_constmuon_ema}, both strategies improve constant step size Muon. Stronger EWA coefficients lead to better validation perplexity, suggesting that external averaging helps stabilize the final model. Similarly, applying a short decay phase from Muon checkpoints substantially lowers perplexity, showing that part of Muon's degradation can be mitigated by an explicit terminal decay. However, none of the EWA variants or WSD decay runs matches AMUSE. Moreover, AMUSE consistently achieves better perplexity across the evaluated training trajectory. These results suggest that AMUSE does not merely imitate post-hoc averaging or learning rate decay. Instead, by stabilizing the gradient-evaluation trajectory during training, AMUSE preserves Muon's fast progress while maintaining better optimization stability.

\begin{figure}[!h]
    \centering
    \includegraphics[width=0.55\linewidth]{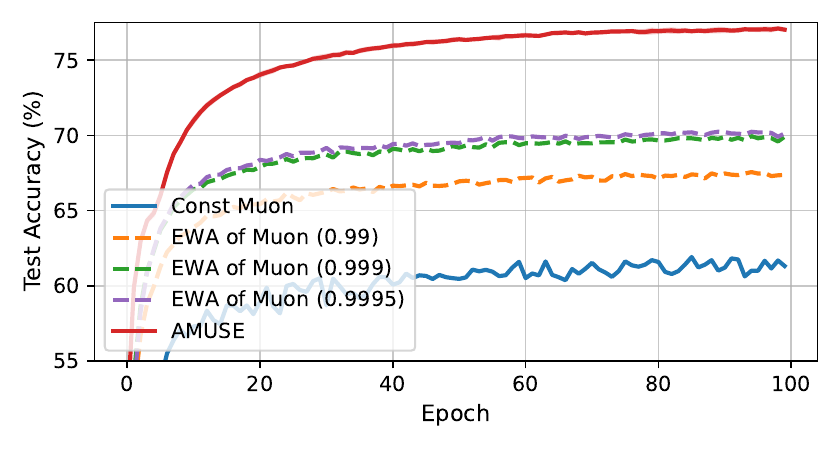}
    \caption{\textbf{Comparison with Muon+EWA on ImageNet.} AMUSE achieves the best performance, outperforming Muon with EWA across all averaging coefficients.}
    \label{fig:comparison_constmuon_imagenet}
\end{figure}

For the image-domain experiments, we also evaluate Muon with EWA on ImageNet, using the same grid of EWA coefficients as in the Llama experiments. We start from the best hyperparameter configuration of the cosine-scheduled Muon baseline, but train Muon with a constant learning rate, and then apply EWA to the resulting trajectory. As shown in Figure~\ref{fig:comparison_constmuon_imagenet}, Muon with EWA does not outperform AMUSE for any averaging coefficient. Together with the Llama results, this suggests that AMUSE does not simply behave like an external weight averaging mechanism. Instead, AMUSE appears to maintain a faster and more stable trajectory during training.

\subsection{Comparison with Post-hoc Weight-Averaging Methods}
\label{app:weight_averaging}
To examine whether AMUSE's improvements can be reproduced
by post-hoc weight averaging, we compare AMUSE with
established methods, including Stochastic Weight Averaging
(SWA)~\citep{izmailov2018averaging},
Latest Weight Averaging (LAWA)~\citep{kaddour2022stop},
and Pre-trained Model Average (PMA)~\citep{li2025model},
on 124M Llama pretraining.

\paragraph{Experimental setup.}
We apply these methods to tuned constant-learning-rate Muon trajectories.
We save checkpoints every 500 iterations for three promising learning rates and tune the averaging hyperparameters for each method, including the averaging start point, window size, checkpoint stride, and exponential decay rate, as applicable.
For PMA-EMA, we additionally tune the exponential decay parameter
$\alpha$.

\paragraph{Results.}
Table~\ref{tab:averaging_results} reports the best results at
different training steps.
All weight-averaging methods substantially improve upon the Muon
baseline. Nevertheless, AMUSE consistently achieves the best
performance across all stopping points.

\begin{table}[t]
\centering
\small
\caption{Comparison with post-hoc weight-averaging methods
on 124M Llama pretraining. Lower is better.}
\label{tab:averaging_results}
\begin{tabular}{lcccc}
\toprule
Method & 8K & 10K & 13K & 16K \\
\midrule
Muon baseline & 36.021 & 34.872 & 33.716 & 32.928 \\
Muon + SWA & 33.169 & 31.952 & 30.773 & 29.956 \\
Muon + LAWA & 33.572 & 32.087 & 30.784 & 29.950 \\
Muon + PMA-EMA & 33.754 & 32.142 & 30.765 & 29.910 \\
Muon + PMA-WMA & 33.410 & 31.981 & 30.719 & 29.906 \\
\midrule
AMUSE & \textbf{31.716} & \textbf{30.447}
      & \textbf{29.262} & \textbf{28.484} \\
\bottomrule
\end{tabular}
\end{table}

Beyond the performance difference, AMUSE operates online without requiring checkpoint storage or post-hoc selection of averaging configurations. In contrast, LAWA and the windowed PMA variants require access to multiple saved checkpoints in our evaluation. This highlights the distinction between post-hoc averaging of an existing trajectory and AMUSE's online modification of the gradient-evaluation point.

\subsection{Longer Training Results}
\label{app:training_longer_iteration}

We further train AMUSE and the main baselines on 124M and 720M Llama models for extended horizons, corresponding to approximately $2.5\times$ and $4.5\times$ the Chinchilla token budget, respectively. Following~\citet{semenov2026benchmarking}, we retain
the hyperparameter configurations tuned for $16{,}000$ iterations
and evaluate performance at different stopping points.

\begin{figure}[ht]
\centering
\begin{subfigure}{0.48\linewidth}
    \centering
    \includegraphics[width=\linewidth]{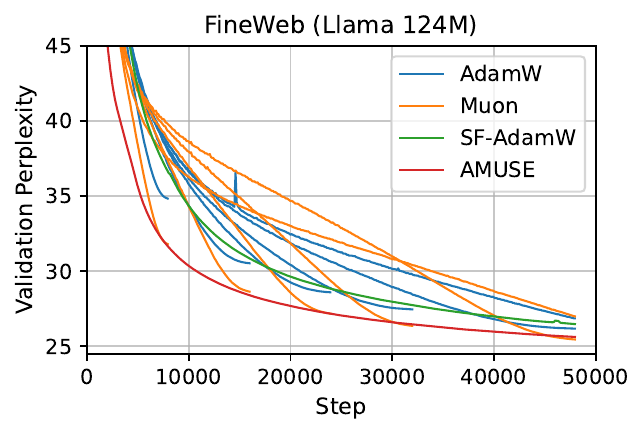}
    \caption{124M Longer-horizon Training}
    \label{fig:fineweb_124m_longer}
\end{subfigure}
\hfill
\begin{subfigure}{0.48\linewidth}
    \centering
    \includegraphics[width=\linewidth]{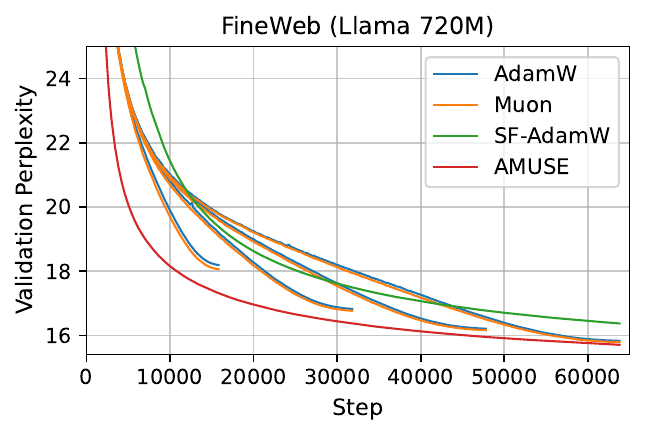}
    \caption{720M Longer-horizon Training}
    \label{fig:fineweb_720m_longer}
\end{subfigure}
\caption{\textbf{Longer-horizon language model pretraining on FineWeb.}
We train AMUSE and the main baselines on 124M and 720M Llama models for approximately $2.5\times$ and $4.5\times$ the Chinchilla token budget, respectively. Unlike cosine-scheduled baselines that require a predefined training horizon, a single AMUSE trajectory performs strongly across training horizons, highlighting its anytime training capability.}
\label{fig:fineweb_longer}
\end{figure}

As shown in Figure~\ref{fig:fineweb_longer}, AMUSE achieves the lowest validation perplexity at most evaluated stopping points for the 124M model and consistently outperforms all baselines across the evaluated training budgets for the 720M model.
Notably, unlike cosine-scheduled baselines whose learning-rate schedules must be specified for a predefined training horizon, AMUSE requires only a single run with no explicit decay schedule. This \textit{single AMUSE trajectory} performs strongly across different training horizons, suggesting that its averaged trajectory remains effective beyond the standard training budget.

\subsection{Hyperparameter Sensitivity}
\label{app:hyper_sensitivity}
\begin{figure}[H]
\centering
\begin{subfigure}{0.48\linewidth}
    \centering
    \includegraphics[width=\linewidth]{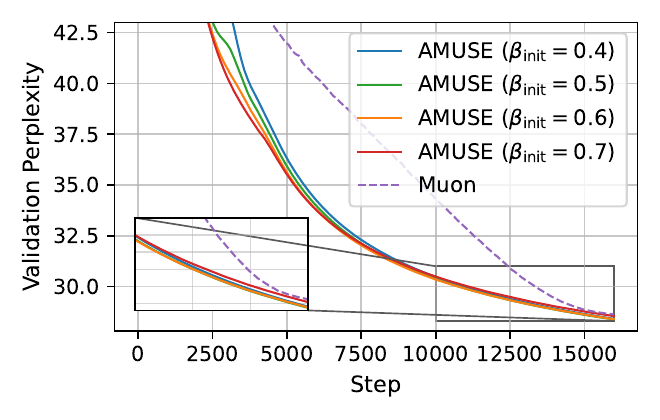}
    \caption{$\beta_{\rm init}$ Sweep Results}
    \label{fig:fineweb_sweep_beta}
\end{subfigure}
\hfill
\begin{subfigure}{0.48\linewidth}
    \centering
    \includegraphics[width=\linewidth]{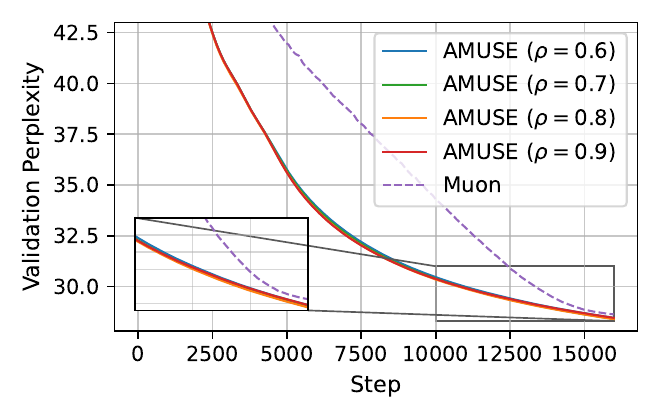}
    \caption{$\rho$ Sweep Results}
    \label{fig:fineweb_sweep_rho}
\end{subfigure}
\caption{
\textbf{Hyperparameter sensitivity varying $\beta_{\rm init}$ and $\rho$ on Llama-124M training.}
AMUSE remains robust to both $\beta_{\rm init}$ and $\rho$, exhibiting stable learning curves across the tested configurations.}
\label{fig:fineweb_sweep}
\end{figure}

\paragraph{Language Model Pretraining (FineWeb with 124M Llama).}\mbox{}
Figure~\ref{fig:fineweb_sweep} and Table~\ref{tab:amuse_hparam_sensitivity} show the sensitivity of AMUSE to $\beta_{\rm init}$ and $\rho$ in the 124M Llama setting. AMUSE shows only mild variation in validation perplexity across the tested ranges.

\begin{table}[H]
\centering
\caption{\textbf{Hyperparameter sensitivity of AMUSE on 124M Llama pretraining.}}
\label{tab:amuse_hparam_sensitivity}
\small
\begin{tabular}{lcc}
\hline
\textbf{Sweep} & \textbf{Hyperparameter value} & \textbf{Validation perplexity} \\
\hline
\multirow{4}{*}{$\beta_{\rm init}$ sweep, $\rho=0.8$}
& $0.4$ & 28.50 \\
& $0.5$ & 28.47 \\
& $0.6$ & 28.48 \\
& $0.7$ & 28.64 \\
\hline
\multirow{4}{*}{$\rho$ sweep, $\beta_{\rm init}=0.6$}
& $0.6$ & 28.55 \\
& $0.7$ & 28.50 \\
& $0.8$ & 28.48 \\
& $0.9$ & 28.56 \\
\hline
Muon with cosine decay & -- & 28.63 \\
\hline
\end{tabular}
\end{table}

\begin{figure}[H]
\centering
\begin{subfigure}{0.48\linewidth}
    \centering
    \includegraphics[width=\linewidth]{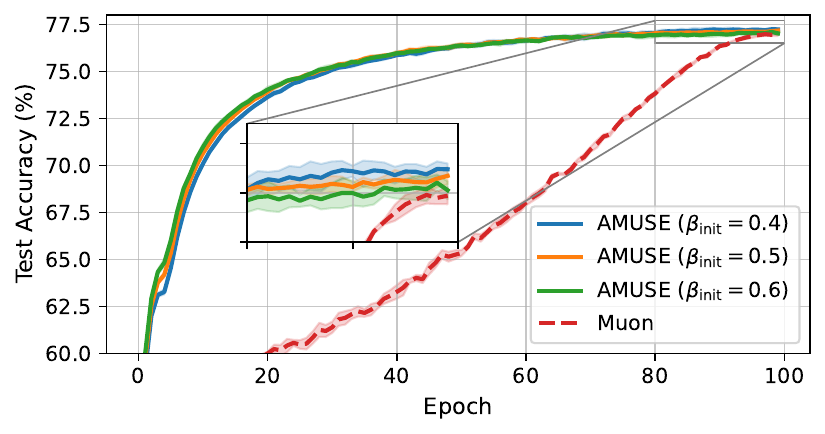}
    \caption{$\beta_{\rm init}$ Sweep Results}
    \label{fig:imagenet_sweep_beta}
\end{subfigure}
\hfill
\begin{subfigure}{0.48\linewidth}
    \centering
    \includegraphics[width=\linewidth]{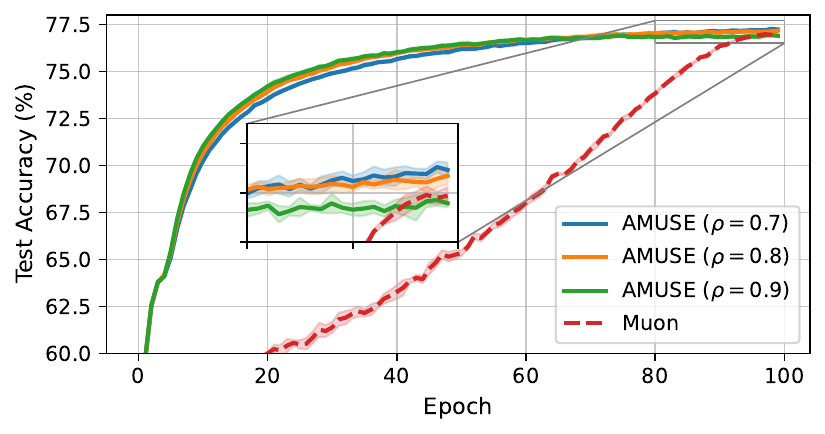}
    \caption{$\rho$ Sweep Results}
    \label{fig:imagenet_sweep_rho}
\end{subfigure}

\caption{
\textbf{Hyperparameter sensitivity varying $\beta_{\rm init}$ and $\rho$ on ImageNet training.} AMUSE exhibits stable performance across a range of hyperparameter choices. Results are averaged over five runs.
}
\label{fig:imagenet_sweep}
\end{figure}

\paragraph{Image Classification (ImageNet with ResNet-50).}\mbox{}
Figure~\ref{fig:imagenet_sweep} and Table~\ref{tab:amuse_hparam_sensitivity_imagenet} show analogous results in the image-domain setting, using ImageNet with ResNet-50. AMUSE is similarly robust to both $\beta_{\rm init}$ and $\rho$: its learning curves remain stable across the tested ranges, and nearly all configurations outperform the tuned Muon baseline with cosine learning-rate decay.

\begin{table}[H]
\centering
\caption{\textbf{Hyperparameter sensitivity of AMUSE using ImageNet with ResNet-50.}}
\label{tab:amuse_hparam_sensitivity_imagenet}
\small
\begin{tabular}{lcc}
\hline
\textbf{Sweep} & \textbf{Hyperparameter value} & \textbf{Test Accuracy (\%)} \\
\hline
\multirow{3}{*}{$\beta_{\rm init}$ sweep, $\rho=0.8$}
& $0.4$ & 77.24 \\
& $0.5$ & 77.17 \\
& $0.6$ & 77.03 \\
\hline
\multirow{3}{*}{$\rho$ sweep, $\beta_{\rm init}=0.5$}
& $0.7$ & 77.23 \\
& $0.8$ & 77.17 \\
& $0.9$ & 76.89 \\
\hline
Muon with cosine decay & -- & 76.97 \\
\hline
\end{tabular}
\end{table}

\subsection{Wall-clock Time Comparison}
\label{app:wall_clock}

We compare the per-iteration wall-clock time of different optimizers in the 124M Llama pretraining setting. Following common practice in optimizer runtime comparisons, we report the average training iteration time, including forward pass, backward pass, and optimizer update. All optimizers are evaluated with the same model, sequence length, global batch size, precision, hardware, and distributed training configuration. AMUSE has a per-iteration cost similar to Muon, while SF-AdamW has a per-iteration cost similar to AdamW. The main runtime difference between these two groups comes from the Newton--Schulz iterations used to orthogonalize Muon updates, which are also used by AMUSE for Muon-updated layers.

\begin{table}[H]
\centering
\caption{\textbf{Wall-clock time comparison on 124M Llama pretraining.}
We report average iteration time on four NVIDIA A6000 GPUs after discarding warmup iterations. Each iteration processes $256 \times 512$ tokens. Relative time is normalized by AdamW.}
\label{tab:wall_clock_124m}
\small
\begin{tabular}{lccc}
\hline
\textbf{Optimizer} & \textbf{Iter. time (s)} & \textbf{Relative time} & \textbf{Throughput (tokens/s)} \\
\hline
AdamW    & 0.674 & $1.000$ & 194469 \\
SF-AdamW & 0.684 & 1.015 & 191626 \\
Muon     & 0.706 & 1.047 & 185654 \\
AMUSE    & 0.719 & 1.067 & 182298 \\
\hline
\end{tabular}
\end{table}

\begin{figure}[t]
    \centering
    \includegraphics[width=\linewidth]{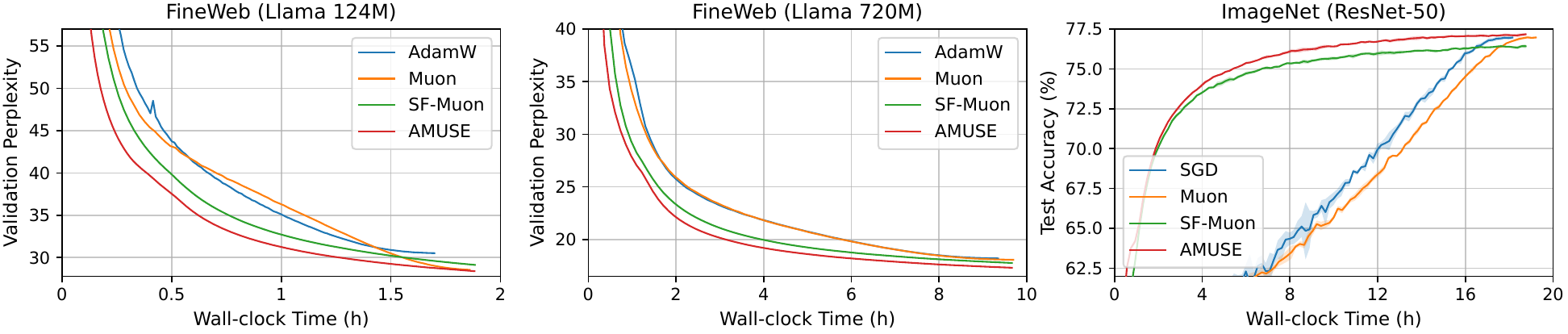}
    \caption{Validation perplexity and test accuracy versus wall-clock time.  Curves end at the last recorded evaluation.}
    \label{fig:wallclock_curves}
\end{figure}

Figure~\ref{fig:wallclock_curves} compares validation perplexity and test accuracy against wall-clock time for 124M and 720M Llama pretraining and ResNet-50 training on ImageNet. AMUSE retains its performance advantage with a total runtime comparable to Muon. We obtain the time coordinates by scaling logged training progress by each method's total runtime, assuming a constant step or epoch cost; SF-Muon uses the same runtime as AMUSE.

\paragraph{Time to target.}
Table~\ref{tab:time_to_target} reports the training progress and wall-clock time needed to reach fixed performance targets, using the measured average training throughput over 16K LLM iterations or 100 ImageNet epochs. For ImageNet SF-Muon, we use the first recorded evaluation of the five-seed mean curve in Figure~\ref{fig:wallclock_curves} that exceeds the target, counting completed training epochs and using AMUSE's total runtime. Timing measurements use one NVIDIA B200 GPU for 124M Llama and ImageNet, and four B200 GPUs with data parallelism for 720M Llama.

\begin{table}[!ht]
\centering
\small
\caption{Training progress and wall-clock time needed to first reach fixed performance targets. Times are rounded to the nearest minute.}
\label{tab:time_to_target}
\setlength{\tabcolsep}{8pt}
\begin{tabular}{llrr}
\toprule
Setting and target & Optimizer & Steps / epochs & Time to target \\
\midrule
\multirow{4}{*}{\shortstack[l]{Llama 124M\\Validation PPL $<33$}}
& AdamW & 11,270 & 1h\,12m \\
& Muon & 10,810 & 1h\,16m \\
& SF-Muon & 8,050 & 57m \\
& AMUSE & \textbf{6,670} & \textbf{47m} \\
\midrule
\multirow{4}{*}{\shortstack[l]{Llama 720M\\Validation PPL $<19$}}
& AdamW & 12,000 & 7h\,06m \\
& Muon & 11,600 & 7h\,07m \\
& SF-Muon & 9,000 & 5h\,34m \\
& AMUSE & \textbf{7,000} & \textbf{4h\,17m} \\
\midrule
\multirow{4}{*}{\shortstack[l]{ImageNet (ResNet-50)\\Test accuracy $>75\%$}}
& SGD & 84 & 15h\,15m \\
& Muon & 85 & 16h\,20m \\
& SF-Muon & 36 & 6h\,45m \\
& AMUSE & \textbf{28} & \textbf{5h\,15m} \\
\bottomrule
\end{tabular}
\end{table}

AMUSE reaches every target in the least wall-clock time, improving time to target by approximately $1.21\times$, $1.30\times$, and $1.29\times$ over the next-fastest method for 124M Llama, 720M Llama, and ImageNet, respectively.

\subsection{Ablation Studies}
\label{app: ablation studies}
\begin{figure}[ht]
    \centering
    \begin{subfigure}{0.48\linewidth}
        \centering
        \includegraphics[width=\linewidth]{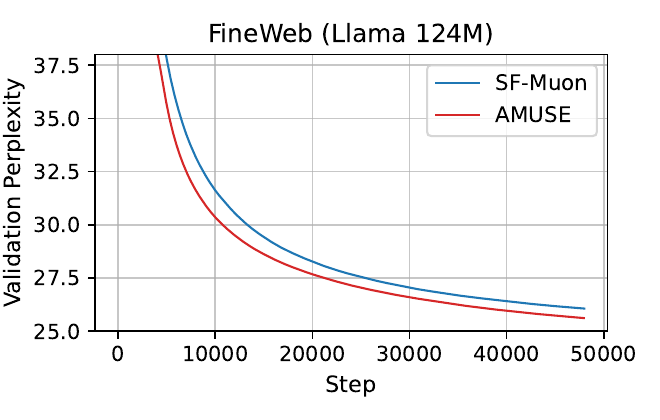}
        \caption{FineWeb Llama 124M}
        \label{fig:fineweb_sf-muon_comparison_124M}
    \end{subfigure}
    \hfill
    \begin{subfigure}{0.485\linewidth}
        \centering
        \includegraphics[width=\linewidth]{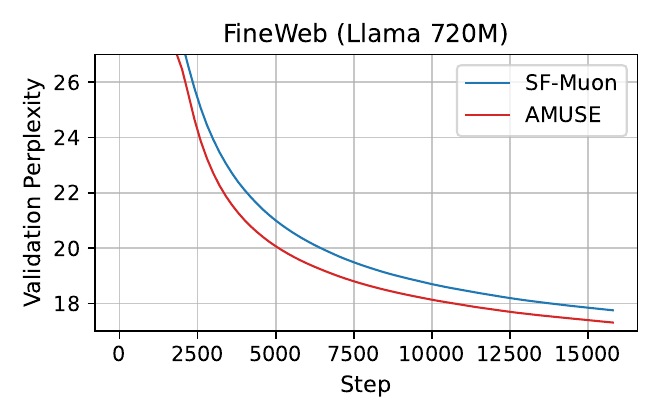}
        \caption{FineWeb Llama 720M}
        \label{fig:fineweb_sf-muon_comparison_720M}
    \end{subfigure}
    \caption{
\textbf{AMUSE and fixed-$\beta$ SF-Muon on 124M/720M Llama pretraining.}
    We compare AMUSE with SF-Muon using the tuned fixed interpolation value, $\beta=0.95$. AMUSE achieves lower validation perplexity at both model scales.
    }
    \label{fig:fineweb_sf-muon_comparison}
\end{figure}

\begin{figure}[ht]
    \centering
    \includegraphics[width=1\linewidth]{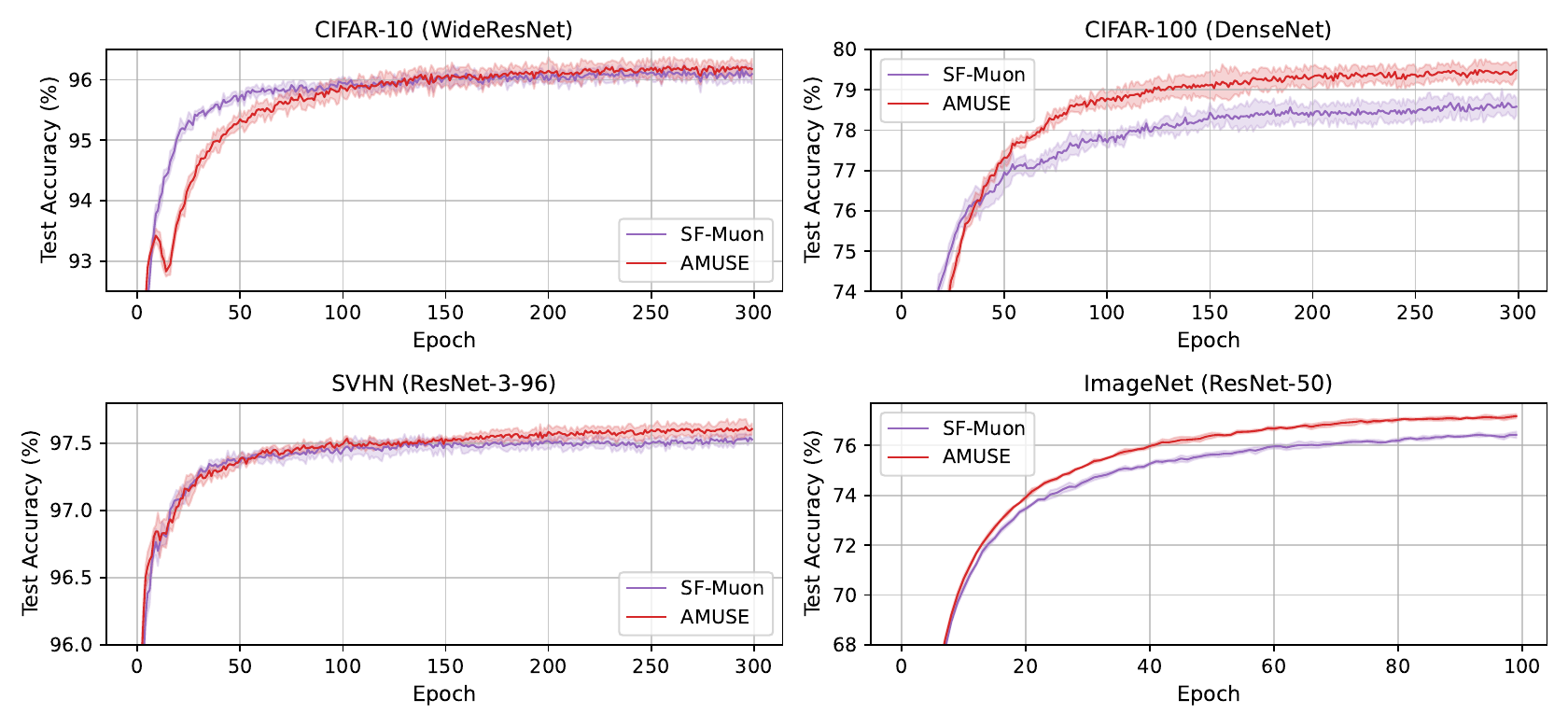}
    \caption{\textbf{AMUSE and fixed-$\beta$ SF-Muon on Image domain experiments.} We compare AMUSE with SF-Muon in image classification benchmarks.}
    \label{fig:image_sf-muon_comparison}
\end{figure}

\paragraph{Comparison with SF-Muon.}\mbox{}
We compare AMUSE with fixed-$\beta$ SF-Muon.
SF-Muon is tuned over the same hyperparameter grid, including multiple fixed $\beta$ values. As shown in Figures~\ref{fig:fineweb_sf-muon_comparison_124M}
and~\ref{fig:fineweb_sf-muon_comparison_720M},
AMUSE consistently achieves lower validation perplexity
at both Llama model scales.
On image classification benchmarks, Figure~\ref{fig:image_sf-muon_comparison} shows that AMUSE achieves higher test accuracy, while fixed-$\beta$ SF-Muon tends to saturate earlier in training.

These results suggest that a fixed interpolation coefficient cannot effectively balance rapid early-stage progress with consistent improvement later in training.
In contrast, by gradually increasing $\beta_t$, AMUSE enables faster progress in the early stages while maintaining stable improvements later in training.

\FloatBarrier
\paragraph{Evaluating the Proposed $\beta_t$ Schedule.}\mbox{}
\label{app:beta_schedule_comparison}
To evaluate the design of the $\beta_t$ schedule in Eq.~\eqref{eq:amuse_beta}, we compare it with tuned linear and cosine alternatives. This ablation tests whether simply increasing $\beta_t$ is sufficient to explain AMUSE's gains. Table~\ref{tab:beta_schedule_comparison} shows that the proposed schedule yields the lowest validation perplexity at 16K iterations. The alternatives require the final iteration $T$ through normalized training progress, whereas Eq.~\eqref{eq:amuse_beta} is independent of the training horizon.

\begin{table}[!ht]
\centering
\caption{Comparison of the $\beta_t$ schedule in Eq.~\eqref{eq:amuse_beta} with tuned linear and cosine alternatives. Lower validation perplexity is better.}
\label{tab:beta_schedule_comparison}
\small
\begin{tabular}{lc}
\toprule
$\beta_t$ schedule & Validation perplexity at 16K \\
\midrule
Linear & 29.18 \\
Cosine & 29.90 \\
AMUSE (Eq.~\eqref{eq:amuse_beta}) & \textbf{28.48} \\
\bottomrule
\end{tabular}
\end{table}

\FloatBarrier

\paragraph{Can We Stop Increasing $\beta_t$ After It Becomes Large?}\mbox{}
We further test whether AMUSE's improvement comes only from reaching a specific large interpolation value. 
To do so, we use the best hyperparameter for AMUSE, $\beta_{\rm init}=0.6$ and $\rho = 0.8$, but cap $\beta_t$ at the best fixed value used by SF-Muon, $\beta=0.95$. 
That is, after $\beta_t$ reaches $0.95$, we keep it fixed instead of allowing it to continue increasing toward $1$.

\begin{figure}[ht]
\centering
\includegraphics[width=1\linewidth]{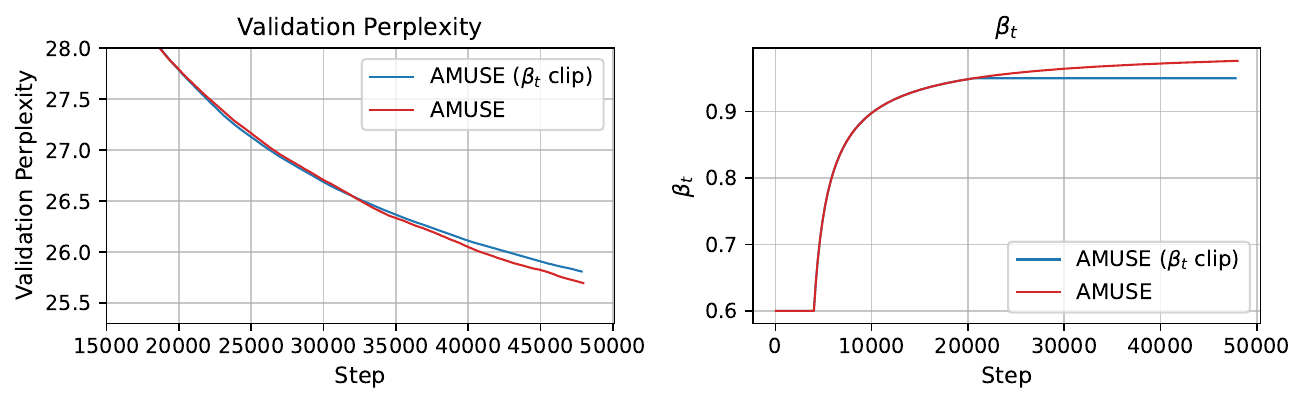}
\caption{
\textbf{Effect of stopping the increase of $\beta_t$ in AMUSE.}
(\textbf{Left}) Validation perplexity on 124M Llama pretraining. The clipped variant underperforms AMUSE, with the gap widening later in training.
(\textbf{Right}) Evolution of $\beta_t$. While both methods reach similar large values, AMUSE continues to increase $\beta_t$, whereas the clipped variant saturates to 0.95.
}
\label{fig:amuse_beta_clip}
\end{figure}

As shown in Figure~\ref{fig:amuse_beta_clip}, the clipped schedule performs worse than AMUSE, and the gap increases later in training. This indicates that AMUSE benefits not only from using a large $\beta_t$, but also from continuously shifting the evaluation point toward the averaged trajectory.

\begin{figure}[ht]
    \centering
    \begin{subfigure}{0.48\linewidth}
        \centering
        \includegraphics[width=\linewidth]{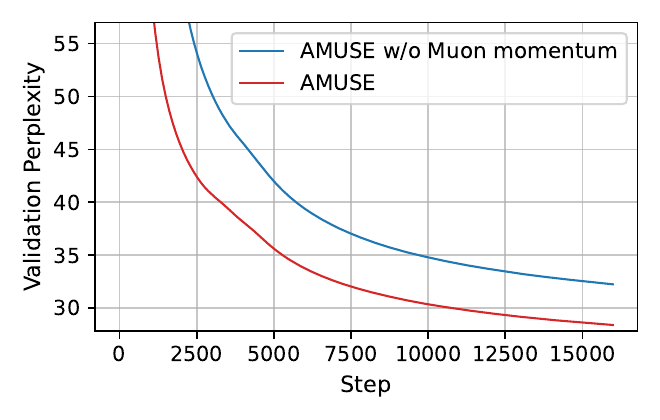}
        \caption{AMUSE w/o Muon momentum}
        \label{fig:fineweb_amuse_wo_momentum}
    \end{subfigure}
    \hfill
    \begin{subfigure}{0.485\linewidth}
        \centering
        \includegraphics[width=\linewidth]{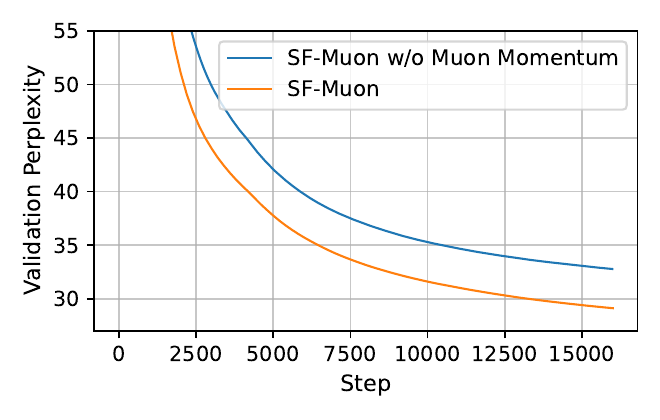}
        \caption{SF-Muon w/o Muon momentum}
        \label{fig:fineweb_sfmuon_wo_momentum}
    \end{subfigure}
    \caption{
\textbf{    Momentum ablation on 124M Llama pretraining.}
    We compare each method with and without Muon momentum, keeping all other hyperparameters fixed.
    Removing Muon momentum substantially degrades both AMUSE and SF-Muon, showing that momentum is essential for stable Muon-based optimization.
    }
    \label{fig:fineweb_muon_wo_momentum}
\end{figure}

\paragraph{AMUSE without Muon Momentum.}\mbox{}
To examine the role of Muon momentum in AMUSE and SF-Muon, we remove the momentum buffer used before the orthogonalization step. 
As shown in Figure~\ref{fig:fineweb_muon_wo_momentum}, removing Muon momentum causes a large performance drop for both methods. 
This indicates that momentum is crucial for Muon-based optimization: it smooths stochastic gradient noise before orthogonalization and accumulates a more reliable update direction, including useful bulk components. 
Without momentum, the orthogonalization step can amplify noisy components, resulting in less stable and less effective training.

\begin{figure}[ht]
    \centering
    \includegraphics[width=0.6\linewidth]{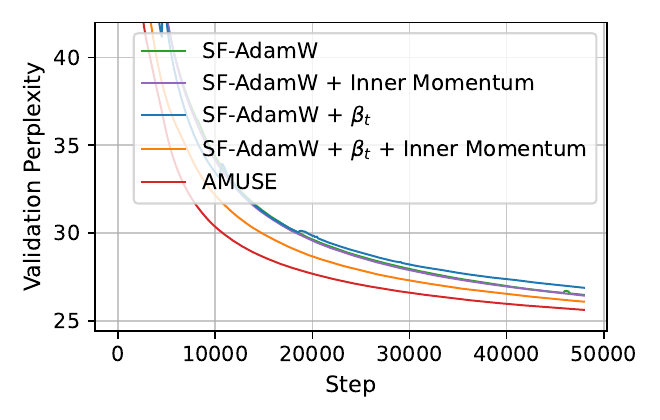}
    \caption{\textbf{Effect of $\beta_t$ scheduling and momentum in SF-AdamW.}
    We evaluate whether the modified $\beta_t$ schedule used in AMUSE can improve SF-AdamW in 124M Llama pretraining. Applying only the $\beta_t$ schedule ($\beta_{\text{init}}=0.6$, $\rho=0.6$) degrades performance relative to the original SF-AdamW. However, when combined with AdamW's first-moment ($\beta_1 =0.9$), the $\beta_t$ schedule improves validation perplexity throughout training, benefiting both the early and late training stages. Hyperparameters were tuned separately for the schedule variant.}
    \label{fig:sfadamw_beta_t}
\end{figure}

\paragraph{Applying $\beta_t$ Scheduling and Momentum to SF-AdamW.}\mbox{}
\label{app:sfadamw_beta}
We investigate whether the modified $\beta_t$ schedule introduced in Eq.~\eqref{eq:amuse_beta} can also improve the performance of SF-AdamW. As shown in Figure~\ref{fig:sfadamw_beta_t}, applying only the $\beta_t$ schedule does not improve performance over the original SF-AdamW. In contrast, when combined with AdamW's first-moment ($\beta_1=0.9$), the $\beta_t$ schedule consistently improves performance throughout training, yielding lower validation perplexity in both the early and late stages.

Interestingly, neither the $\beta_t$ schedule nor first-moment alone improves performance over the original SF-AdamW. In contrast, applying both together consistently improves validation perplexity throughout training, indicating that the two components are effective only when used in combination.

However, incorporating first-moment in AdamW into SF-AdamW requires maintaining an additional momentum buffer, resulting in higher memory consumption than either AdamW or AMUSE.

\subsection{Robustness Across Random Seeds}
\label{app:llm_multiseed}
We repeat 124M Llama pretraining for 16K iterations with five random seeds for each method. All hyperparameters are fixed across seeds, with no per-seed retuning, and each run is evaluated on the full evaluation set. Table~\ref{tab:llm_multiseed_124m} reports the mean and standard deviation across seeds. AMUSE outperforms both baselines in all five matched-seed runs, reducing mean validation perplexity by $0.164$ relative to Muon and by $0.557$ relative to SF-Muon.

\begin{table}[H]
\centering
\small
\caption{Validation perplexity of 124M Llama models after 16K pretraining iterations, evaluated on the full evaluation set. Values are mean $\pm$ standard deviation across five random seeds; lower is better.}
\label{tab:llm_multiseed_124m}
\begin{tabular}{lc}
\toprule
Method & Validation PPL $\downarrow$ \\
\midrule
SF-Muon & $29.073 \pm 0.041$ \\
Muon & $28.680 \pm 0.042$ \\
AMUSE & $\mathbf{28.516} \pm \mathbf{0.041}$ \\
\bottomrule
\end{tabular}
\end{table}

\clearpage

%% file: checklist.tex
\section*{NeurIPS Paper Checklist}

\begin{enumerate}

\item {\bf Claims}
    \item[] Question: Do the main claims made in the abstract and introduction accurately reflect the paper's contributions and scope?
    \item[] Answer:\answerYes{} 
    \item[] Justification: The abstract and introduction clearly outline our main contributions, including the proposal of the AMUSE optimizer and its motivation, which are fully supported by our empirical findings.

\item {\bf Limitations}
    \item[] Question: Does the paper discuss the limitations of the work performed by the authors?
    \item[] Answer: \answerYes{} 
    \item[] Justification: We discuss the limitations of our approach in the Implementation Details (Section~\ref{sec:Impldetails}) and Conclusion (Section~\ref{sec:conclusion}). Specifically, we address the introduction of the new hyperparameter $\rho$ and the computational overhead of maintaining the averaged iterates.

\item {\bf Theory assumptions and proofs}
    \item[] Question: For each theoretical result, does the paper provide the full set of assumptions and a complete (and correct) proof?
    \item[] Answer: \answerNA{} 
    \item[] Justification: This paper does not contain formal theorems or lemmas requiring mathematical proofs.

    \item {\bf Experimental result reproducibility}
    \item[] Question: Does the paper fully disclose all the information needed to reproduce the main experimental results of the paper to the extent that it affects the main claims and/or conclusions of the paper (regardless of whether the code and data are provided or not)?
    \item[] Answer: \answerYes{}
    \item[] Justification: We provide all necessary details for reproducibility, including model architectures, dataset specifications, and the exact hyperparameter configurations in Section~\ref{sec:Impldetails} and Appendix~\ref{app:image domain detail} and \ref{app:llm_detail}. The core algorithm is also summarized in Algorithm~\ref{alg:amuse}.

\item {\bf Open access to data and code}
    \item[] Question: Does the paper provide open access to the data and code, with sufficient instructions to faithfully reproduce the main experimental results, as described in supplemental material?
    \item[] Answer: \answerYes{}
    \item[] Justification: Our implementation and usage instructions are publicly available at \url{https://github.com/kjeiun/amuse}.

\item {\bf Experimental setting/details}
    \item[] Question: Does the paper specify all the training and test details (e.g., data splits, hyperparameters, how they were chosen, type of optimizer) necessary to understand the results?
    \item[] Answer: \answerYes{}
    \item[] Justification: We specify all experimental configurations, including dataset details (e.g., FineWeb-100B), model architectures, and comprehensive hyperparameter tables for all baselines in Appendix~\ref{app:image domain detail} and~\ref{app:llm_detail}.

\item {\bf Experiment statistical significance}
    \item[] Question: Does the paper report error bars suitably and correctly defined or other appropriate information about the statistical significance of the experiments?
    \item[] Answer: \answerYes{}
    \item[] Justification: Most image-domain results are averaged over multiple independent runs and reported with standard deviations or shaded error bands. We additionally report mean and standard deviation across five random seeds for 124M LLM pretraining (Appendix~\ref{app:llm_multiseed}). Other LLM pretraining results are reported as single runs.

\item {\bf Experiments compute resources}
    \item[] Question: For each experiment, does the paper provide sufficient information on the computer resources (type of compute workers, memory, time of execution) needed to reproduce the experiments?
\item[] Answer: \answerYes{}
    \item[] Justification: Detailed information regarding the hardware resources used for our experiments, including the specific GPU types and memory is provided in Appendix~\ref{app:implementation_details}.
    
\item {\bf Code of ethics}
    \item[] Question: Does the research conducted in the paper conform, in every respect, with the NeurIPS Code of Ethics \url{https://neurips.cc/public/EthicsGuidelines}?
\item[] Answer: \answerYes{}
    \item[] Justification: Our research adheres to the NeurIPS Code of Ethics. The study focuses on optimization algorithms and does not involve human subjects, sensitive private data, or any ethically controversial practices.

\item {\bf Broader impacts}
    \item[] Question: Does the paper discuss both potential positive societal impacts and negative societal impacts of the work performed?
    \item[] Answer: \answerNA{}
    \item[] Justification: This work focuses on foundational optimization algorithms and does not have a direct societal impact. As a general-purpose mathematical tool for training neural networks, any potential impact would depend entirely on the specific application it is used for.
    
\item {\bf Safeguards}
    \item[] Question: Does the paper describe safeguards that have been put in place for responsible release of data or models that have a high risk for misuse (e.g., pre-trained language models, image generators, or scraped datasets)?
\item[] Answer: \answerNA{}
    \item[] Justification: This paper proposes a general-purpose optimization algorithm and does not release high-risk assets such as large-scale pre-trained models or sensitive datasets that would require specific safeguards.

\item {\bf Licenses for existing assets}
    \item[] Question: Are the creators or original owners of assets (e.g., code, data, models), used in the paper, properly credited and are the license and terms of use explicitly mentioned and properly respected?
\item[] Answer: \answerYes{}
    \item[] Justification: We properly cite the original creators of the datasets and baseline models used in our study. All assets are used in compliance with their respective open-source licenses.

\item {\bf New assets}
    \item[] Question: Are new assets introduced in the paper well documented and is the documentation provided alongside the assets?
\item[] Answer: \answerYes{}
    \item[] Justification: We introduce the AMUSE optimizer as a new asset. A complete description of the algorithm is provided in the main text, and documented implementation code is included in the supplemental material for reproducibility.

\item {\bf Crowdsourcing and research with human subjects}
    \item[] Question: For crowdsourcing experiments and research with human subjects, does the paper include the full text of instructions given to participants and screenshots, if applicable, as well as details about compensation (if any)? 
    \item[] Answer: \answerNA{}
    \item[] Justification: This research does not involve crowdsourcing nor research with human subjects.

\item {\bf Institutional review board (IRB) approvals or equivalent for research with human subjects}
    \item[] Question: Does the paper describe potential risks incurred by study participants, whether such risks were disclosed to the subjects, and whether Institutional Review Board (IRB) approvals (or an equivalent approval/review based on the requirements of your country or institution) were obtained?
    \item[] Answer: \answerNA{}
    \item[] Justification: This research does not involve human subjects, and therefore does not require IRB approval or equivalent ethical reviews.

\item {\bf Declaration of LLM usage}
    \item[] Question: Does the paper describe the usage of LLMs if it is an important, original, or non-standard component of the core methods in this research? Note that if the LLM is used only for writing, editing, or formatting purposes and does \emph{not} impact the core methodology, scientific rigor, or originality of the research, declaration is not required.
    \item[] Answer: \answerNA{}
    \item[] Justification: LLMs were not used as a component of the core methodology or for generating original research results. Their usage was limited to standard writing and editing assistance, which does not require declaration according to the NeurIPS guidelines.

\end{enumerate}